\documentclass{article}

\PassOptionsToPackage{noabbrev}{cleveref}
\usepackage[final]{neurips_2020}

\usepackage[utf8]{inputenc} 
\usepackage[T1]{fontenc}    
\usepackage{hyperref}       
\usepackage{url}            
\usepackage{booktabs}       
\usepackage{amsfonts}       
\usepackage{nicefrac}       
\usepackage{microtype}      
\usepackage{lipsum}
\usepackage{graphicx}
\usepackage{amsmath}
\usepackage{cleveref}
\usepackage{wrapfig}
\usepackage{placeins}
\usepackage{caption}
\usepackage{subcaption}

\usepackage{tikz}
\usetikzlibrary{arrows, arrows.meta, bayesnet, chains, positioning, decorations.pathreplacing, fit, calc, decorations.pathmorphing}

\usepackage{layouts}  
\tikzset{scaled/.style={scale=#1}}
\tikzset{scaled/.default=1.}

\definecolor{depth1}{HTML}{ff7f0e}
\definecolor{depth2}{HTML}{2ca02c}
\definecolor{depth3}{HTML}{d62728}
\definecolor{depth4}{HTML}{9467bd}
\definecolor{depth5}{HTML}{17becf}
\tikzset{nicecolor/.style={draw=#1!50!black, fill=#1!40!white}}
    
\tikzset{block/.style={rectangle, text width=2ex, align=center, minimum height=2ex, nicecolor=#1, font=\scshape, rounded corners}}
\tikzset{block/.default=red}

\tikzset{opnode/.style={circle, draw, minimum size=1ex, text width=1ex}}
\tikzset{plus/.style ={opnode, font={$+$}}}

\tikzset{line/.style={line width=.15ex, draw=black}}
\tikzset{arrow/.style={-{Latex[length=1.3ex,width=0.8ex]},line}}
\tikzset{arrowsm/.style={-{Latex[length=1.ex,width=0.66ex]},line}}
\tikzset{snake/.style={decoration={snake, pre length=0.01mm, segment length=2mm, amplitude=0.2mm, post length=1mm}, decorate}}

\definecolor{mplblue}{rgb}{0.12156862745098039, 0.4666666666666667, 0.7058823529411765}
\definecolor{mplorange}{rgb}{1.0, 0.4980392156862745, 0.054901960784313725}
\definecolor{mplgreen}{rgb}{0.17254901960784313, 0.6274509803921569, 0.17254901960784313}
\definecolor{mplred}{rgb}{0.8392156862745098, 0.15294117647058825, 0.1568627450980392}
\definecolor{mplpurple}{rgb}{0.5803921568627451, 0.403921568627451, 0.7411764705882353}
\definecolor{mplbrown}{rgb}{0.5490196078431373, 0.33725490196078434, 0.29411764705882354}
\definecolor{mplpink}{rgb}{0.8901960784313725, 0.4666666666666667, 0.7607843137254902}
\definecolor{mplgrey}{rgb}{0.4980392156862745, 0.4980392156862745, 0.4980392156862745}
\definecolor{mplyellow}{rgb}{0.7372549019607844, 0.7411764705882353, 0.13333333333333333}
\definecolor{mplcyan}{rgb}{0.09019607843137255, 0.7450980392156863, 0.8117647058823529}

\title{Depth Uncertainty Networks for Active Learning}

\author{%
  Chelsea Murray \hspace{2.5mm} 
  James U. Allingham \hspace{2.5mm}
  Javier Antor\'an \hspace{2.5mm}
  Jos\'e Miguel Hern\'andez-Lobato \\
  Department of Engineering \\
  University of Cambridge \\
  \texttt{\{clm88, jua23, ja666, jmh233\}@cam.ac.uk}
}

\begin{document}

\maketitle

\begin{abstract}
  In active learning, the size and complexity of the training dataset changes over time.
  Simple models that are well specified by the amount of data available at the start of active learning might suffer from bias as more points are actively sampled. 
  Flexible models that might be well suited to the full dataset can suffer from overfitting towards the start of active learning.
  We tackle this problem using Depth Uncertainty Networks (DUNs), a BNN variant in which the depth of the network, and thus its complexity, is inferred. 
  We find that DUNs outperform other BNN variants on several active learning tasks. Importantly, we show that on the tasks in which DUNs perform best they present notably less overfitting~than~baselines.  
\end{abstract}

\section{Introduction}

In many important applications of machine learning, labelled data can be scarce and expensive to obtain. Active learning is one approach to improving data efficiency in such cases. Given a fixed labelling budget, the objective of active learning is to identify examples that maximise the expected gain in model performance. Bayesian Neural Networks (BNNs) are a natural fit for active learning problems as they provide uncertainty estimates that can be used as a measure of the informativeness of unlabelled data points. This work focuses on a recently proposed form of BNN, Depth Uncertainty Networks (DUNs), in which probabilistic inference is performed over the depth of the network \citep{antoran2020depth}. We hypothesise that the ability to infer depth enables DUNs to adapt model complexity to the varying size of the training dataset as additional labels are acquired. This is expected to result in reduced overfitting in DUNs in the early stages of active learning, and in better performance of DUNs relative to other BNN methods.

\section{Background}

\subsection{Active learning}

In the active learning framework, a model is initially trained on a small labelled subset of the available data, and additional unlabelled points are selected via an \textit{acquisition function} to be labelled by an external \textit{oracle} (e.g. a human expert) \citep{settles2009active}. Given a model with parameters $\theta$ trained on training data $\mathcal{D}_{\text {train}}=\left\{\mathbf{x}_{i}, \mathbf{y}_{i}\right\}_{i=1}^{N}$, the acquisition function $\alpha(\cdot)$ 
scores all unlabelled examples in the pool set $\mathcal{D}_{\text {pool }} = \left\{\mathbf{x}_{j}\right\}_{j=1}^{N_{\text {pool }}}$. These scores are used to select the next point $\mathbf{x}^{\star}$ to be labelled:
\begin{equation}
\mathbf{x}^{\star}=\underset{\mathbf{x} \in \mathcal{D}_{\text {pool }}}{\operatorname{argmax}}\ \alpha\left(\mathbf{x} ; \theta, \mathcal{D}_{\text {train}}\right). 
\end{equation}
\citet{houlsby2011bayesian} propose an aquistion called Bayesian Active Learning by Disagreement (BALD),
 \begin{equation}
    \alpha_{\mathrm{BALD}}\left(\mathbf{x} ; \theta, \mathcal{D}_{\text {train}}\right) = \mathbb{H}\left[\mathbf{y} \mid \mathbf{x}, \mathcal{D}_{\text {train}}\right]-\mathbb{E}_{\theta \sim p\left(\theta \mid \mathcal{D}_{\text {train}}\right)}[\mathbb{H}[\mathbf{y} \mid \mathbf{x}, \theta]] ,\label{eq:bald_houlsby}
\end{equation}
which selects points for which the predictions of individual parameterisations maximally disagree---i.e., where there is high uncertainty in the predictive posterior on average---but the predictions of individual parameter settings are confident. To avoid acquiring correlated points when performing batch acquisition, we implement a stochastic relaxation of BALD, 
\begin{equation}
    \mathbf{x}^{\star} \sim \alpha_{\mathrm{BALDstoch}}(\mathbf{x}; \theta, \mathcal{D}_{\text{train}}) = \mathtt{softmax}\left(T\alpha_{\mathrm{BALD}}(\mathbf{x}; \theta, \mathcal{D}_{\text{train}})\right). 
\end{equation}
We use temperature $T = 10$; see \cref{app:temp} for discussion of this choice. \citet{kirsch2021stochastic} provide extensive analysis showing that such stochastic relaxations are never worse than their deterministic counterparts, and can even outperform computationally expensive batch-aware methods.

\subsection{Depth uncertainty networks}

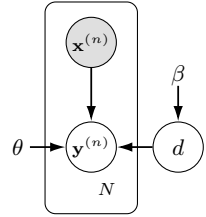
\begin{wrapfigure}[9]{r}{0.18\textwidth}
\vspace{-0.35in}
    \centering
    \begin{tikzpicture}[every node/.style={scale=0.95}]
      \node[obs] (xn) {\scriptsize $\mathbf{x}^{(n)}$};
      \node[latent, below=4.5ex of xn] (yn) {\scriptsize $\mathbf{y}^{(n)}$};
      \node[const, left=3ex of yn] (theta) {\footnotesize $\theta \ $};
      \node[latent, right=3ex of yn] (d) {\footnotesize $d$};
      \node[const, above=3ex of d] (beta) {\footnotesize $\beta$};
      
      \edge[arrow] {xn} {yn} ; %
      \edge[arrow] {theta} {yn} ;
      \edge[arrow] {d} {yn} ;
      \edge[arrow] {beta} {d} ;
      
      \plate[inner sep=2ex] {} {(yn)(xn)} {\scriptsize $N$} ;
    \end{tikzpicture}
\caption{DUN graphical model.}
\label{fig:dun_only_graph_model}
\vspace{-0.1in}
\end{wrapfigure}


DUNs, depicted in \cref{fig:dun_only_graph_model}, are a BNN variant in which the depth of the network $d \in [1,2,\!...,\!D]$ is treated as a random variable. Model weights $\theta$ are kept deterministic, simplifying optimisation. We place a categorical prior over depth $p(d)\,{=}\,\mathrm{Cat}(d| \{\beta_{i}\}_{i=0}^{D})$, resulting in a categorical posterior which can be marginalised in closed form. This procedure can be seen as Bayesian model averaging over an ensemble of subnetworks of increasing depth.
Thus, for DUNs, the predictive distribution and $\alpha_{\mathrm{BALD}}(\cdot)$ are tractable and cheap to compute. See \cref{app:duns} or \citep{antoran2020depth} for a detailed description.

In DUNs, predictions from different depths of the network correspond to different kinds of functions---shallow networks induce simple functions, while deeper networks induce more complex functions. We hypothesise that DUNs will automatically adapt model flexibility during active learning by inferring initially shallow networks, and progressively deeper networks as more data are observed. This capability is expected to reduce overfitting in the small data regimes typical of active learning, and result in improved performance for DUNs relative to other BNN methods. It is worth noting that automatically adapting model complexity to observed data is also a promised feature of traditional BNNs. However, unlike in DUNs where inference is exact, weight space models require crude approximations which fall short in this regard~\citep{JMLR:v20:19-236}.

\section{Results}

We perform experiments on
five toy regression datasets, nine UCI regression datasets \citep{hernandez2015probabilistic}, and MNIST. Concrete, Energy, and Kin8nm are presented in the following sections, with the rest provided in \cref{app:add_results}. The experimental setup is described in \cref{app:exp_setup}. Experiments are repeated 40 times, with the mean and standard deviations reported in our figures.

\subsection{DUN posteriors}

\begin{figure}[h!] 
\vspace{-.5em}
\centering
    \begin{subfigure}[b]{0.315\textwidth}
        \centering
        \includegraphics[width=\linewidth]{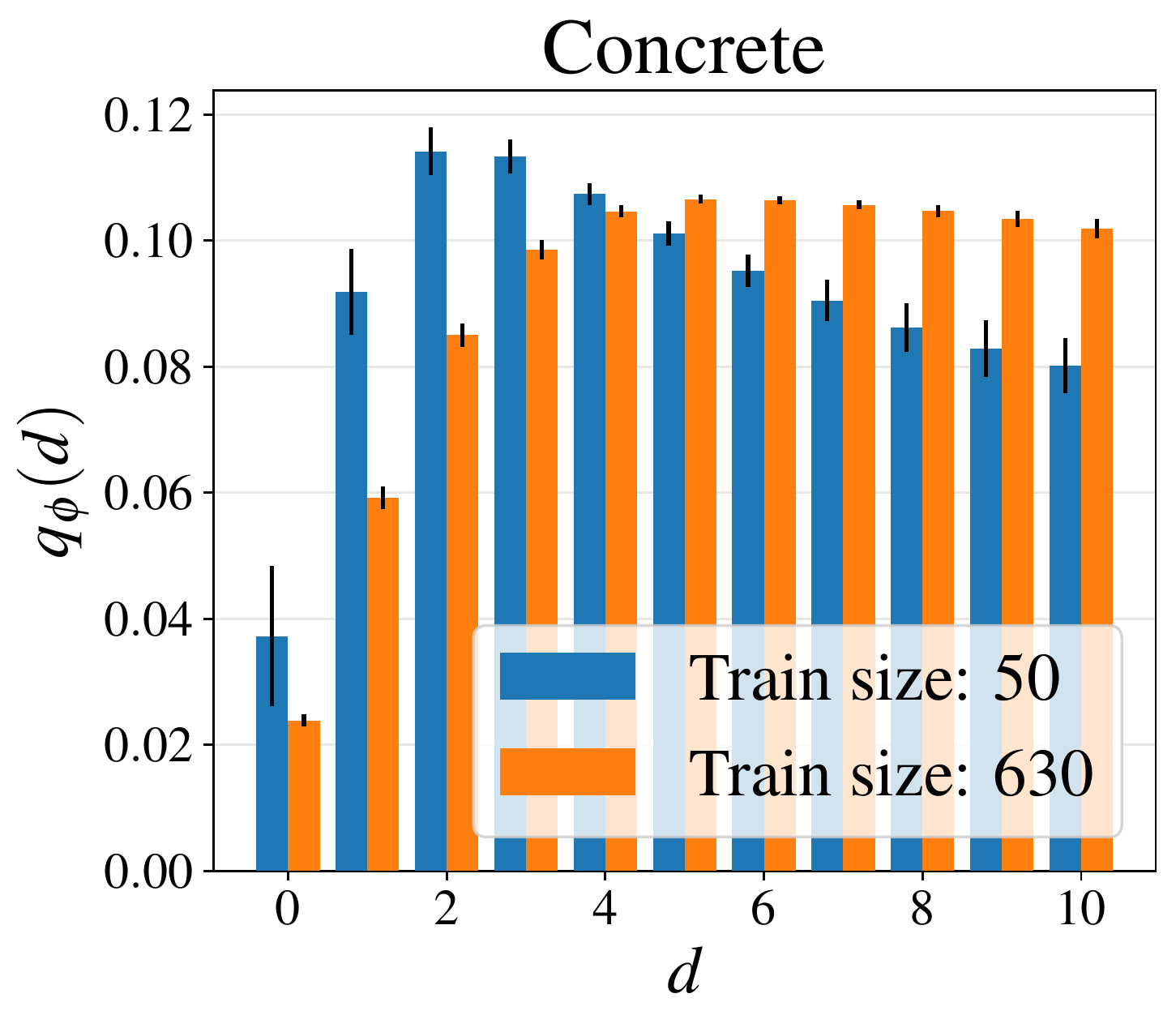}
    \end{subfigure}
    \begin{subfigure}[b]{0.29\textwidth}
        \centering
        \includegraphics[width=\linewidth]{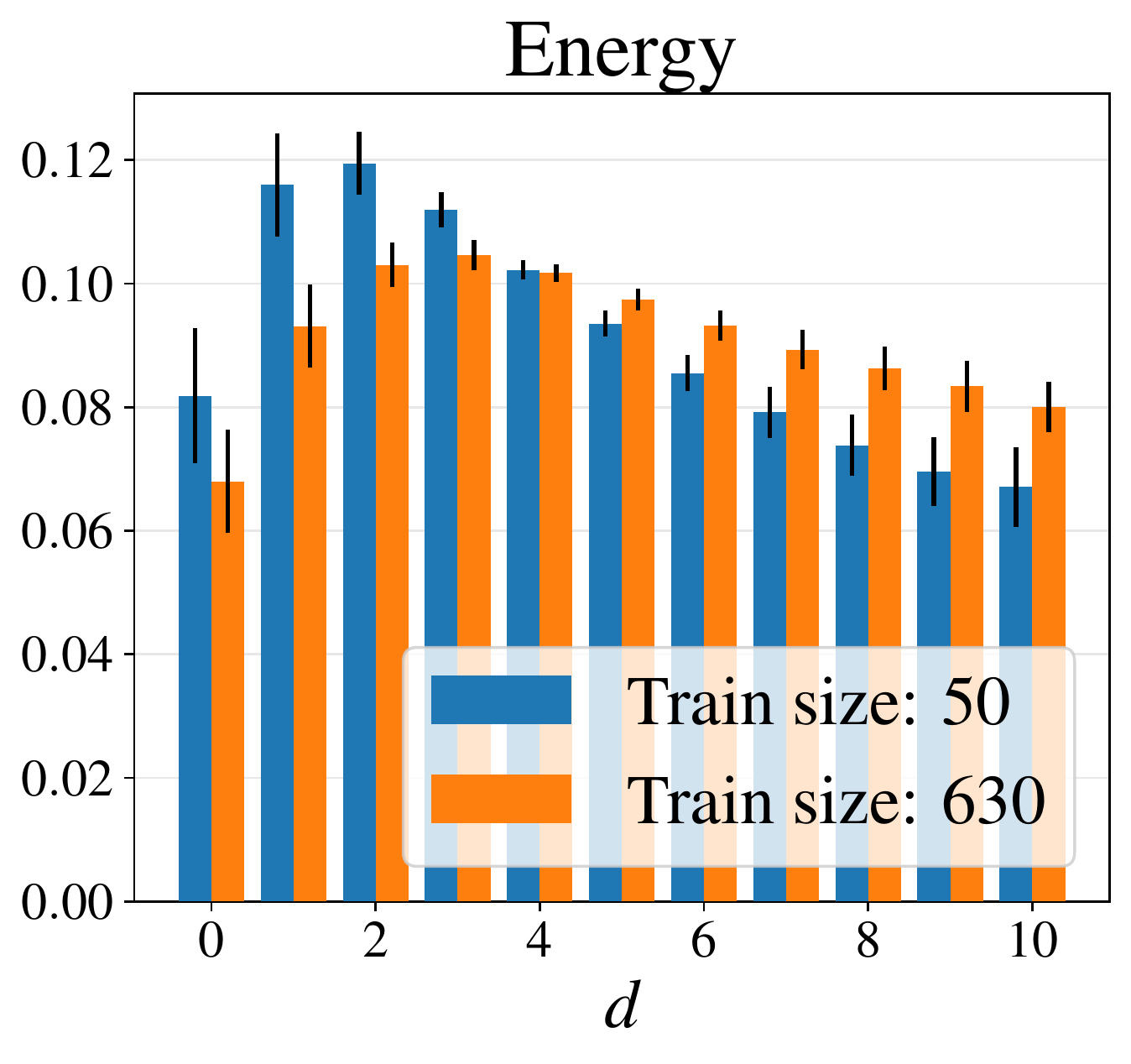}
    \end{subfigure} 
    \begin{subfigure}[b]{0.29\textwidth}
        \centering
        \includegraphics[width=\linewidth]{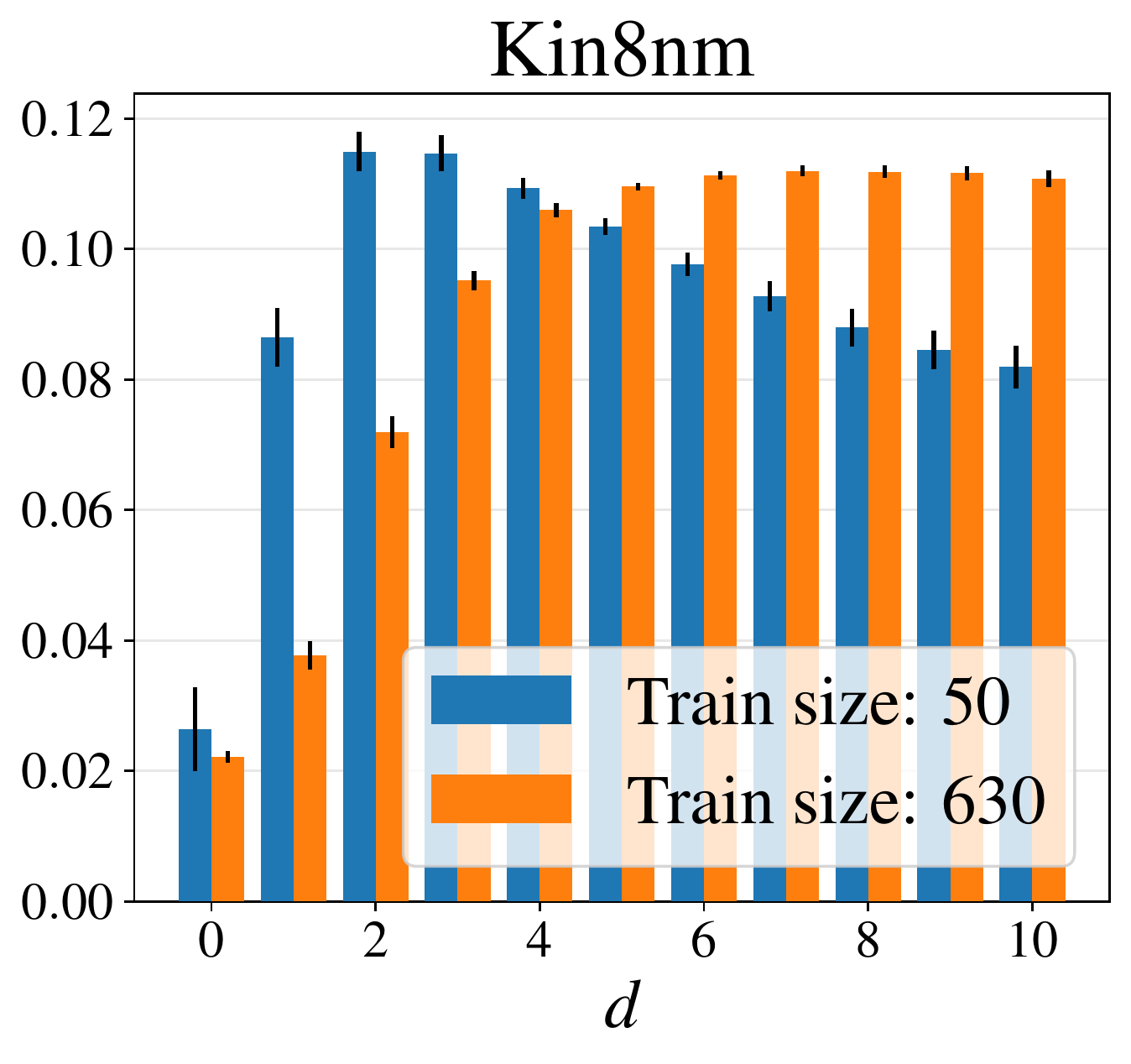} 
    \end{subfigure}
    \vspace{-1.0em}
    \caption{DUN depth posteriors for three UCI regression datasets, with the \textbf{\textcolor{mplblue}{smallest}} and \textbf{\textcolor{mplorange}{largest}} labelled datasets used during active learning. Additional datasets' results are in \cref{fig:app_res_reg_posts_app,fig:mnist_post}.}
    \label{fig:res_reg_posts}
    \vspace{-1.5em}
\end{figure}

First, we investigate whether our hypothesis that DUNs adapt their inferred depth to the size of the dataset holds. \Cref{fig:res_reg_posts} compares the posterior probabilities over depth for DUNs trained on datasets from the first and final steps of active learning, and illustrates that this does occur in practice. 

\subsection{DUNs versus other BNN methods}

\begin{figure}[h] 
\centering
\vspace{-0.3em}
    \begin{subfigure}[b]{0.305\textwidth}
        \centering
        \includegraphics[width=\linewidth]{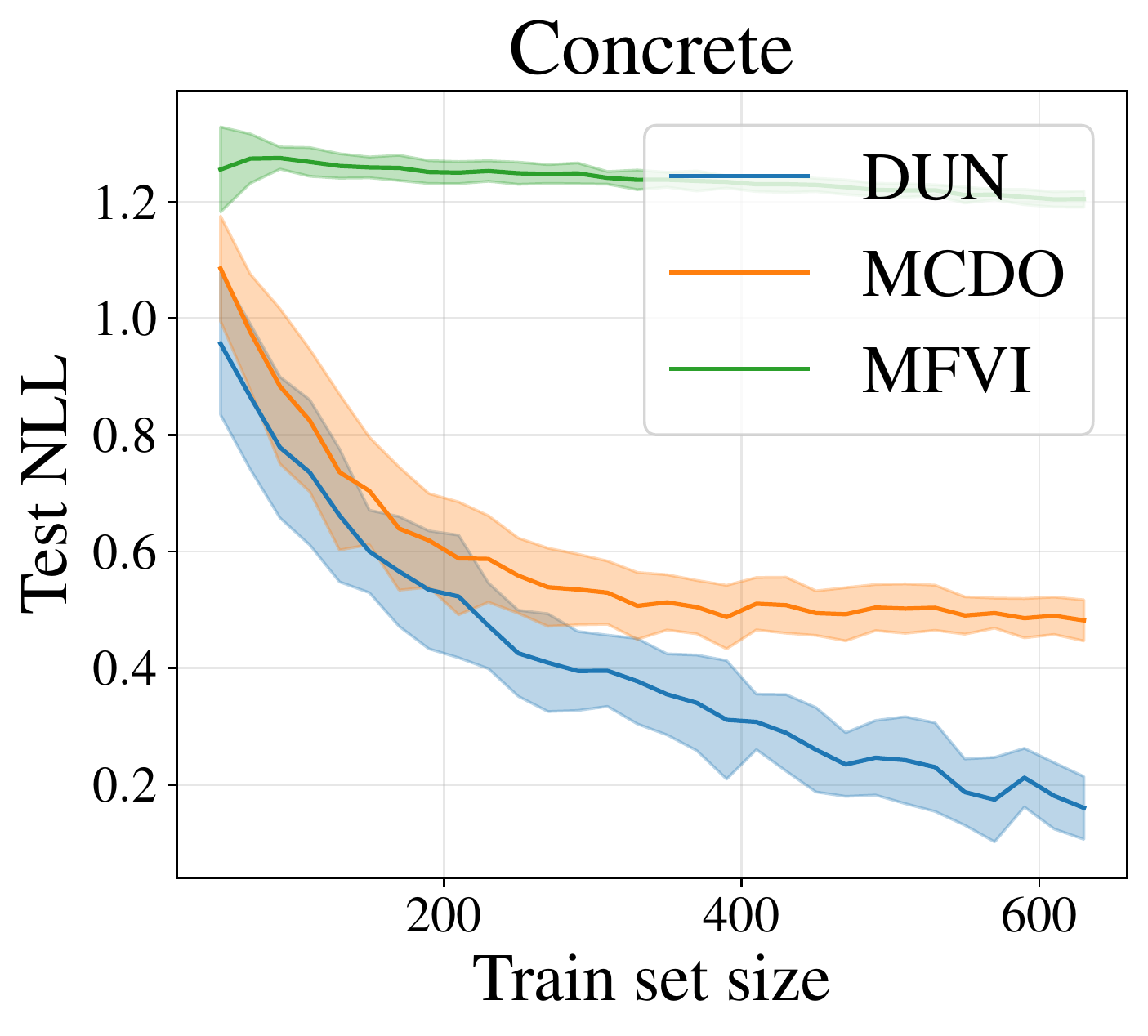}
    \end{subfigure}
    \begin{subfigure}[b]{0.303\textwidth}
        \centering
        \includegraphics[width=\linewidth]{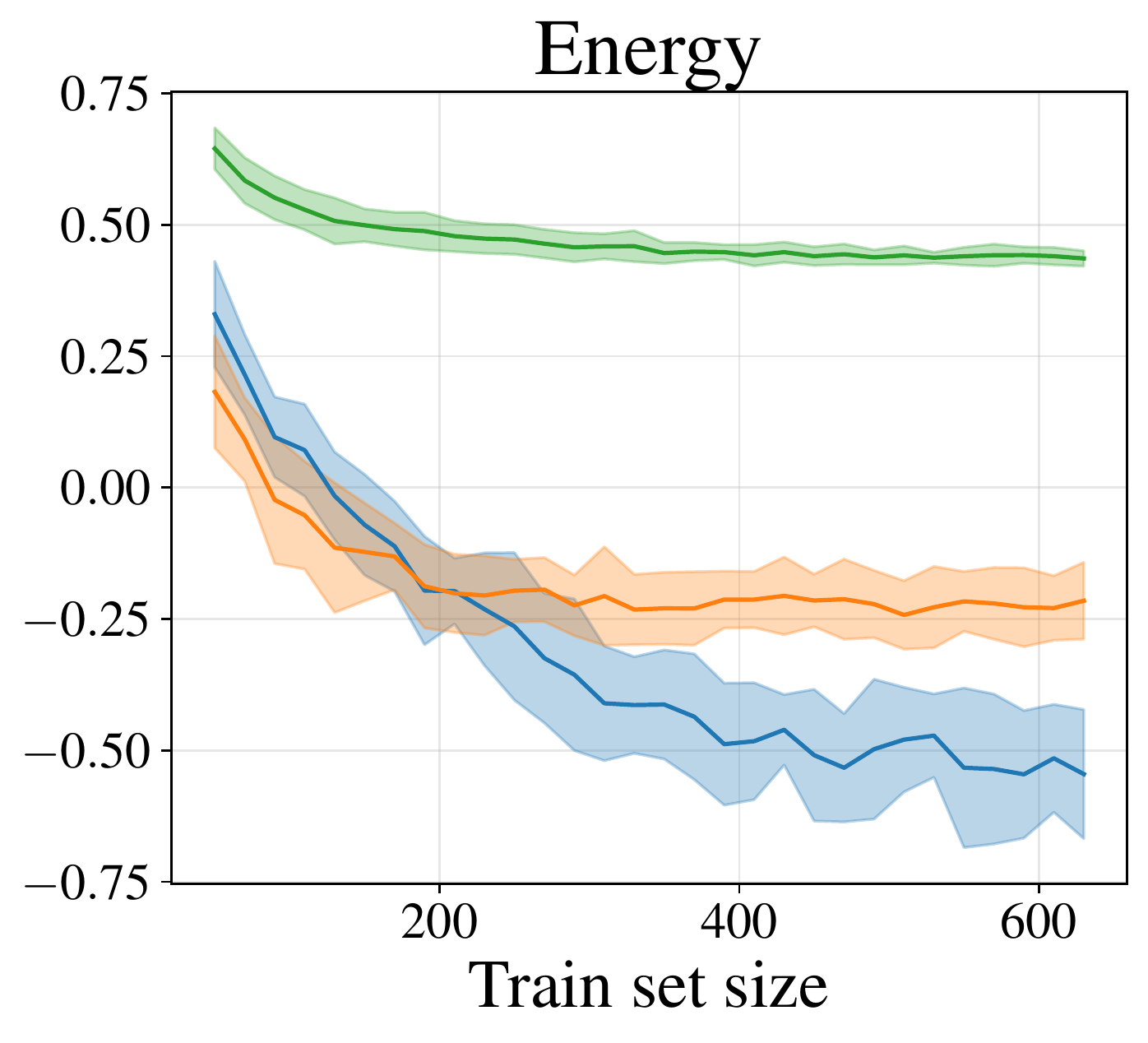}
    \end{subfigure} 
    \begin{subfigure}[b]{0.285\textwidth}
        \centering
        \includegraphics[width=\linewidth]{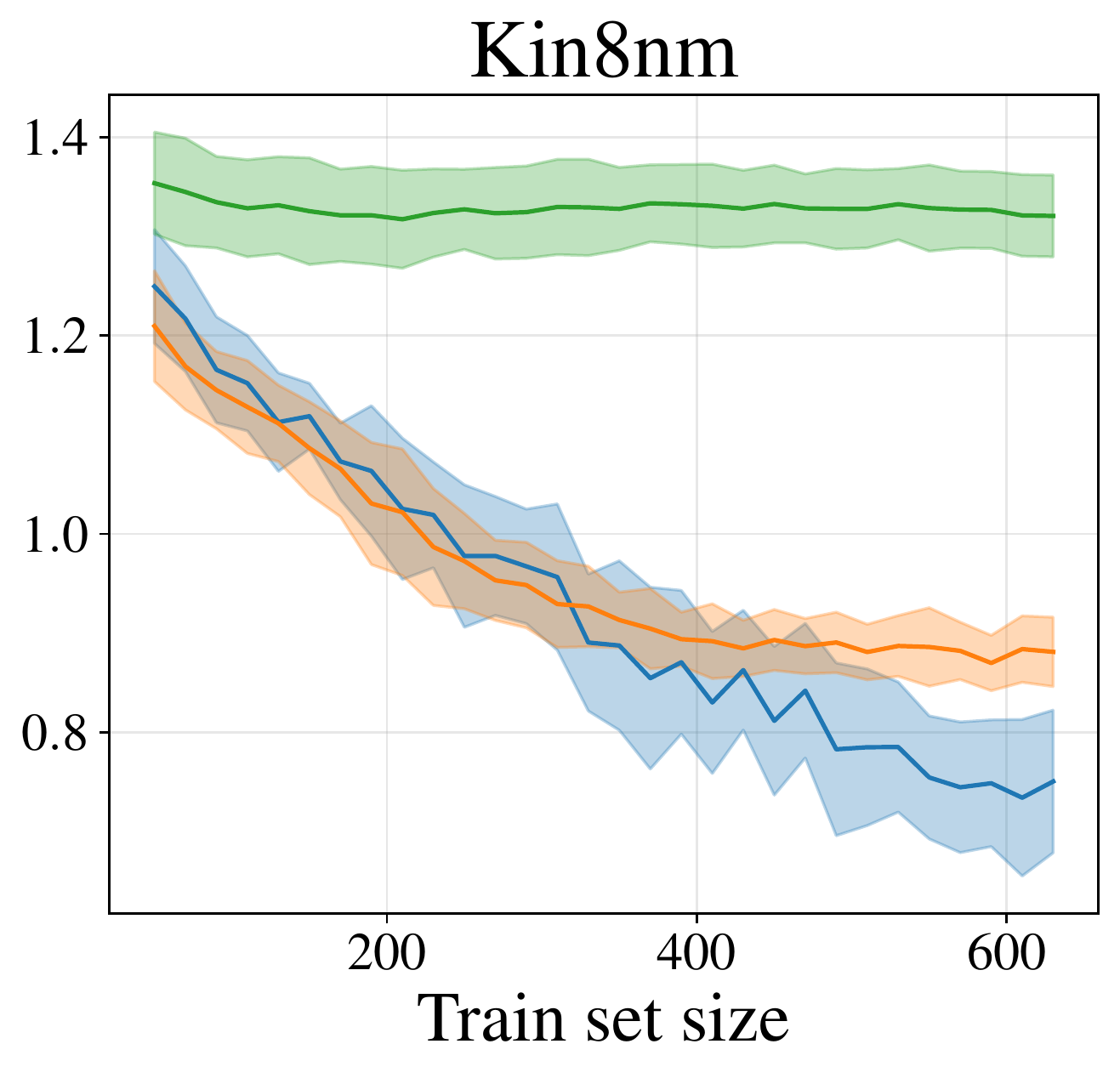} 
    \end{subfigure}
    \vspace{-1mm}
    \caption[NLL vs. number of training points for DUNs, MCDO and MFVI evaluated on UCI datasets.]{Test NLL vs. number of training points using stochastic BALD acquisition function evaluated on UCI datasets. \textbf{\textcolor{mplblue}{DUNs}}, \textbf{\textcolor{mplorange}{MCDO}} and \textbf{\textcolor{mplgreen}{MFVI}} are compared. Additional datasets' results are in \cref{fig:app_res_reg_methods_nll,fig:mnist_nll,fig:res_toy_methods_nll}. Results for classification error rather than NLL are in \cref{fig:res_reg_methods_err}. }   \label{fig:res_reg_methods_nll}
    \vspace{-0.5em}
\end{figure}

\begin{figure}[h!] 
\vspace{-0.42em}
\centering
    \begin{subfigure}[b]{0.315\textwidth}
        \centering
        \includegraphics[width=\linewidth]{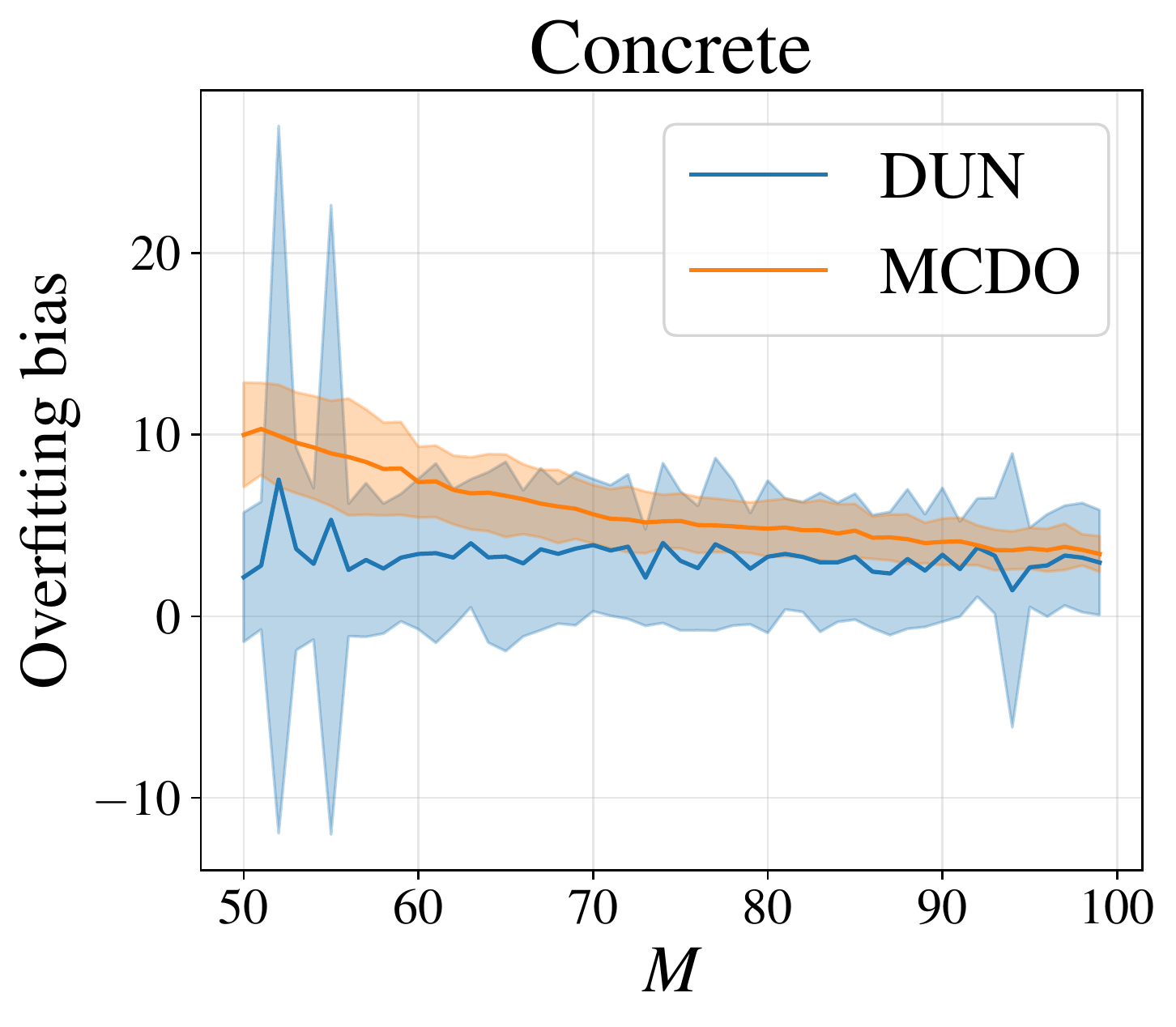}
    \end{subfigure}
    \begin{subfigure}[b]{0.288\textwidth}
        \centering
        \includegraphics[width=\linewidth]{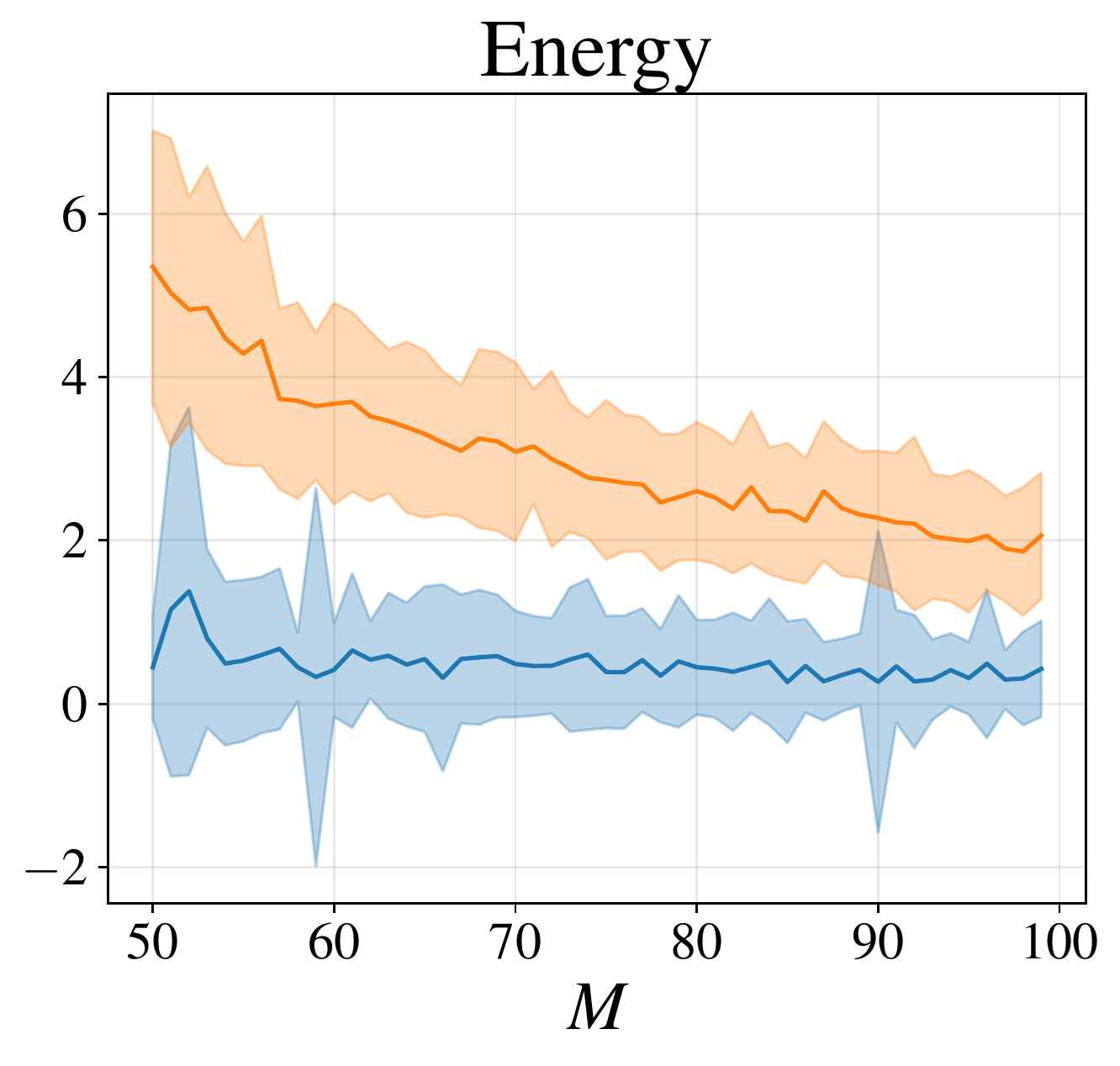}
    \end{subfigure} 
    \begin{subfigure}[b]{0.295\textwidth}
        \centering
        \includegraphics[width=\linewidth]{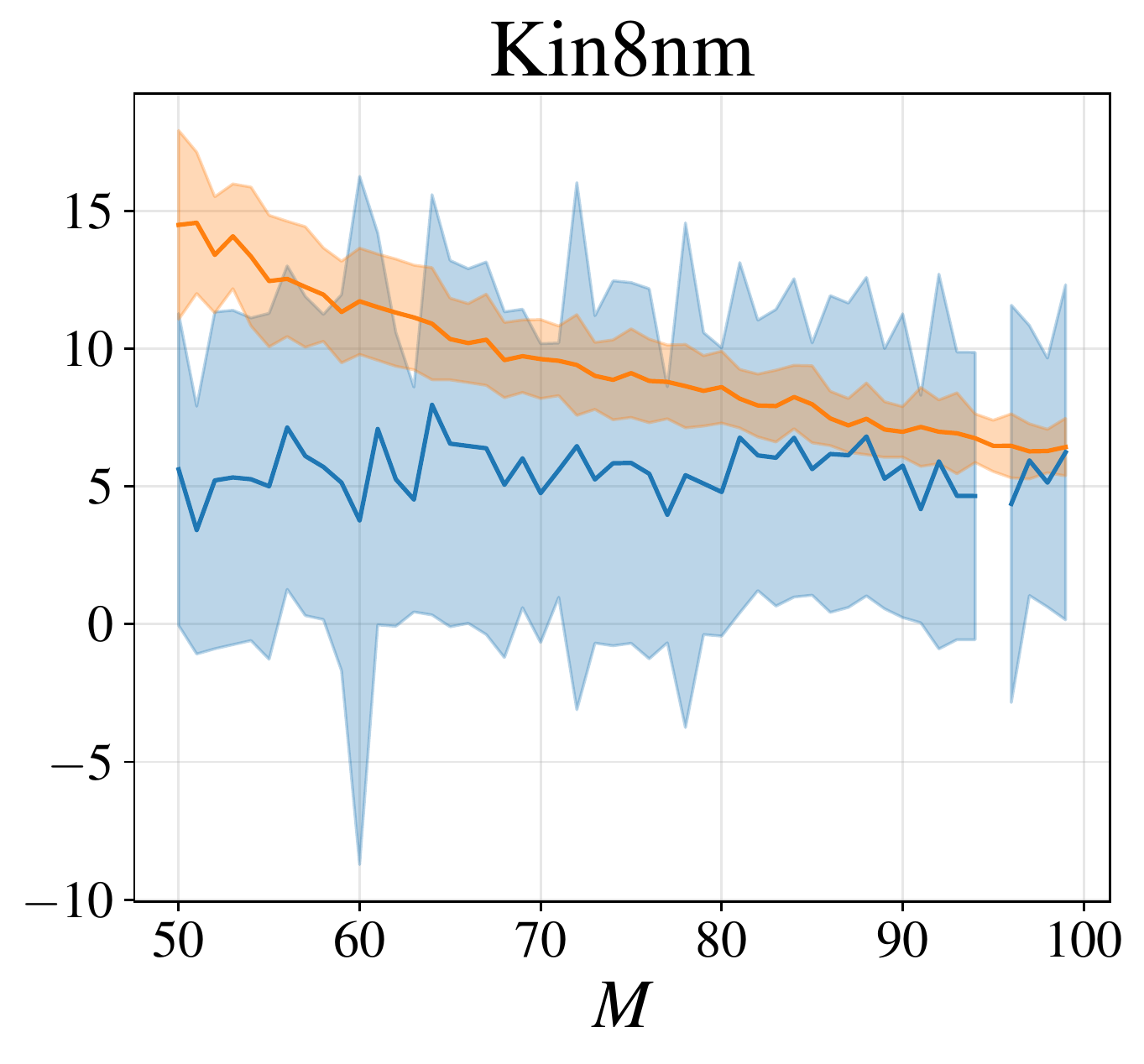} 
    \end{subfigure}
    \vspace{-1mm}
    \caption{Evolution of ``overfitting bias'', as described by~\citet{farquhar_statistical_2020}. See \cref{app:al_bias} for more details. For \textbf{\textcolor{mplblue}{DUNs}} and \textbf{\textcolor{mplorange}{MCDO}} as the amount of acquired data $M$ grows. DUNs are more robust to overfitting than MCDO. Results for additional datasets are in \cref{fig:app_res_ofb_DUNvMCDO}.}
    \label{fig:res_ofb_DUNvMCDO}
    \vspace{-1em}
\end{figure}

We compare the negative log-likelihood (NLL) performance of DUNs to mean-field variational inference (MFVI) \citep{blundell2015weight, graves2011practical} and Monte Carlo dropout (MCDO) \citep{gal2016dropout} in \cref{fig:res_reg_methods_nll}. The test NLL for MFVI and MCDO is evaluated using 10 Monte Carlo samples. DUNs perform either better than, or comparably to, MCDO, and clearly outperform MFVI. Following an approach proposed by \citet{farquhar_statistical_2020} for quantifying overfitting in the context of active learning problems, \cref{fig:res_ofb_DUNvMCDO} shows that the magnitude of overfitting in DUNs tends to be smaller relative to MCDO. Given the results in \cref{fig:res_reg_posts} and \cref{fig:res_ofb_DUNvMCDO}, it is plausible to conclude that DUNs' flexibility over depth is useful in active learning settings. DUNs' performance may also be attributable to better quality uncertainty estimates, as reported by \citet{antoran2020depth}, or to the fact that $\alpha_{\text{BALD}}$ can be computed exactly for DUNs.

\section{Discussion}
\vspace{-1mm}
Our preliminary investigation into the application of DUNs to active learning has lead to some promising results. DUNs do, as hypothesised, change their complexity as the dataset grows. As a result of this property, DUNs not only exhibit improved performance---on 10 of the 15 datasets analysed---but also overfit to a lesser extent relative to alternative BNN methods. Overfitting tends to be lower for DUNs only for the datasets on which DUNs also outperform MCDO, suggesting that robustness to overfitting is an important factor in DUNs' superior performance.  Interestingly, DUNs tend to outperform MCDO more as the number of actively sampled points increases. Intuitively, reduced overfitting should have the greatest impact for the smallest training set sizes. This may be explained by DUNs having better quality uncertainty estimates \citep{antoran2020depth}, which are more important later in active learning when it is more difficult to differentiate between the informativeness of candidate data points. Alternatively, the extra depth of the DUNs may enable better performance as the dataset becomes more complex.
Or, perhaps the negative effects of overfitting early in the active selection process accumulate and have a bigger influence later on.

We suggest two avenues for further investigation. Firstly, the observations made in this work should be confirmed using more complex datasets (preliminary experiments were also conducted on MNIST, with results shown in \cref{fig:app_res_mnist}). We also conducted investigations into the effect of using decaying prior probabilities over depth in place of a uniform prior, but found that this has almost no impact on performance (\cref{fig:res_reg_prior_decay_nll}). Further work could investigate the development of an improved DUN in which the prior over depth plays a larger role in controlling model complexity.


\clearpage
\section*{Acknowledgments}
JA acknowledges support from Microsoft Research, through its PhD Scholarship Programme, and from the EPSRC.
JUA acknowledges funding from the EPSRC and the Michael E. Fisher Studentship in Machine Learning.
This work has been performed using resources provided by the Cambridge Tier-2 system operated by the University of Cambridge Research Computing Service (http://www.hpc.cam.ac.uk) funded by EPSRC Tier-2 capital grant EP/P020259/1.

\bibliographystyle{plainnat}
\bibliography{references}

\clearpage
\appendix
\renewcommand\thefigure{\thesection.\arabic{figure}}  
\renewcommand\thetable{\thesection.\arabic{table}} 

\section{Depth uncertainty networks}\label{app:duns}
\setcounter{figure}{0} 
BNNs translate uncertainty in weight-space into predictive uncertainty by marginalising out the posterior over weights. The intractability of this posterior necessitates the use of approximate inference methods. DUNs, in contrast, leverage uncertainty about the appropriate \textit{depth} of the network in order to obtain predictive uncertainty estimates. There are two primary advantages of this approach: 1) the posterior over depth is tractable, mitigating the need for limiting approximations; and 2) due to the sequential structure of feed-forward neural networks, both inference and prediction can be performed with a single forward pass, making DUNs suitable for deployment in low-resource environments \citep{antoran2020depth}. 

\begin{figure}[h!]
\centering
    \begin{tikzpicture}[every node/.style={scale=0.85}]
    	\node[block=depth1] (RB0) {$f_{0}$};
    	\node[block=depth2, below right=.0ex and 4ex of RB0] (RB1) {$f_{1}$};
    	\node[block=depth3, below right=.0ex and 4ex of RB1] (RB2) {$f_{2}$};
    	\node[block=depth4, below right=.0ex and 4ex of RB2] (RB3) {$f_{3}$};
    	\node[block=depth5, below right=.0ex and 4ex of RB3] (RB4) {$f_{4}$};
    	\node[block=teal!80!white, below right=2ex and 7ex of RB4, text width=2.5ex] (RBD) {$f_{D}$};

    	\node[left=4ex of RB0] (X) {$\mathbf{x}$};

    	\draw[arrow] (X) -- (RB0);
    	\draw[arrow] (RB0) |- (RB1);
    	\draw[arrow] (RB1) |- (RB2);
    	\draw[arrow] (RB2) |- (RB3);
    	\draw[arrow] (RB3) |- (RB4);
    	\draw[line] (RB4) -- ([yshift=-.5ex] RB4.south);
    	\draw[line, dashed] ([yshift=-.5ex] RB4.south) |- ([xshift=-4.ex] RBD.west);
    	\draw[arrow] ([xshift=-4.ex] RBD.west) -- (RBD);
    
        \draw[arrow, snake] (RBD) -- ([xshift=4ex] RBD.east) node[coordinate] (end) {};
        \draw[arrow, snake] (RB0) -- (RB0 -| end);
        \draw[arrow, snake] (RB1) -- (RB1 -| end);
        \draw[arrow, snake] (RB2) -- (RB2 -| end);
        \draw[arrow, snake] (RB3) -- (RB3 -| end);
        \draw[arrow, snake] (RB4) -- (RB4 -| end);
        
        \draw [decorate,decoration={brace,amplitude=1ex, raise=.5ex}, xshift=2ex, thick] (RB0 -| end.west) -- (end) node [block=blue!60!white, midway, xshift=6ex, text width=5ex, thin] (OB) {$f_{D+1}$};
        
    	\node[right=4ex of OB] (YD) {$\mathbf{\hat y}_{i}$};
    	\draw[arrow] (OB.east) -- (YD);
    \end{tikzpicture}
\caption[DUN computational model.]{DUN computational model \citep{antoran2020depth}. Each layer's activations are passed through the output block, producing per-depth predictions.}\label{fig:dun_comp_model}
\end{figure}
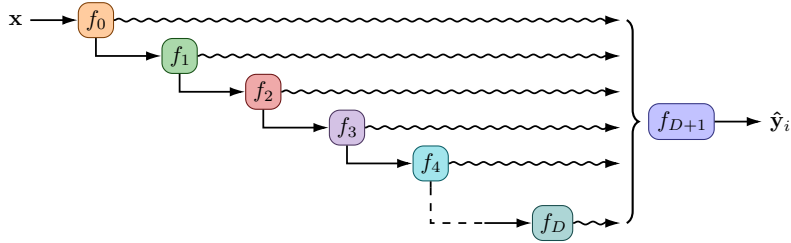

DUNs are composed of subnetworks of increasing depth, with each subnetwork contributing one prediction to the final model. The computational model for DUNs is shown in \cref{fig:dun_comp_model}. Inputs are passed through an input block, $f_0(\cdot)$, and then sequentially through each of $D$ intermediate blocks $\{f_i(\cdot)\}_{i=1}^{D}$, with the activations of the previous layer acting as the inputs of the current layer. The activations of each of the $D+1$ layers are passed through the output block $f_{D+1}(\cdot)$ to generate per-depth predictions. These are combined via marginalisation over the depth posterior, equivalent to forming an ensemble with networks of increasing depth. Variation between the predictions from each depth can be used to obtain predictive uncertainty estimates, in much the same way that variance in the predictions of different sampled weight configurations yield predictive uncertainties in BNNs.

\subsection{Uncertainty over depth}

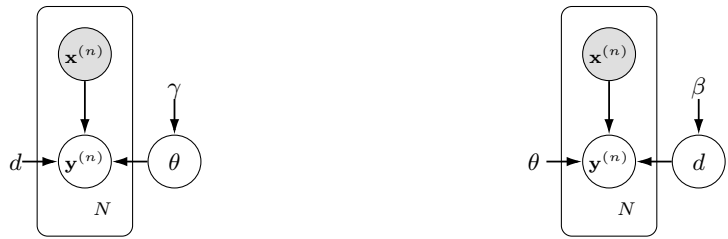
\begin{figure}[h]
\centering
\begin{subfigure}[b]{0.49\textwidth}
    \centering
    \begin{tikzpicture}[every node/.style={scale=1}]
      \node[obs] (xn) {\scriptsize $\mathbf{x}^{(n)}$};
      \node[latent, below=4.5ex of xn] (yn) {\scriptsize $\mathbf{y}^{(n)}$};
      \node[latent, right=3ex of yn] (theta) {\footnotesize $\theta$};
      \node[const, above=3ex of theta] (gamma) {\footnotesize $\gamma$};
      \node[const, left=3ex of yn] (d) {\footnotesize $d$};
      
      \edge[arrow] {xn} {yn} ; %
      \edge[arrow] {theta} {yn} ;
      \edge[arrow] {gamma} {theta} ;
      \edge[arrow] {d} {yn}
      
      \plate[inner sep=1.7ex] {} {(yn)(xn)} {\scriptsize $N$} ;
    \end{tikzpicture}
\end{subfigure}
\begin{subfigure}[b]{0.49\textwidth}
    \centering
    \begin{tikzpicture}[every node/.style={scale=1}]
      \node[obs] (xn) {\scriptsize $\mathbf{x}^{(n)}$};
      \node[latent, below=4.5ex of xn] (yn) {\scriptsize $\mathbf{y}^{(n)}$};
      \node[const, left=3ex of yn] (theta) {\footnotesize $\theta \ $};
      \node[latent, right=3ex of yn] (d) {\footnotesize $d$};
      \node[const, above=3ex of d] (beta) {\footnotesize $\beta$};
      
      \edge[arrow] {xn} {yn} ; %
      \edge[arrow] {theta} {yn} ;
      \edge[arrow] {d} {yn} ;
      \edge[arrow] {beta} {d} ;
      
      \plate[inner sep=1.7ex] {} {(yn)(xn)} {\scriptsize $N$} ;
    \end{tikzpicture}
\end{subfigure}
\caption[Graphical models of BNNs and DUNs.]{Left: graphical model of a BNN. Right: graphical model of a DUN.}
\label{fig:dun_graph_model}
\end{figure}

\Cref{fig:dun_graph_model} compares the graphical model representations of a BNN and a DUN. In BNNs, the weights $\theta$ are treated as random variables drawn from a distribution $p_{\gamma}(\theta)$ with hyperparameters $\gamma$. In DUNs, the depth of the network $d$ is assumed to be stochastic, while the weights are learned as hyperparameters. A categorical prior distribution is assumed for $d$, with hyperparameters $\beta$: $p_{\beta}(d)=\operatorname{Cat}(d \mid \{\beta_i \}_{i=0}^D)$. The model's marginal log likelihood (MLL) is given by
\begin{equation} 
\log p(\mathcal{D}_{\text{train}} ; \theta)=\log \sum_{i=0}^{D}\left(p_{\beta}(d=i) \prod_{n=1}^{N} p\left(\mathbf{y}_n \mid \mathbf{x}_n, d=i ; \theta\right)\right),\label{eq:dun_mll}
\end{equation}
where the likelihood for each depth is parameterised by the corresponding subnetwork's output: $p(\mathbf{y} \mid \mathbf{x}, d=i ; \theta)=p\left(\mathbf{y} \mid f_{D+1}\left(\mathbf{a}_{i} ; \theta\right)\right)$, where  $\mathbf{a}_{i}=f_i(\mathbf{a}_{i-1})$. \\

The posterior, 
\begin{equation}
    p(d \mid \mathcal{D}_{\text{train}} ; \theta)=\frac{p(\mathcal{D}_{\text{train}} \mid d ; \theta) p_{\beta}(d)}{p(\mathcal{D}_{\text{train}} ; \theta)}, \label{eq:depth_post}
\end{equation}

is also a categorical distribution, whose probabilities reflect how well each depth subnetwork explains the data. DUNs leverage two properties of modern neural networks: first, that successive layers have been shown to extract features at increasing levels of abstraction \citep{zeiler2014visualizing}; and second, that networks are typically underspecified, meaning that several different models may explain the data well \citep{d2020underspecification}. The first property implies that initial layers in a network specialise on low-level feature extraction, yielding poor predictions from the shallower subnetworks in a DUN. This is handled by the depth posterior assigning low probabilities to earlier layers, in preference for later layers that specialise on prediction. The second property suggests that subnetworks at several depths are able to explain the data simultaneously and thus have non-zero posterior probabilities, yielding the diversity in predictions required to obtain useful estimates of model uncertainty. 

\subsection{Inference in DUNs}

\citet{antoran2020depth} find that directly optimising the MLL \cref{eq:dun_mll} with respect to $\theta$ results in convergence to a local optimum in which all but one layer is attributed a posterior probability of zero. That is, direct optimisation returns a deterministic network of arbitrary depth. The authors instead propose a stochastic gradient variational inference (VI) approach with the aim of separating optimisation of the weights $\theta$ from the posterior. A surrogate categorical distribution $q_{\phi}(d)$ is introduced as the variational posterior. The following evidence lower bound (ELBO) can be derived:
\begin{equation} 
    \mathcal{L}(\phi, \theta)=\sum_{n=1}^{N} \mathbb{E}_{q_{\phi}(d)}\left[\log p\left(\mathbf{y}_n \mid \mathbf{x}_n, d ; \theta\right)\right]-\mathrm{KL}\left(q_{\phi}(d) \| p_{\beta}(d)\right) ,  \label{eq:dun_elbo}
\end{equation}
allowing $\theta$ and the variational parameters $\phi$ to be optimised simultaneously. \citet{antoran2020depth} show that under the VI approach, the variational parameters converge more slowly than the weights, enabling the weights to reach a setting at which a positive posterior probability is learnt for several depths.

It is important to note that \cref{eq:dun_elbo} can be computed exactly, and that optimising $\mathcal{L}(\phi, \theta)$ is equivalent to exact inference of the true posterior $p(d \mid \mathcal{D}_{\text{train}}; \theta)$ in the limit: since both $q$ and $p$ are categorical, \cref{eq:dun_elbo} is convex and tight at the optimum (i.e., $q_{\phi}(d)=p(d \mid \mathcal{D}_{\text{train}}; \theta)$). Since the expectation term can be computed exactly using the activations at each layer (recall $p(\mathbf{y} \mid \mathbf{x}, d=i ; \theta)=p\left(\mathbf{y} \mid f_{D+1}\left(\mathbf{a}_{i} ; \theta\right)\right)$), the ELBO can be evaluated exactly without Monte Carlo sampling.

\clearpage
\section{Bias in active learning} \label{app:al_bias}
\setcounter{figure}{0} 

We use notation similar to \citet{farquhar_statistical_2020}. In the active learning setting, given a loss function $\mathcal{L}( \mathbf{y}, f_{\theta}(\mathbf{x}))$, we aim to find $\theta$ that minimises the population risk over $p_{\text{data}}(\mathbf{y}, \mathbf{x})$: ${\mathbb{E}_{\mathbf{x},  \mathbf{y} \sim p_{\text {data }}}\left[\mathcal{L}\left(\mathbf{y}, f_{\theta}(\mathbf{x})\right)\right]}$. In practice, we only have access to $N$ samples from $p_{\text{data}}$. These yield the empirical risk, ${r=\frac{1}{N} \sum_{n=1}^{N} \mathcal{L}\left(\mathbf{y}_{n}, f_{\theta}\left(\mathbf{x}_{n}\right)\right)}$. $r$ is an unbiased estimator of the population risk when the data are drawn i.i.d. from $p_{\text{data}}$.
If these samples have been used to train $\theta$, our estimators become biased. We refer to estimators biased by computing with the training data with capital letters (i.e., $R$).
In the active learning setting, our model is optimised using a subset of $M$ actively sampled data points:
\begin{equation}
    \tilde{R}=\frac{1}{M} \sum_{m=1}^{M} \mathcal{L}\left(\mathbf{y}_{i_m}, f_{\theta}\left(\mathbf{x}_{i_m}\right)\right), \quad i_m \sim \alpha\left(i_{1: m-1}, \mathcal{D}_{\text {pool }}\right) . \label{eq:r_tilde}
\end{equation} 
The proposal distribution $\alpha\left(i_{m} ; i_{1: m-1}, \mathcal{D}_{\text {pool }}\right)$ over the pool of unlabelled data $\mathcal{D}_{\text{pool}}$ represents the probability of each index being sampled next, given that we have already acquired $\mathcal{D}_{\text{train}} = \{\mathbf{x}_i\}_{1:m-1}$. As a result of active sampling, $\tilde{R}$ is not an unbiased estimator of $R$. \citet{farquhar_statistical_2020} propose $\tilde{R}_{\text{LURE}}$ (``levelled unbiased risk estimator''), which removes active learning bias:
\begin{equation}
    \begin{aligned}
    \tilde{R}_{\text{LURE}} \!\equiv\! \frac{1}{M} \sum_{m=1}^{M} v_{m} \mathcal{L}_{i_{m}} ;
    \quad v_{m} \!\equiv 1+\frac{N-M}{N-m}\left(\frac{1}{(N-m+1)\ \alpha\left(i_{m} ; i_{1: m-1}, \mathcal{D}_{\text{pool}}\right)}-1\right), \label{eq:r_lure} 
    \end{aligned}
\end{equation}
where $\mathcal{L}_{i_{m}} \equiv \mathcal{L}\left(\mathbf{y}_{i_{m}}, f_{\theta}\left(\mathbf{x}_{i_{m}}\right)\right)$.  Intuitively, the estimator works by re-weighting each example's contribution to the total loss by its inverse acquisition probability, such that if unusual examples have particularly high acquisition probabilities, they contribute proportionally less to the total risk. A detailed derivation of this estimator is given by \citet{farquhar_statistical_2020}.

Surprisingly, \citet{farquhar_statistical_2020} find that using $\tilde{R}_{\text{LURE}}$ during training can negatively impact the performance of flexible models. They explain this by noting that the overfitting bias ($r{-}R$) typical in such models has the opposite sign to the bias induced by active learning. That is, active learning encourages the selection of ``difficult'' to predict data points, increasing the overall risk, whereas overfitting results in the total risk being underestimated. Where overfitting is substantial, it is (partially) offset by active learning bias, such that correcting for active learning bias does not improve model performance.

In this paper, however, we do not attempt to use $\tilde{R}_{\text{LURE}}$ to improve the performance of active learning, but rather as a means to measure the degree of overfitting for DUNs and MCDO. The procedure is as follows:
\begin{enumerate}
    \item We train a model on the current active learning train set.
    \item We calculate the test-set empirical risk $r$ of the model.
    \item We calculate the unbiased current train-set empirical risk $\tilde{R}_{\text{LURE}}$ of the model. This quantity is free from active learning bias.
    \item We estimate the degree of overfitting as the difference between these two values of the empirical risk. Intuitively, once active learning bias has been accounted for, the difference between the risk estimated on the train and test sets must be due to overfitting.
\end{enumerate}

\clearpage
\section{Experimental setup} \label{app:exp_setup}
\setcounter{figure}{0} 

We conduct experiments on three groups of datasets: the toy regression datasets used in \citet{antoran2020depth}, nine of the UCI regression datasets \citep{hernandez2015probabilistic}, and the MNIST image classification dataset \citep{lecun1998mnist}. We implement batch-based active learning, with batches of $10$ (for MNIST and most toy datasets) or $20$ (for most UCI datasets) data points acquired in each query, depending on the size of the dataset. An initial training set representing a small proportion of the full data is selected uniformly at random in the first query. For all regression datasets, 80\% of the data are used for training, 10\% for validation and 10\% for testing. The standard train-test split is used for MNIST. Details about dataset sizes, input dimensionality, and active learning specifications for each dataset are provided in \cref{tab:datasets}.
\begin{table}[h]
    \centering
    \caption[Summary of datasets and active learning specifications.]{Summary of datasets and active learning specifications. For the toy and UCI datasets, 80\% of the data are used for training, 10\% for validation and 10\% for testing. For MNIST, the test and train set sizes are shown in parentheses, i.e., (train \& test).}
    \resizebox{\textwidth}{!}{\begin{tabular}{l|ccccc}
    \toprule
         \textsc{Name} & \textsc{Size} & \textsc{Input Dim.} & \textsc{Init. train size} & \textsc{No. queries} & \textsc{Query size} \\
         \midrule
         Simple 1d & 501 & 1 & 10 & 30 & 10 \\
         \citet{izmailov2020subspace} & 400 & 1 & 10 & 30 & 10 \\
         \citet{foong2019between} & 100 & 1 & 10 & 15 & 5 \\
         Matern & 400 & 1 & 10 & 30 & 10 \\
         Wiggle & 300 & 1 & 10 & 20 & 10 \\
         \midrule
         Boston Housing & 506 & 13 & 20 & 17 & 20 \\
         Concrete Strength & 1,030 & 8 & 50 & 30 & 20 \\
         Energy Efficiency & 768 & 8 & 50 & 30 & 20 \\
         Kin8nm & 8,192 & 8 & 50 & 30 & 20 \\
         Naval Propulsion & 11,934 & 16 & 50 & 30 & 20 \\
         Power Plant & 9,568 & 4 & 50 & 30 & 20 \\ 
         Protein Structure & 45,730 & 9 & 50 & 30 & 20 \\
         Wine Quality Red & 1,599 & 11 & 50 & 30 & 20 \\
         Yacht Hydrodynamics & 308 & 6 & 20 & 20 & 10 \\
         \midrule
         MNIST & 70,000 (60,000 \& 10,000) & 784 ($28\,\times\,28$) & 10 & 10 & 10 \\
         \bottomrule
    \end{tabular}}
    \label{tab:datasets}
\end{table}

For the regression problems, we implement a fully-connected network with residual connections, with 100 hidden nodes per layer. The networks contain 10 hidden layers for DUNs, or three hidden layers for MCDO and MFVI. Other depths were tested for the baseline methods, with similar results to the chosen depth of three layers (see \cref{fig:app_depths} for these results). We use ReLU activations, and for DUNs batch normalisation is applied after every layer \citep{ioffe2015batch}. Optimisation is performed over $1,000$ iterations with full-batch gradient descent, with momentum of $0.9$ and a learning rate of $10^{-4}$. A weight decay value of $10^{-5}$ is also used. We do not implement early stopping, but the best model based on evaluation of the evidence lower bound on the validation set is used for reporting final results. MFVI models are trained using five MC samples and the local reparameterisation trick \citep{kingma2015variational}, and prediction for both MCDO and MFVI is based on 10 MC samples. For MCDO models a fixed dropout probability of $0.1$ is used. Unless otherwise specified, DUNs use a uniform categorical prior over depth, while MFVI  networks use a $\mathcal{N}(\mathbf{0}, I)$ prior over weights. These hyperparameter settings largely follow those used in \citet{antoran2020depth} for the toy regression problems.

For image classification experiments we use the convolutional network architecture described in \citet{antoran2020variational}.  DUNs contain 10 intermediate blocks (bottleneck convolutional blocks described in \citet{he2016identity}), while MCDO and MFVI networks contain three. Following PyTorch defaults for MNIST, optimisation is performed over 90 epochs, with the learning rate reduced by a factor of 10 at epochs 40 and 70. The initial learning rate is 0.1. We do not use a validation set to find the best model, instead reporting results on the final model. All other hyperparameters are as for the regression case.

All experiments are repeated 40 times with different weight initialisations and train-test splits (with the exception of experiments on the Protein dataset, which are repeated only 30 times due to the cost of evaluation on the larger test set). Unless otherwise specified, we report the mean and standard deviation of the relevant metric over the repeated experiment runs.

\clearpage
\section{Additional results}\label{app:add_results}
\setcounter{figure}{0} 

\renewcommand\thefigure{\thesection.\arabic{figure}}
\subsection{Remaining UCI regression datasets}

\begin{figure}[h!] 
\centering
    \begin{subfigure}[b]{0.335\textwidth}
        \centering
        \includegraphics[width=\linewidth]{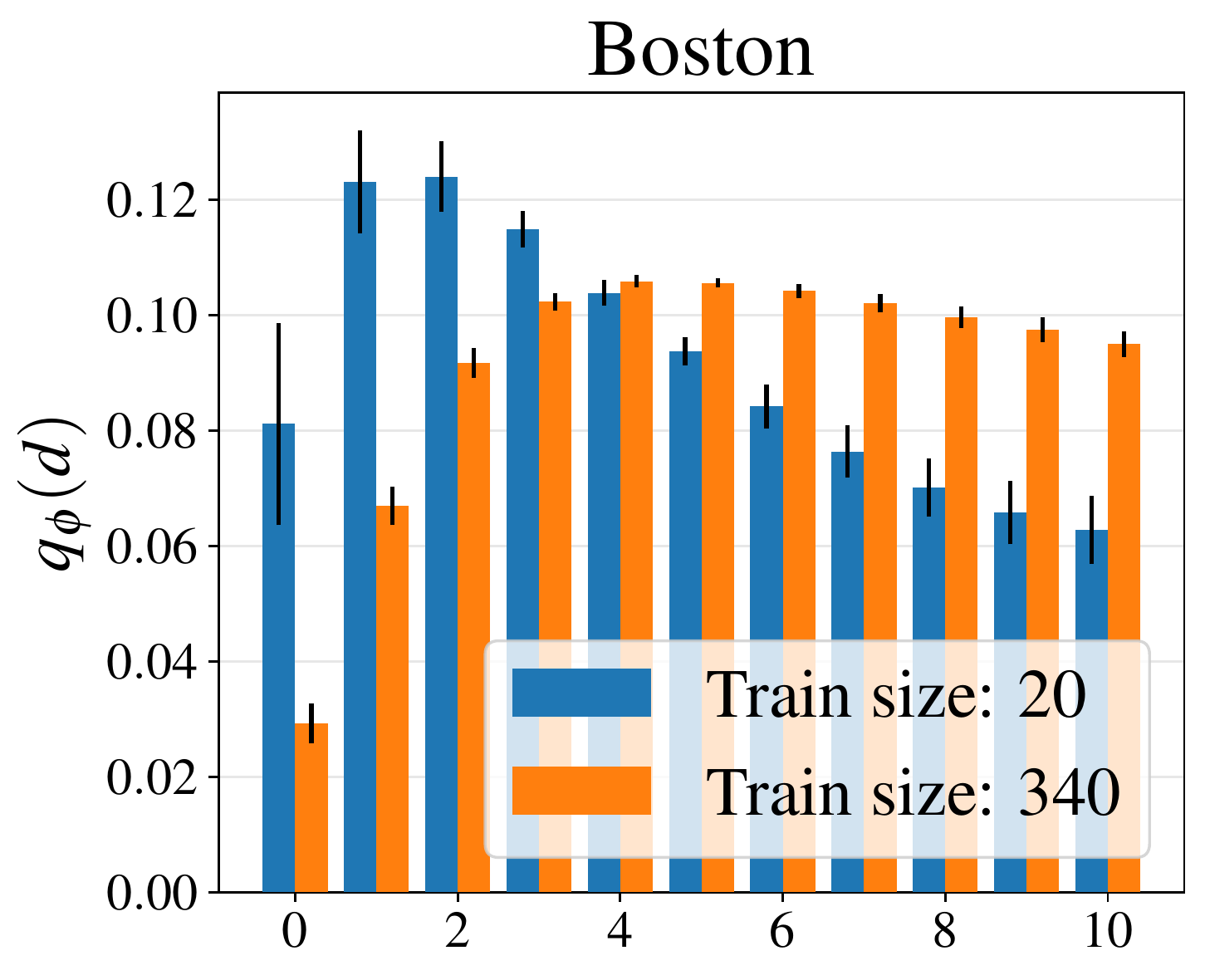}
    \end{subfigure} 
    \begin{subfigure}[b]{0.31\textwidth}
        \centering
        \includegraphics[width=\linewidth]{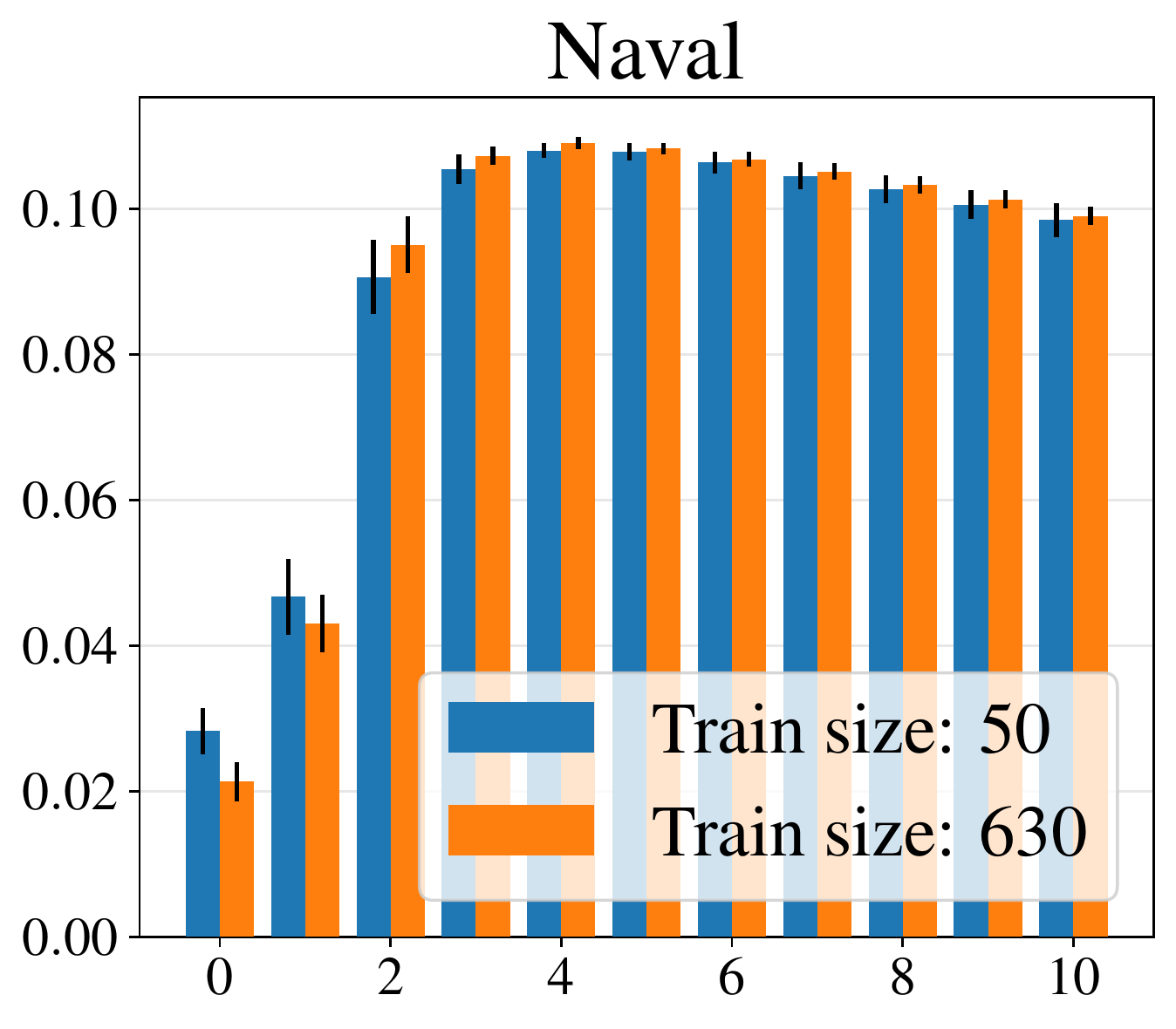}
    \end{subfigure} 
    \begin{subfigure}[b]{0.31\textwidth}
        \centering
        \includegraphics[width=\linewidth]{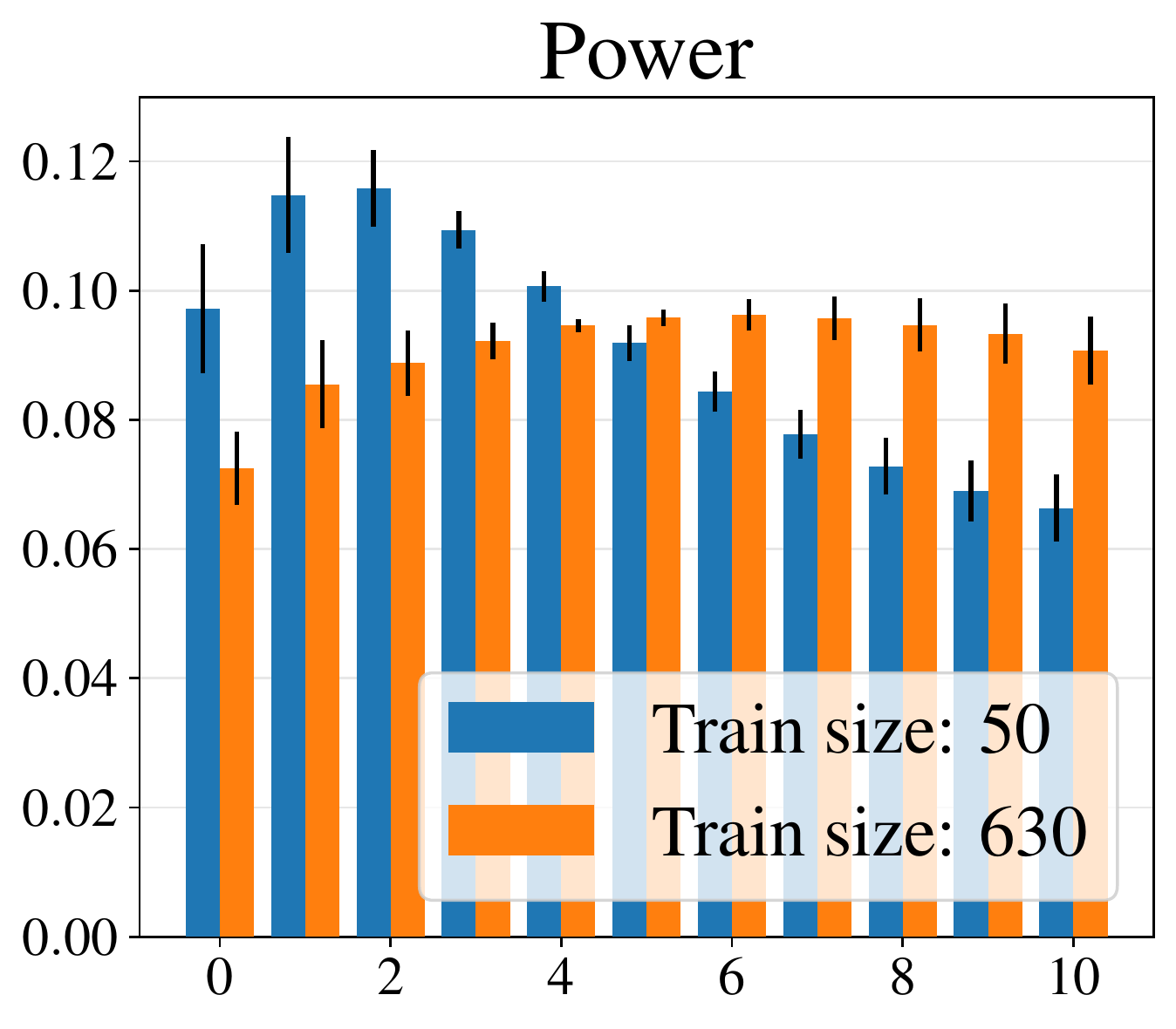}
    \end{subfigure} \\
    \begin{subfigure}[b]{0.34\textwidth}
        \centering
        \includegraphics[width=\linewidth]{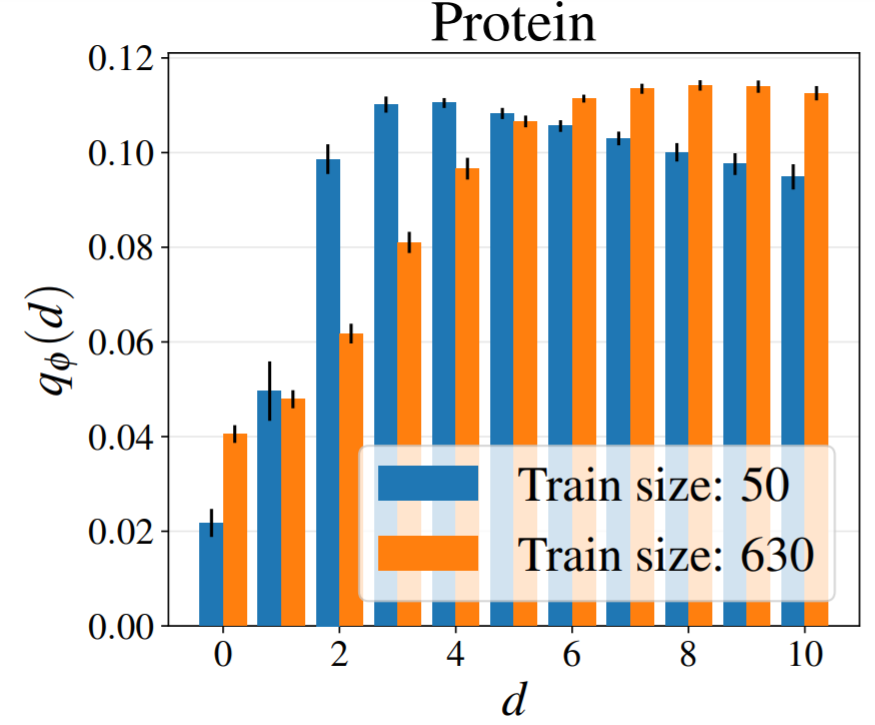}
    \end{subfigure} 
    \begin{subfigure}[b]{0.31\textwidth}
        \centering
        \includegraphics[width=\linewidth]{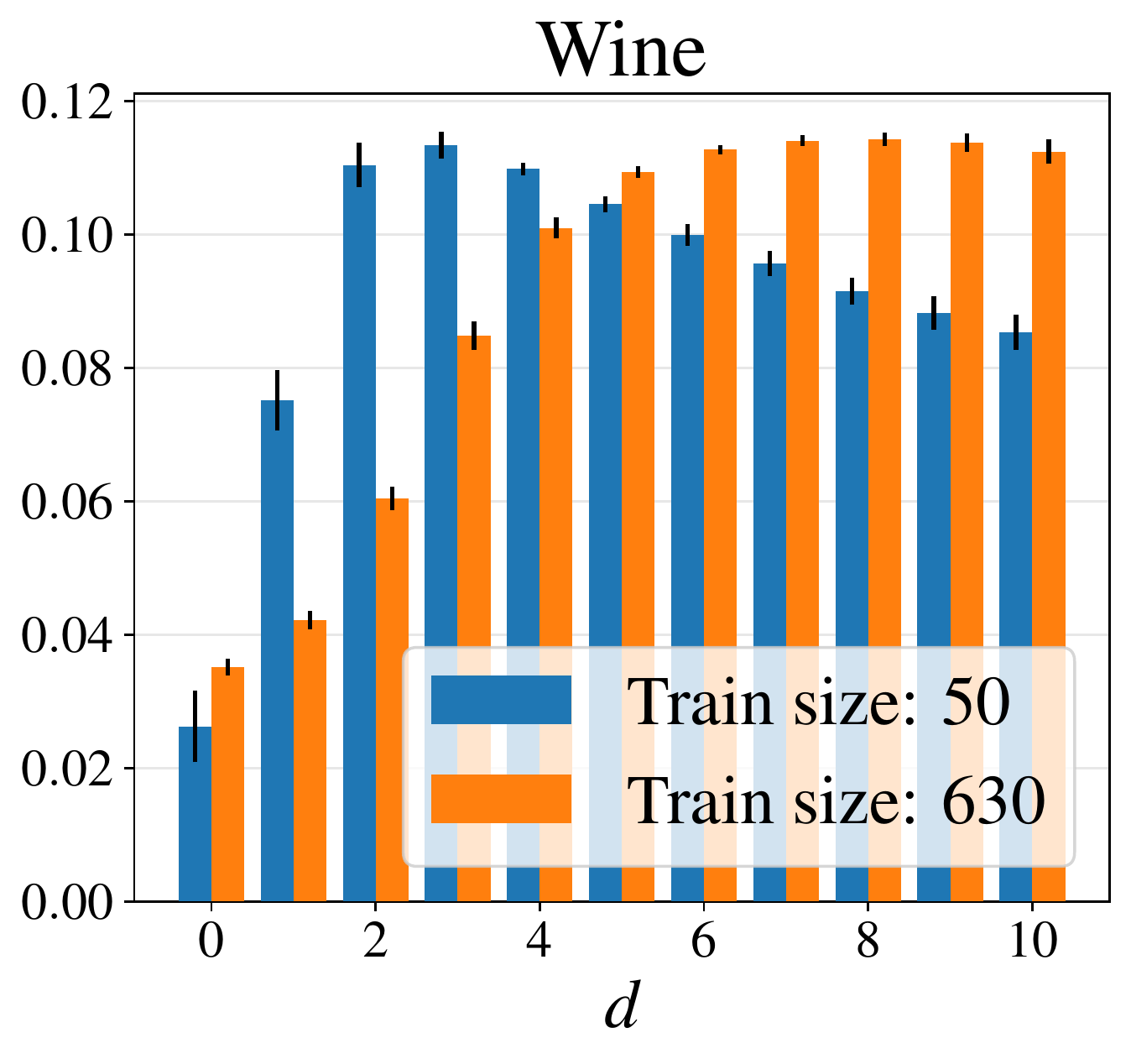}
    \end{subfigure} 
    \begin{subfigure}[b]{0.31\textwidth}
        \centering
        \includegraphics[width=\linewidth]{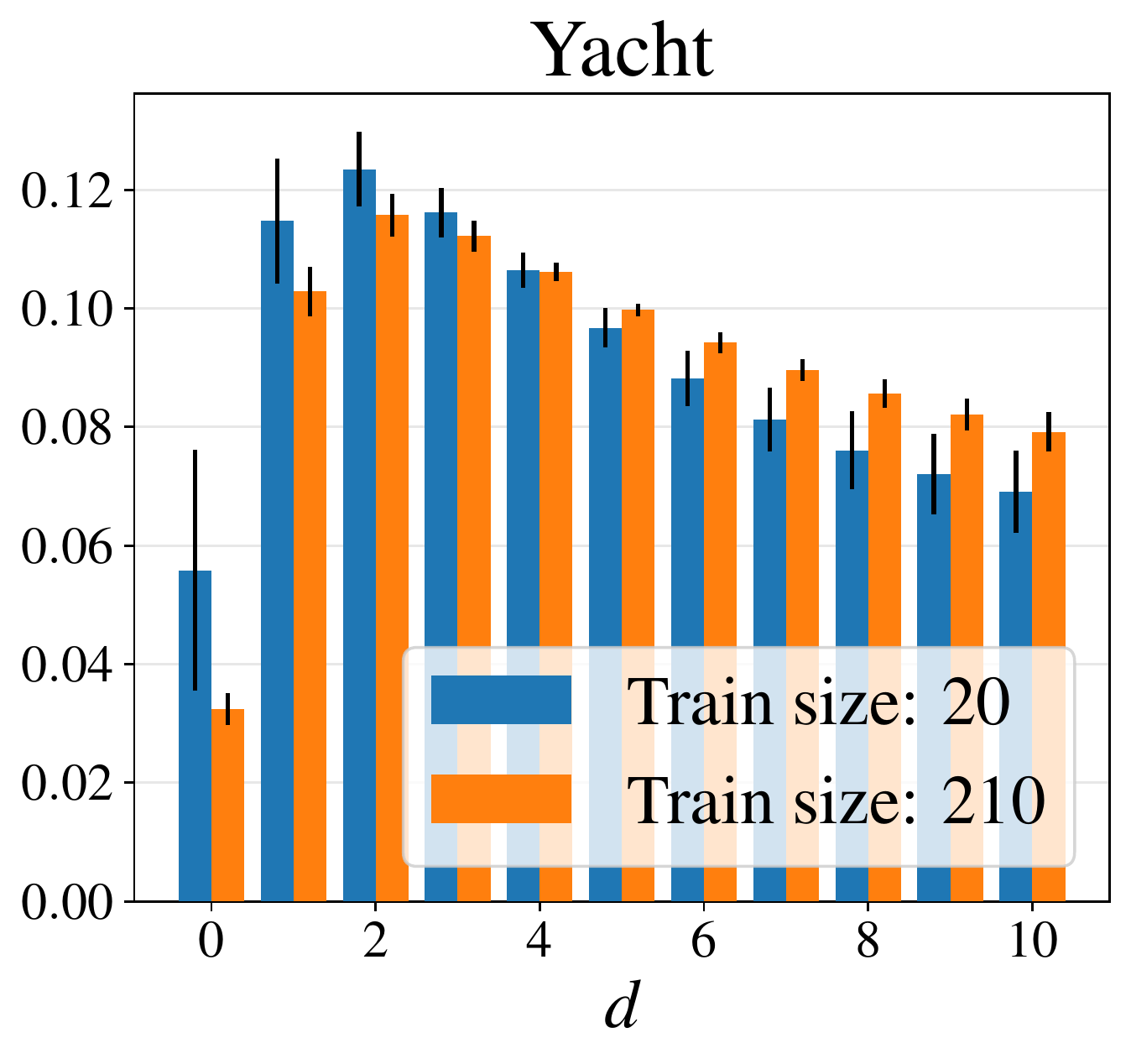}
    \end{subfigure} \\
    \caption{Posterior probabilities over depth for DUNs trained on UCI regression datasets, for the \textbf{\textcolor{mplblue}{smallest}} and \textbf{\textcolor{mplorange}{largest}} labelled datasets used in active learning. Extension of the results in \cref{fig:res_reg_posts}.}
    \label{fig:app_res_reg_posts_app}
\end{figure}

\begin{figure}[h] 
\centering
    \begin{subfigure}[b]{0.33\textwidth}
        \centering
        \includegraphics[width=\linewidth]{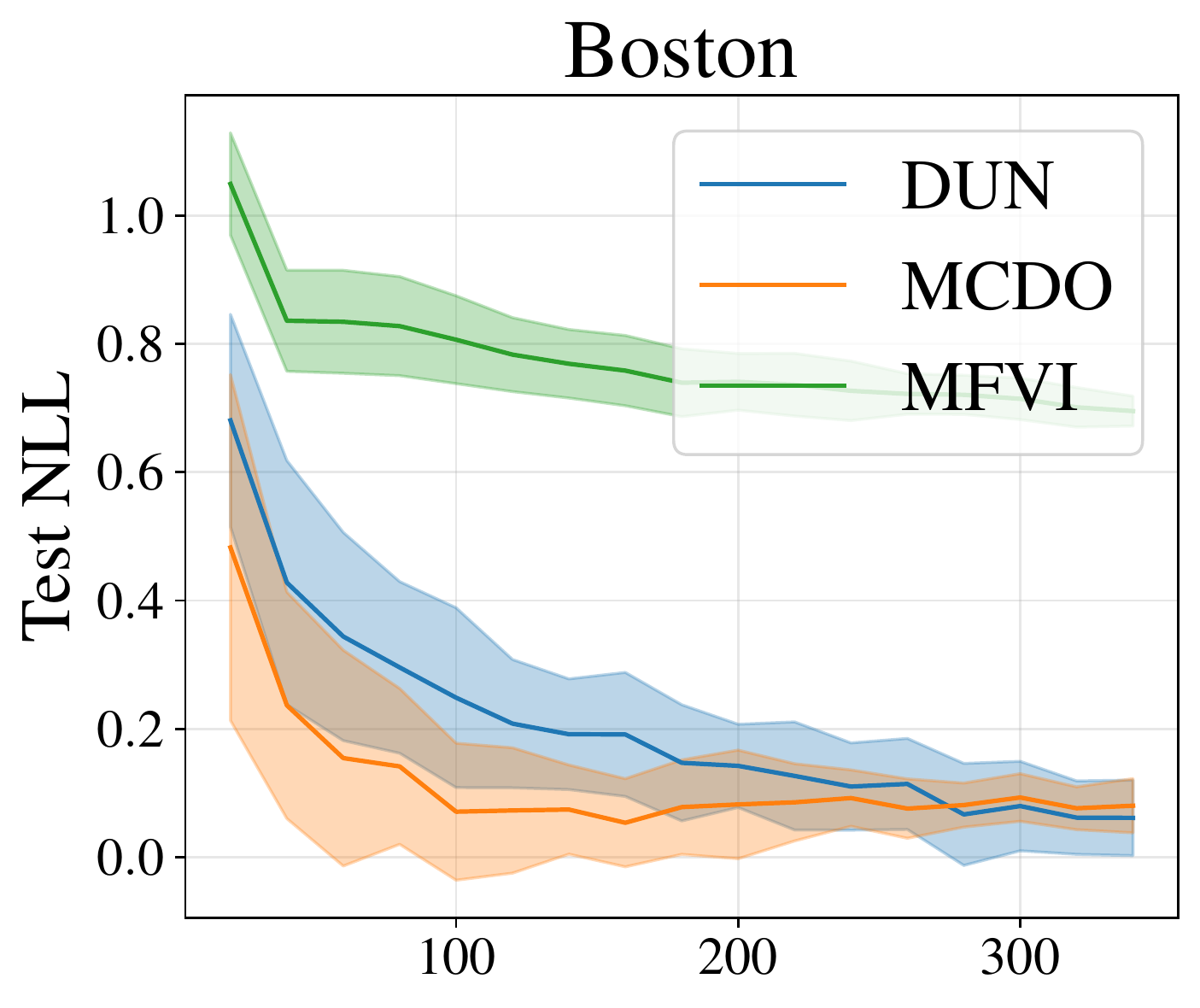}
    \end{subfigure} 
    \begin{subfigure}[b]{0.31\textwidth}
        \centering
        \includegraphics[width=\linewidth]{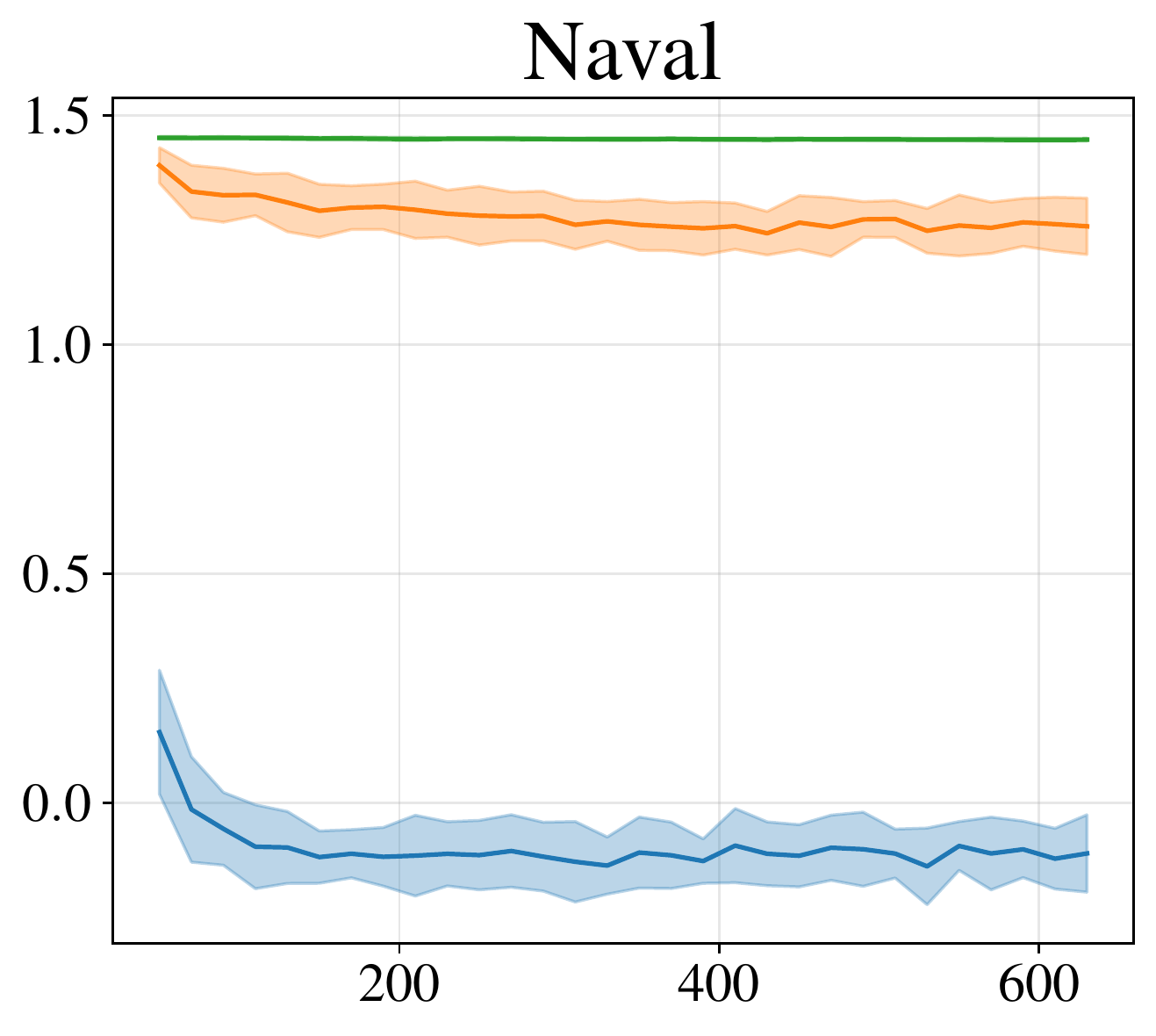}
    \end{subfigure} 
    \begin{subfigure}[b]{0.315\textwidth}
        \centering
        \includegraphics[width=\linewidth]{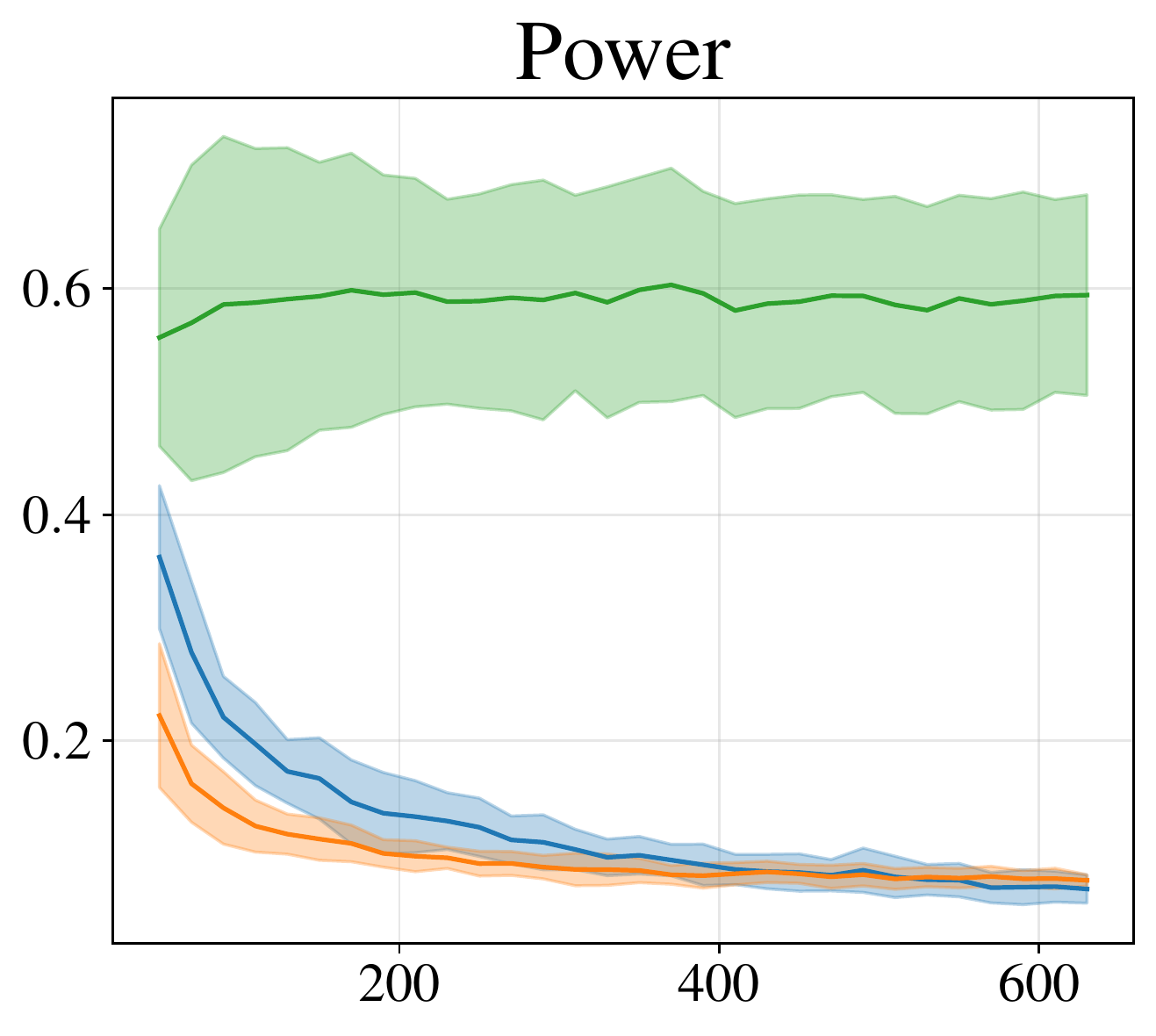}
    \end{subfigure} \\
    \begin{subfigure}[b]{0.33\textwidth}
        \centering
        \includegraphics[width=\linewidth]{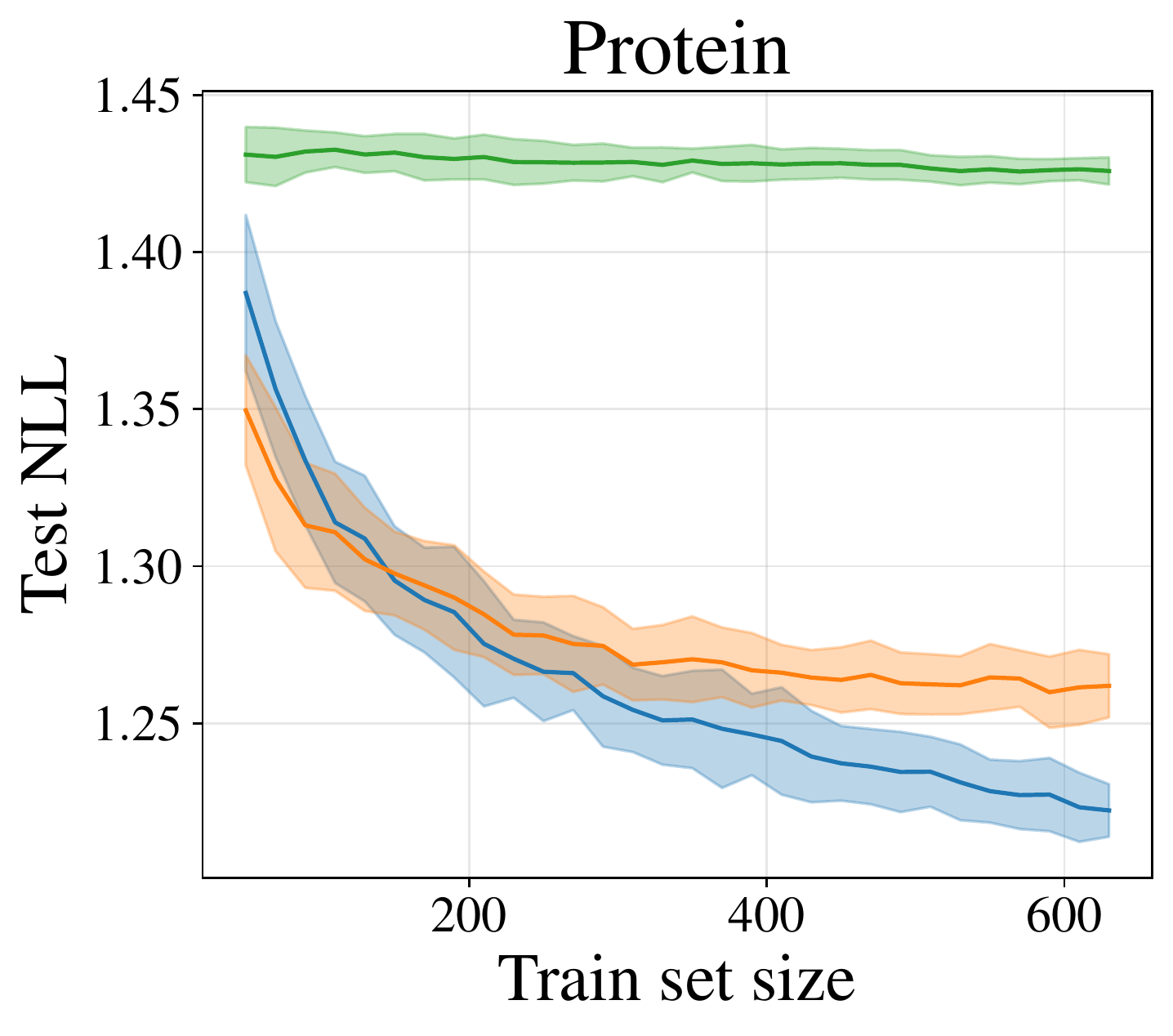}
    \end{subfigure} 
    \begin{subfigure}[b]{0.31\textwidth}
        \centering
        \includegraphics[width=\linewidth]{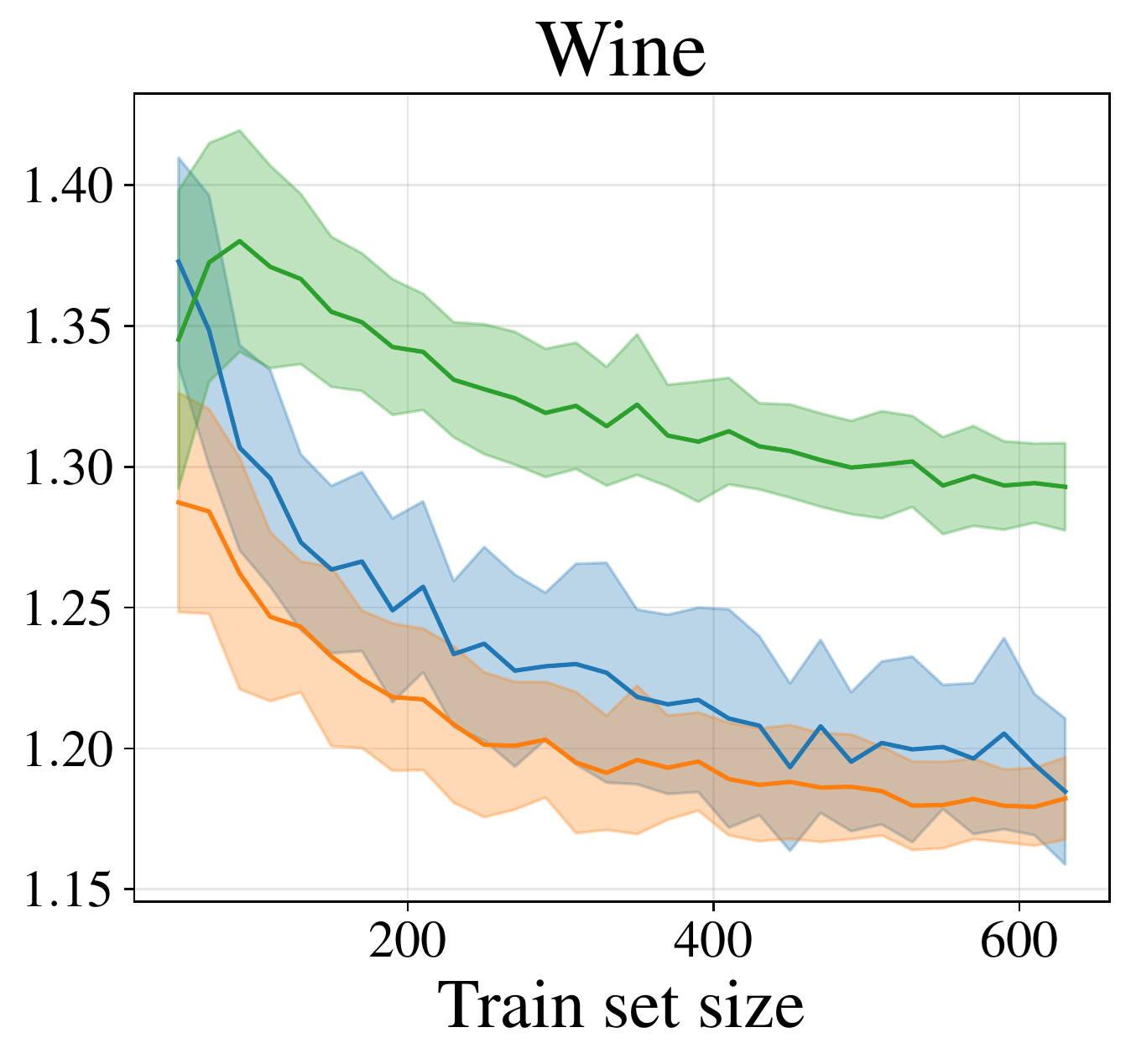}
    \end{subfigure} 
    \begin{subfigure}[b]{0.315\textwidth}
        \centering
        \includegraphics[width=\linewidth]{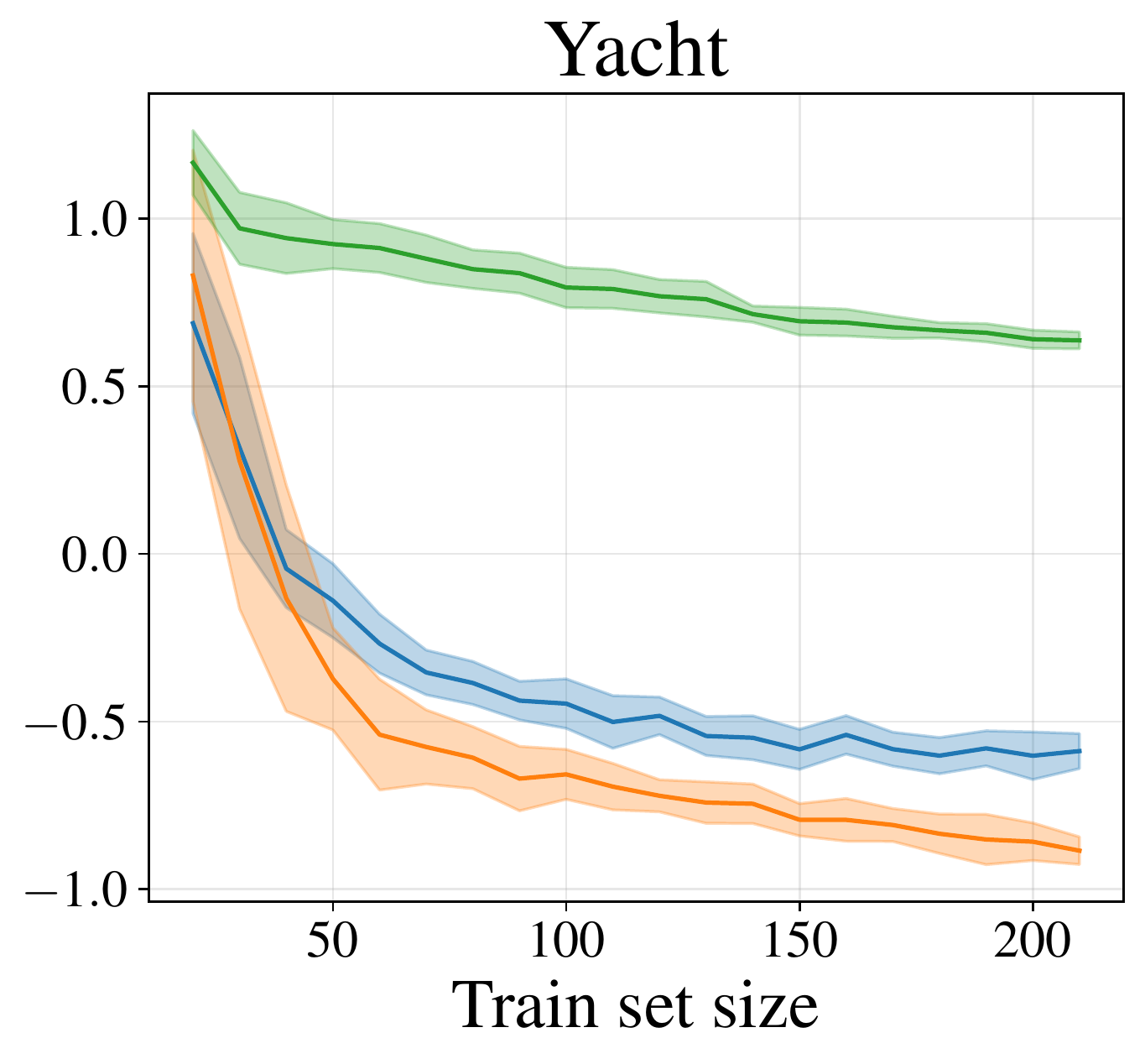}
    \end{subfigure} \\
    \caption[NLL vs. number of training points for \textbf{\textcolor{mplblue}{DUNs}}, \textbf{\textcolor{mplorange}{MCDO}} and \textbf{\textcolor{mplgreen}{MFVI}} evaluated on UCI datasets.]{Test NLL vs. number of training points evaluated on UCI datasets. \textbf{\textcolor{mplblue}{DUNs}}, \textbf{\textcolor{mplorange}{MCDO}} and \textbf{\textcolor{mplgreen}{MFVI}} are compared. Extension of the results in \cref{fig:res_reg_methods_nll}.}   \label{fig:app_res_reg_methods_nll}
\end{figure}

\begin{figure}[h] 
\centering
    \begin{subfigure}[b]{0.33\textwidth}
        \centering
        \includegraphics[width=\linewidth]{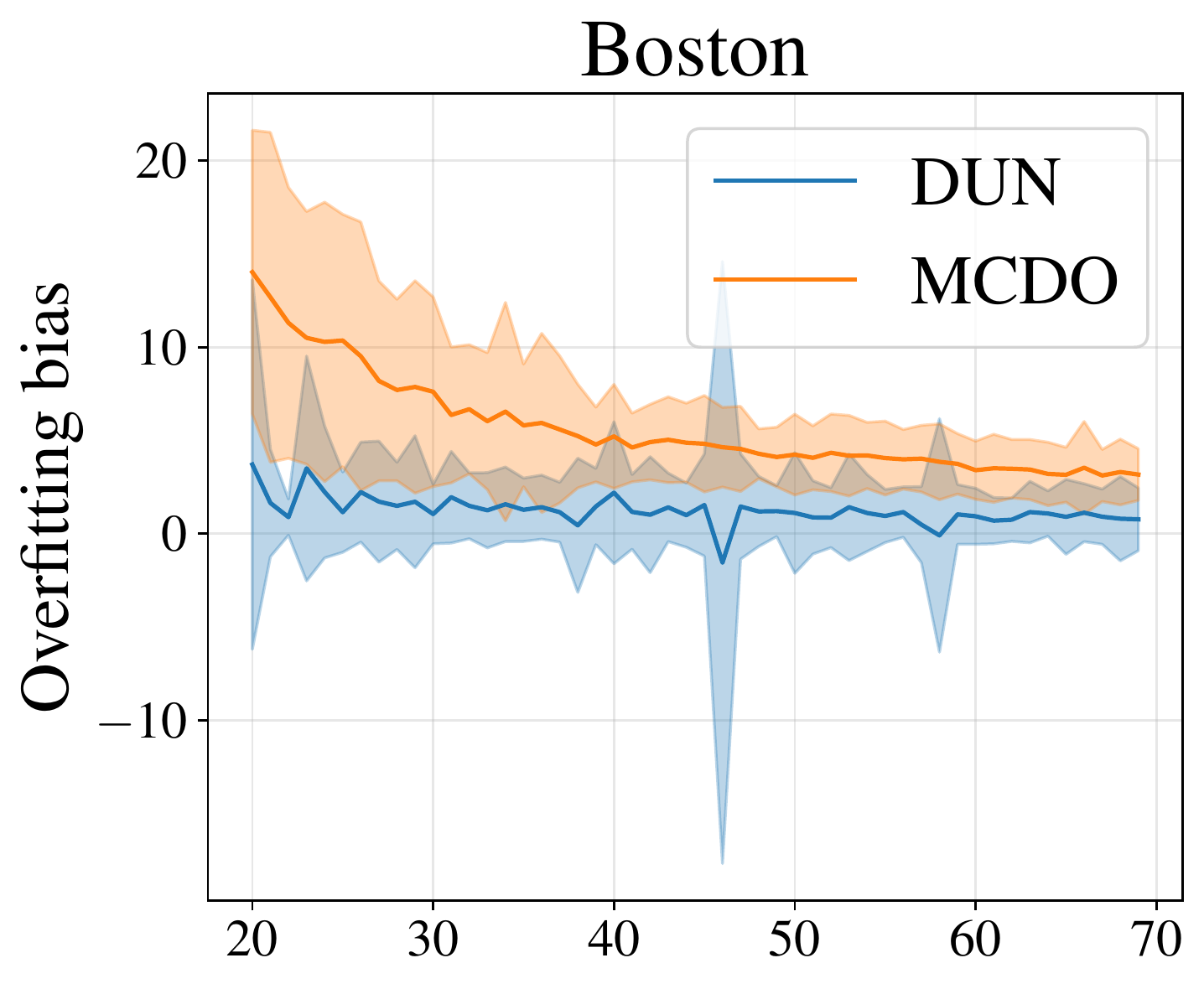}
    \end{subfigure} 
    \begin{subfigure}[b]{0.31\textwidth}
        \centering
        \includegraphics[width=\linewidth]{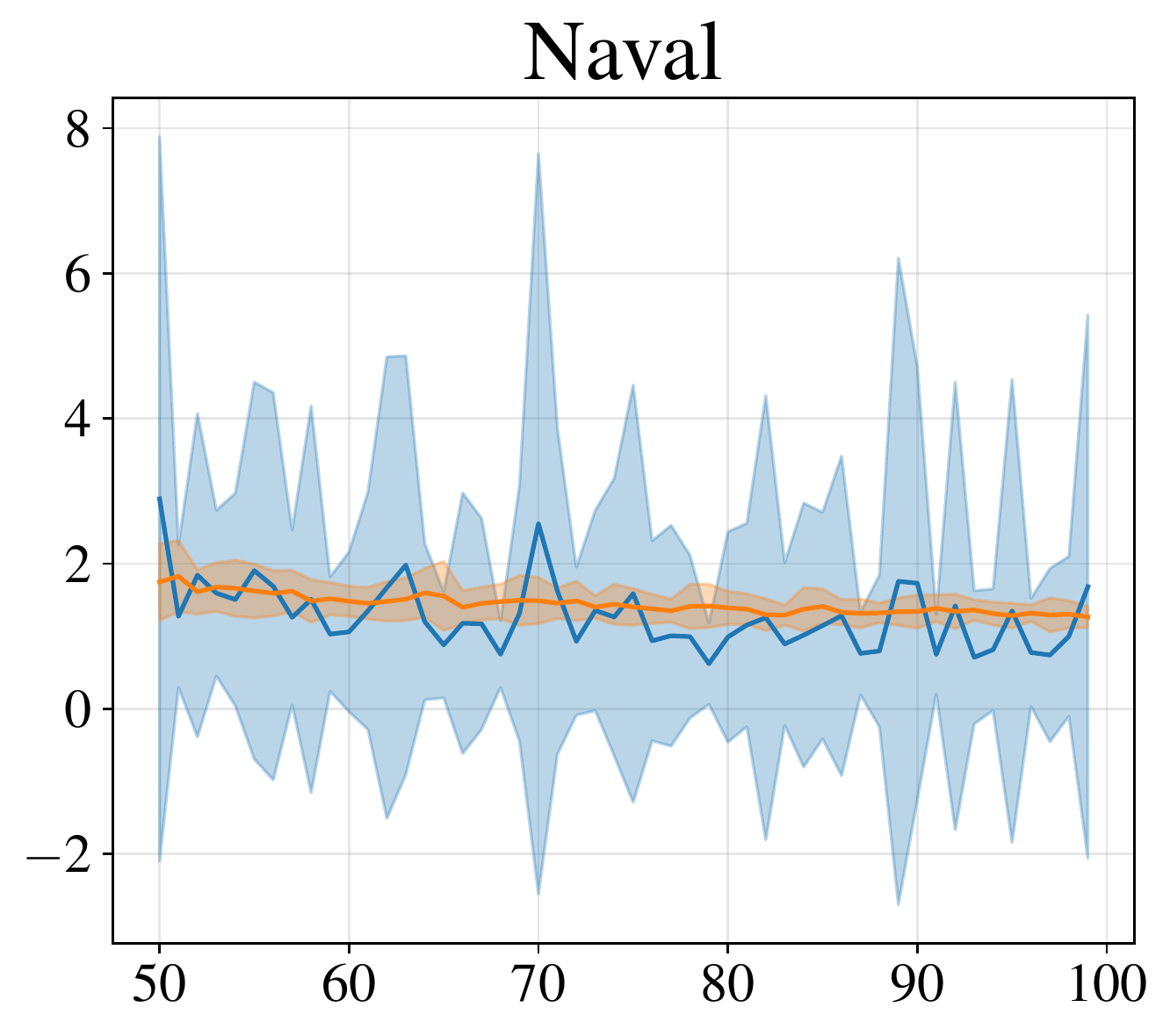}
    \end{subfigure} 
    \begin{subfigure}[b]{0.32\textwidth}
        \centering
        \includegraphics[width=\linewidth]{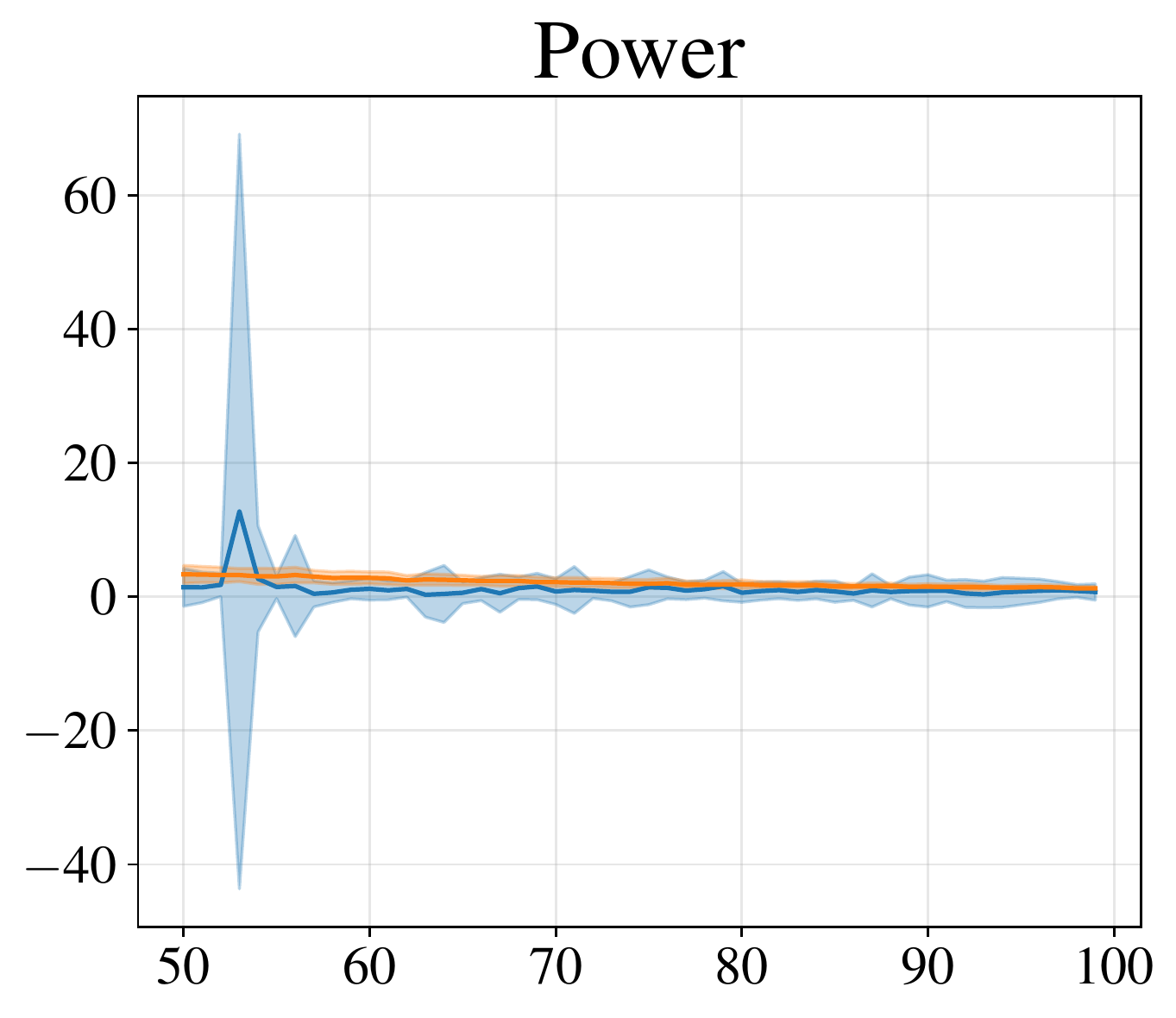}
    \end{subfigure} \\
    \begin{subfigure}[b]{0.335\textwidth}
        \centering
        \includegraphics[width=\linewidth]{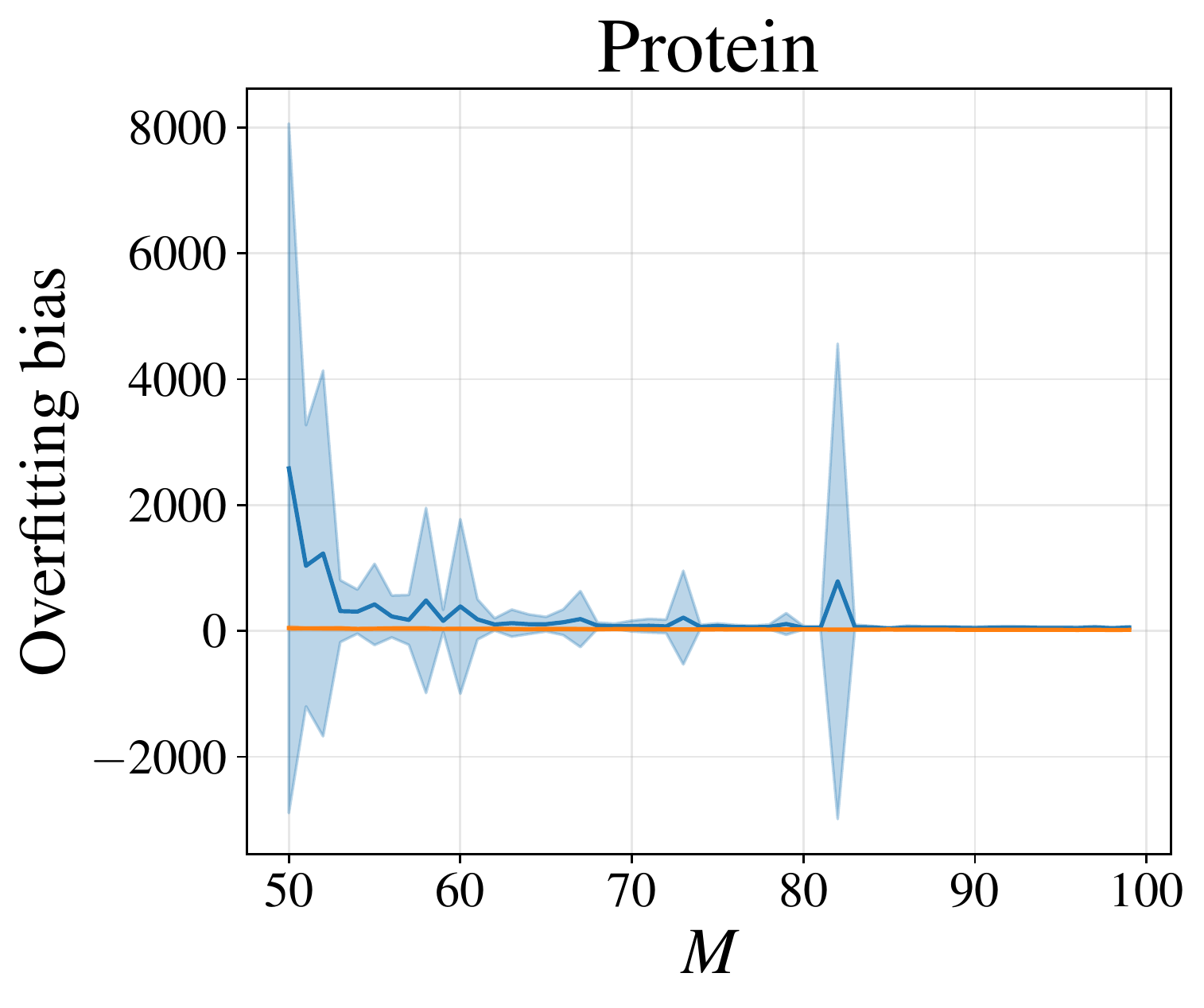}
    \end{subfigure} 
    \begin{subfigure}[b]{0.31\textwidth}
        \centering
        \includegraphics[width=\linewidth]{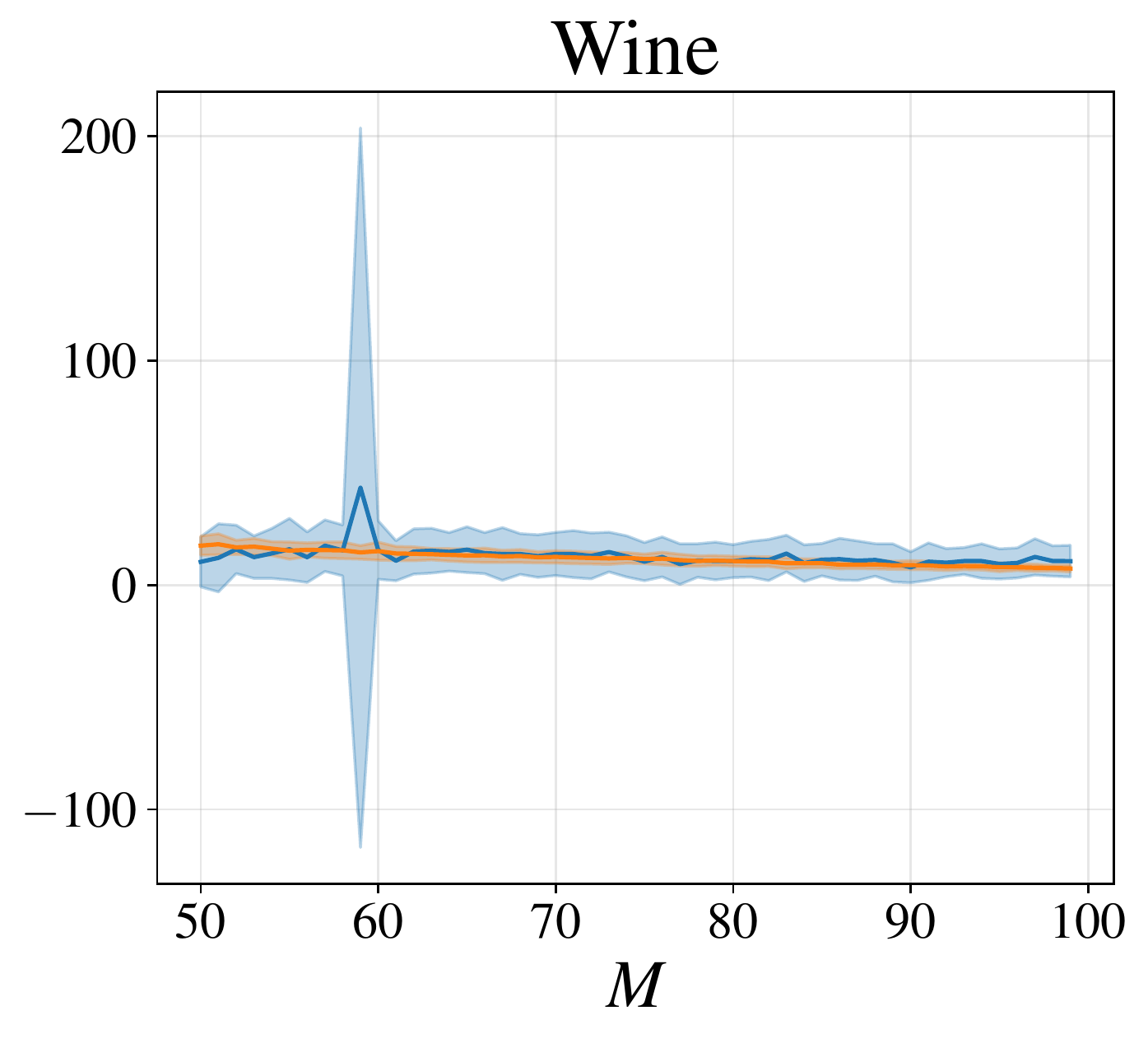}
    \end{subfigure} 
    \begin{subfigure}[b]{0.305\textwidth}
        \centering
        \includegraphics[width=\linewidth]{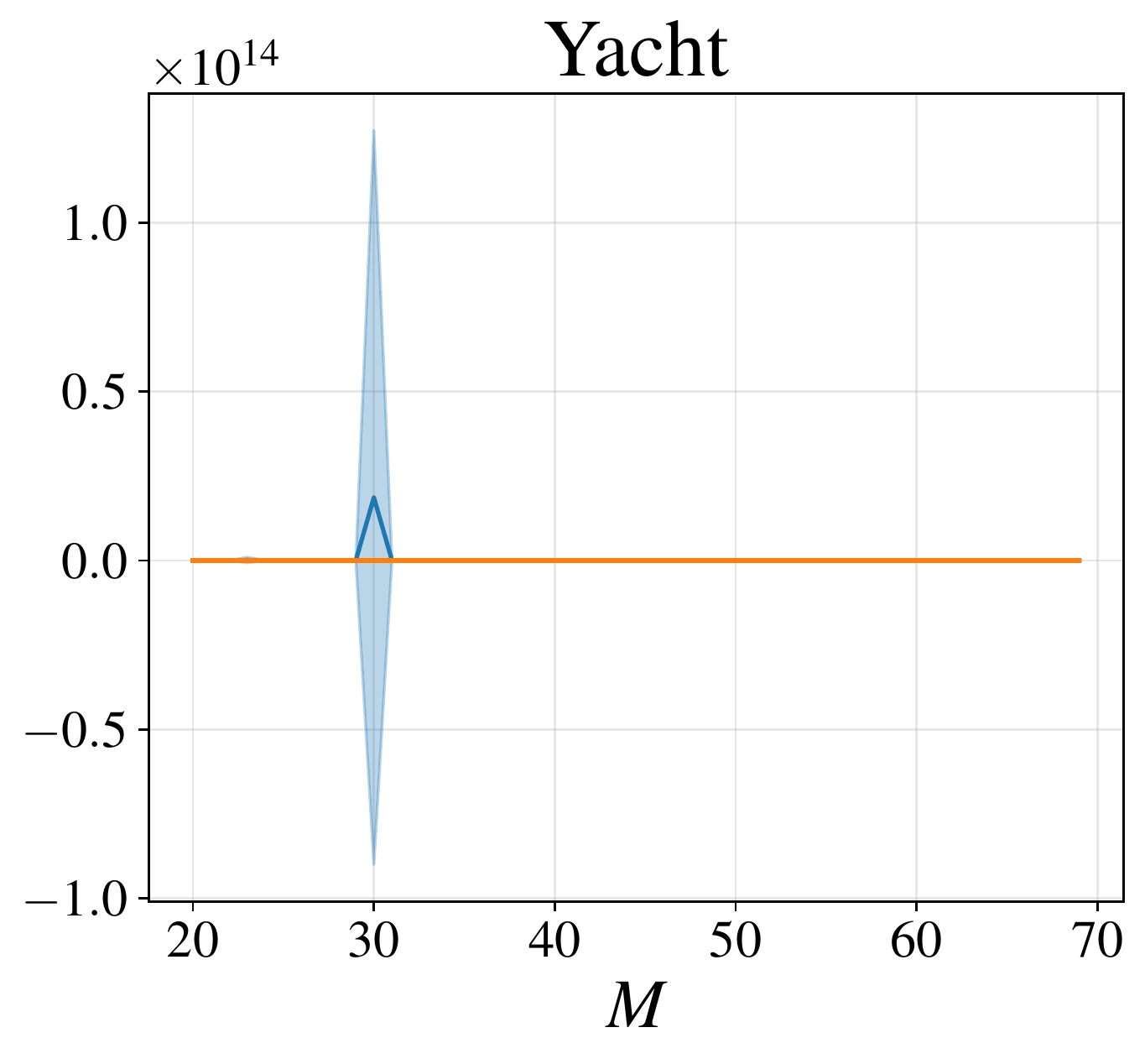}
\end{subfigure} \\
    \caption{Evolution of ``overfitting bias'', as described by~\citet{farquhar_statistical_2020}. See \cref{app:al_bias} for more details. For \textbf{\textcolor{mplblue}{DUNs}} and \textbf{\textcolor{mplorange}{MCDO}} as the amount of acquired data $M$ grows. DUNs are more robust to overfitting than MCDO. Extension of the results in \cref{fig:res_ofb_DUNvMCDO}.}
    \label{fig:app_res_ofb_DUNvMCDO}
\end{figure}

This section presents extensions of the results shown in the main text of this work, for the remaining six of the nine UCI regression datasets investigated. \Cref{fig:app_res_reg_posts_app} shows the DUN posteriors, as in \cref{fig:res_reg_posts}. \Cref{fig:app_res_reg_methods_nll} compares the performance of DUNs on active learning problems to MCDO and MFVI, as in \cref{fig:res_reg_methods_nll}. \Cref{fig:app_res_ofb_DUNvMCDO} compares overfitting in DUNs and MCDO, as in \cref{fig:res_ofb_DUNvMCDO}.

\FloatBarrier
\clearpage

\clearpage
\subsection{Other datasets}

\begin{figure}[h] 
\centering
    \begin{subfigure}[b]{0.32\textwidth}
        \centering
        \includegraphics[width=\linewidth]{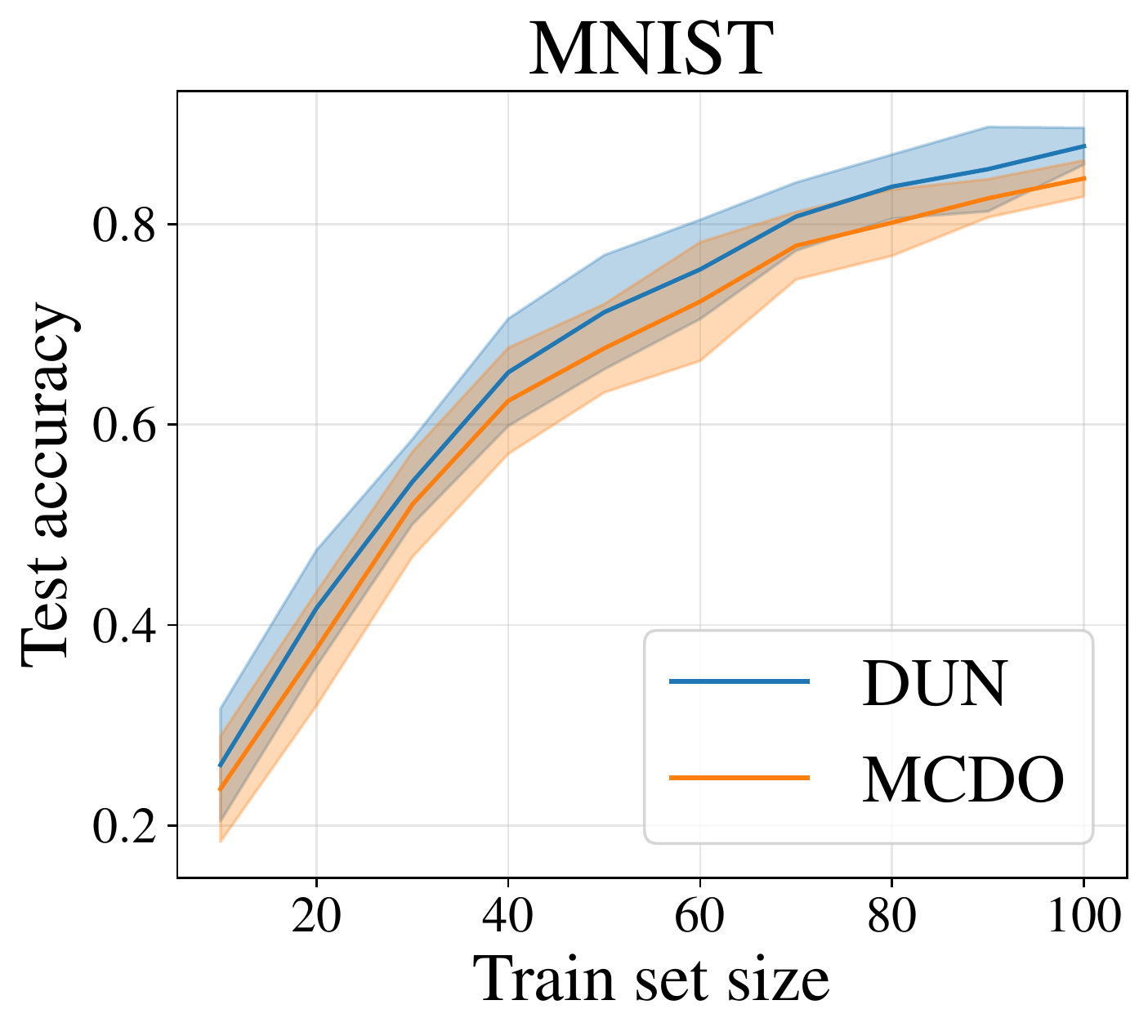}
        \caption{Accuracy}
        \label{fig:mnist_acc}
    \end{subfigure}
    \begin{subfigure}[b]{0.32\textwidth}
        \centering
        \includegraphics[width=\linewidth]{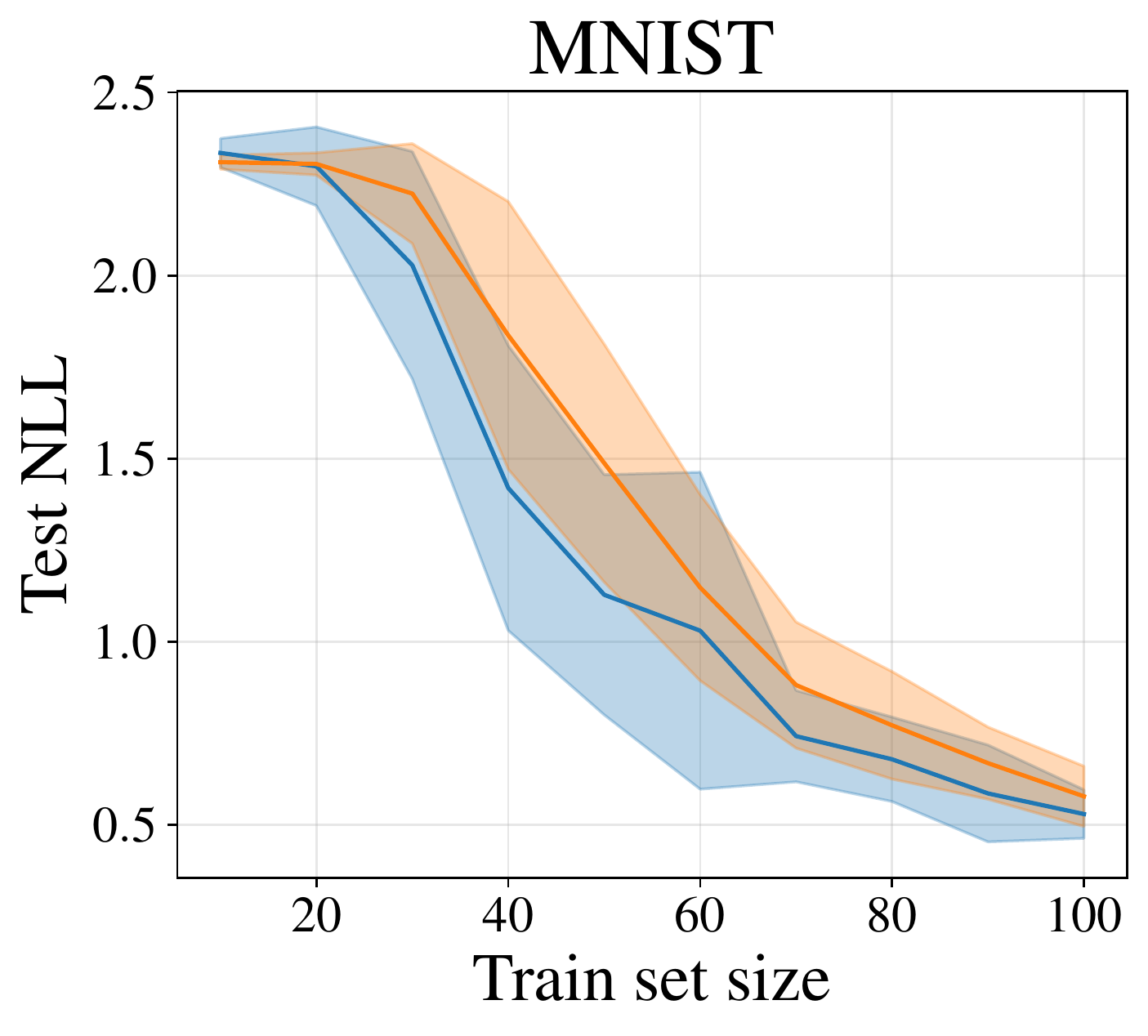}
        \caption{NLL}
        \label{fig:mnist_nll}
    \end{subfigure} 
    \begin{subfigure}[b]{0.32\textwidth}
        \centering
        \includegraphics[width=\linewidth]{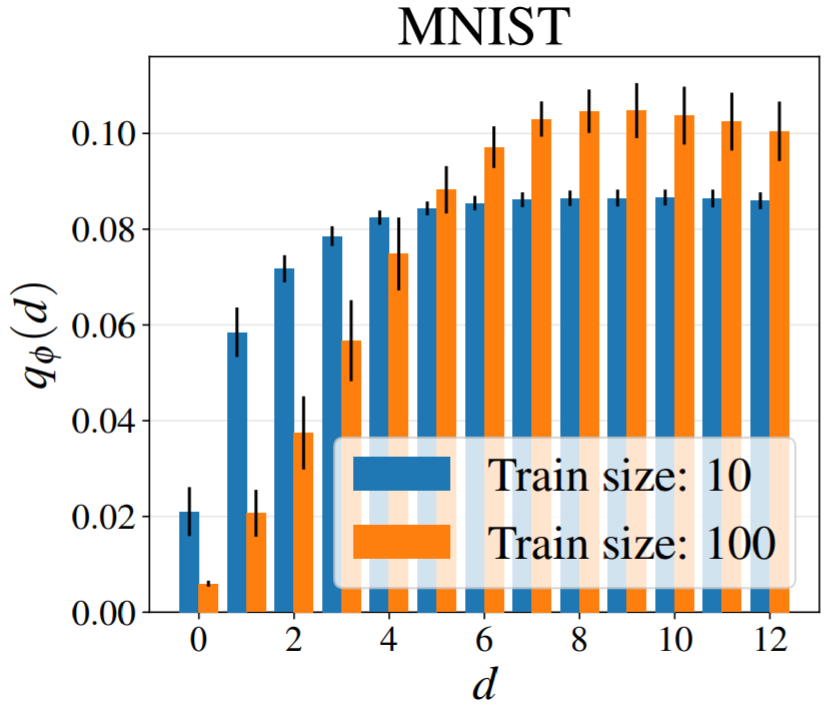}
        \caption{DUN depth posterior}
        \label{fig:mnist_post}
    \end{subfigure}
    \caption{Test accuracy (left) and NLL (middle) vs. number of training points evaluated on MNIST. The performance of \textbf{\textcolor{mplblue}{DUNs}} and \textbf{\textcolor{mplorange}{MCDO}} is compared. Right: DUN posterior probabilities over depth, for the \textbf{\textcolor{mplblue}{smallest}} and \textbf{\textcolor{mplorange}{largest}} labelled datasets used in active learning.} \label{fig:app_res_mnist}
\end{figure}

\begin{figure}[h] 
\centering
    \begin{subfigure}[b]{0.325\textwidth}
        \centering
        \includegraphics[width=\linewidth]{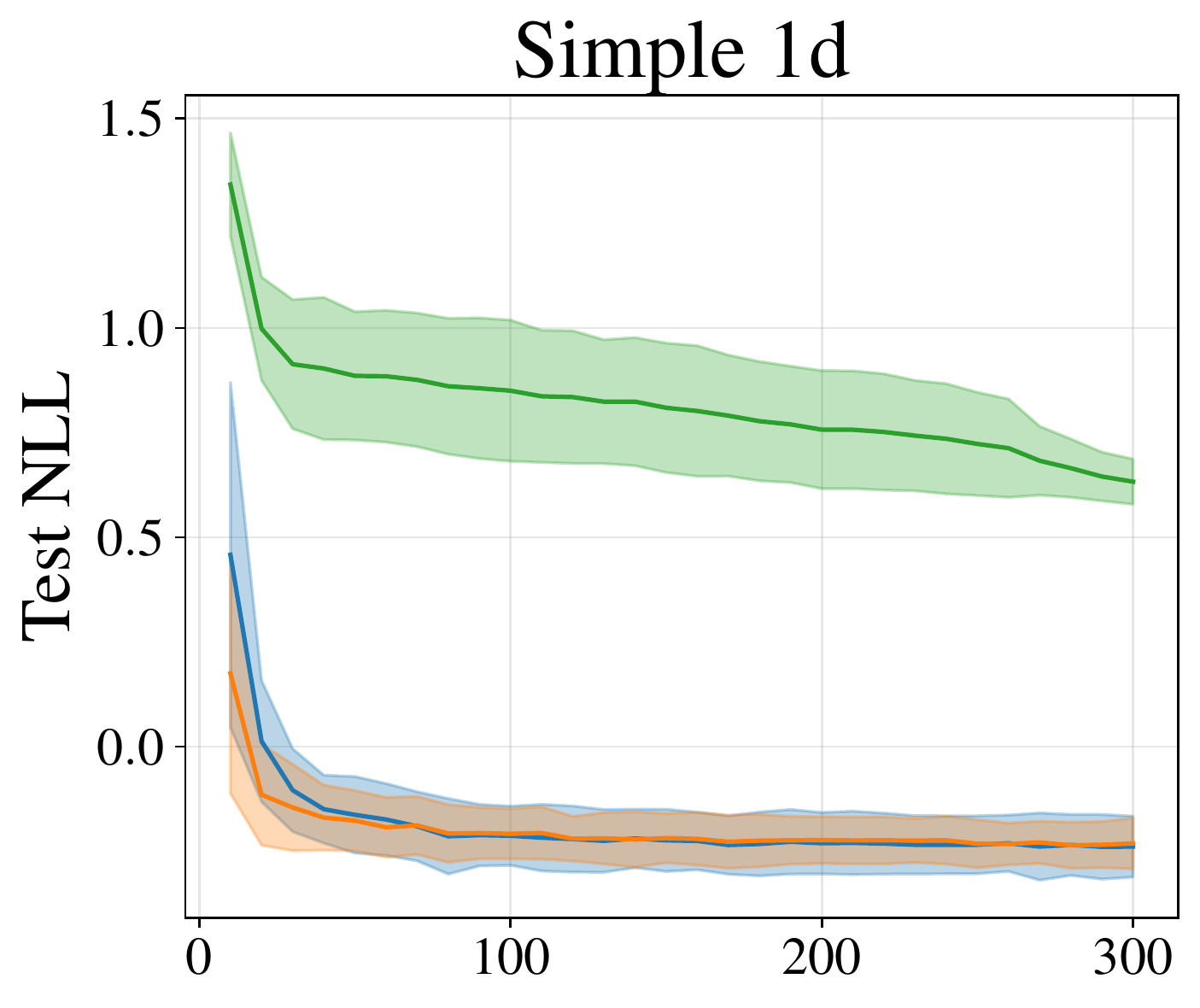}
    \end{subfigure}
    \begin{subfigure}[b]{0.315\textwidth}
        \centering
        \includegraphics[width=\linewidth]{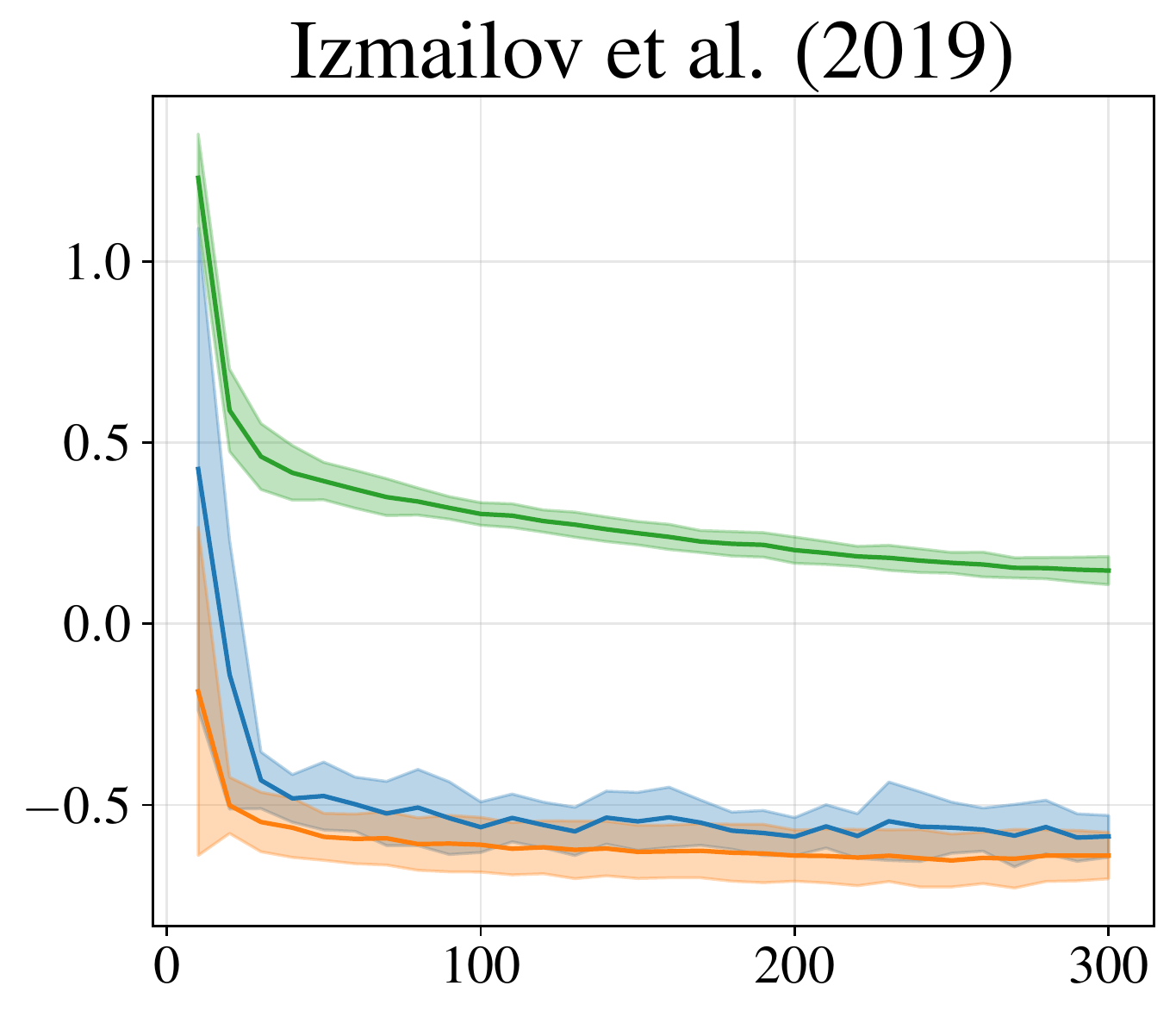}
    \end{subfigure} 
    \begin{subfigure}[b]{0.315\textwidth}
        \centering
        \includegraphics[width=\linewidth]{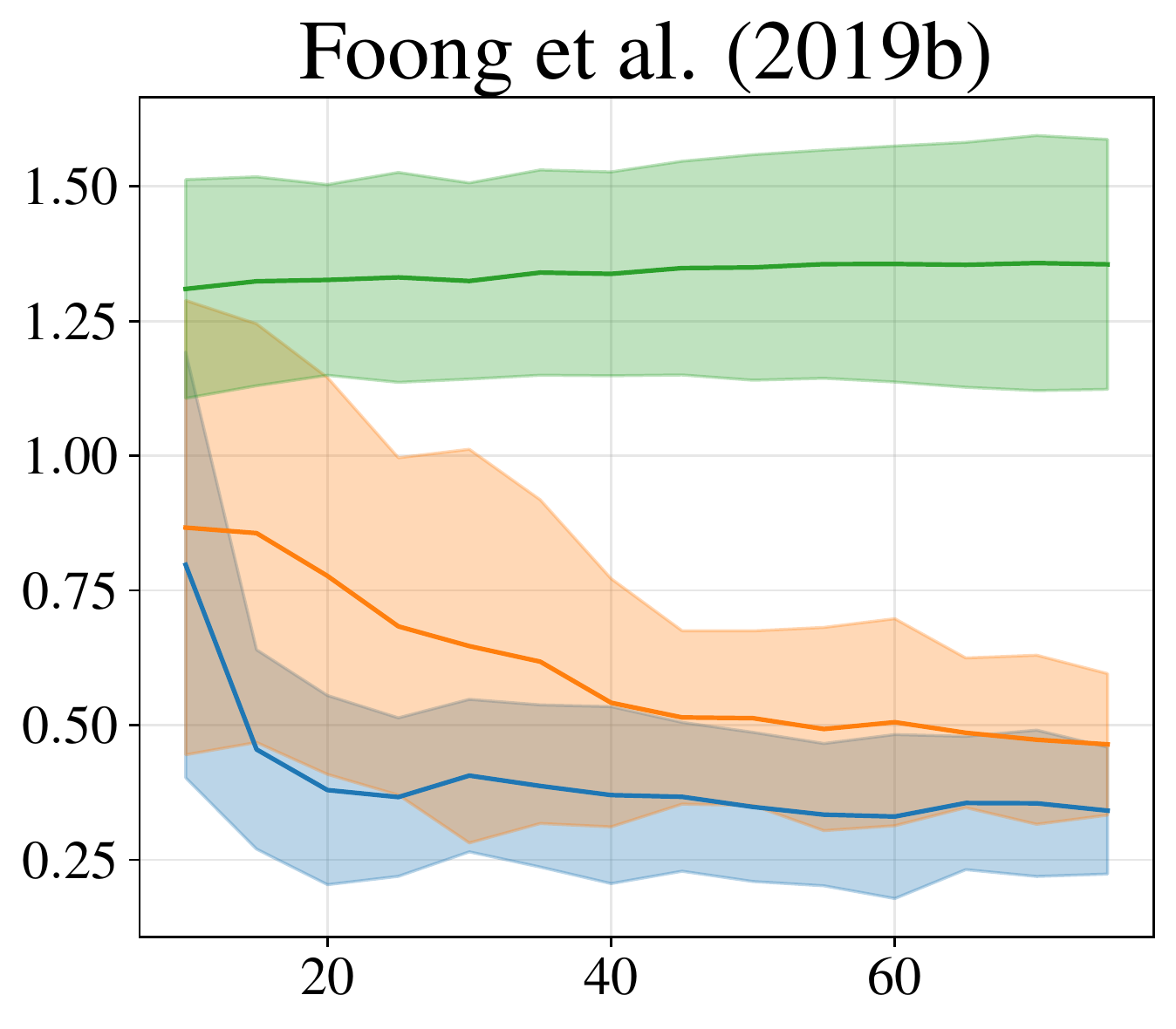} 
    \end{subfigure} \\
    \begin{subfigure}[b]{0.34\textwidth}
        \centering
        \includegraphics[width=\linewidth]{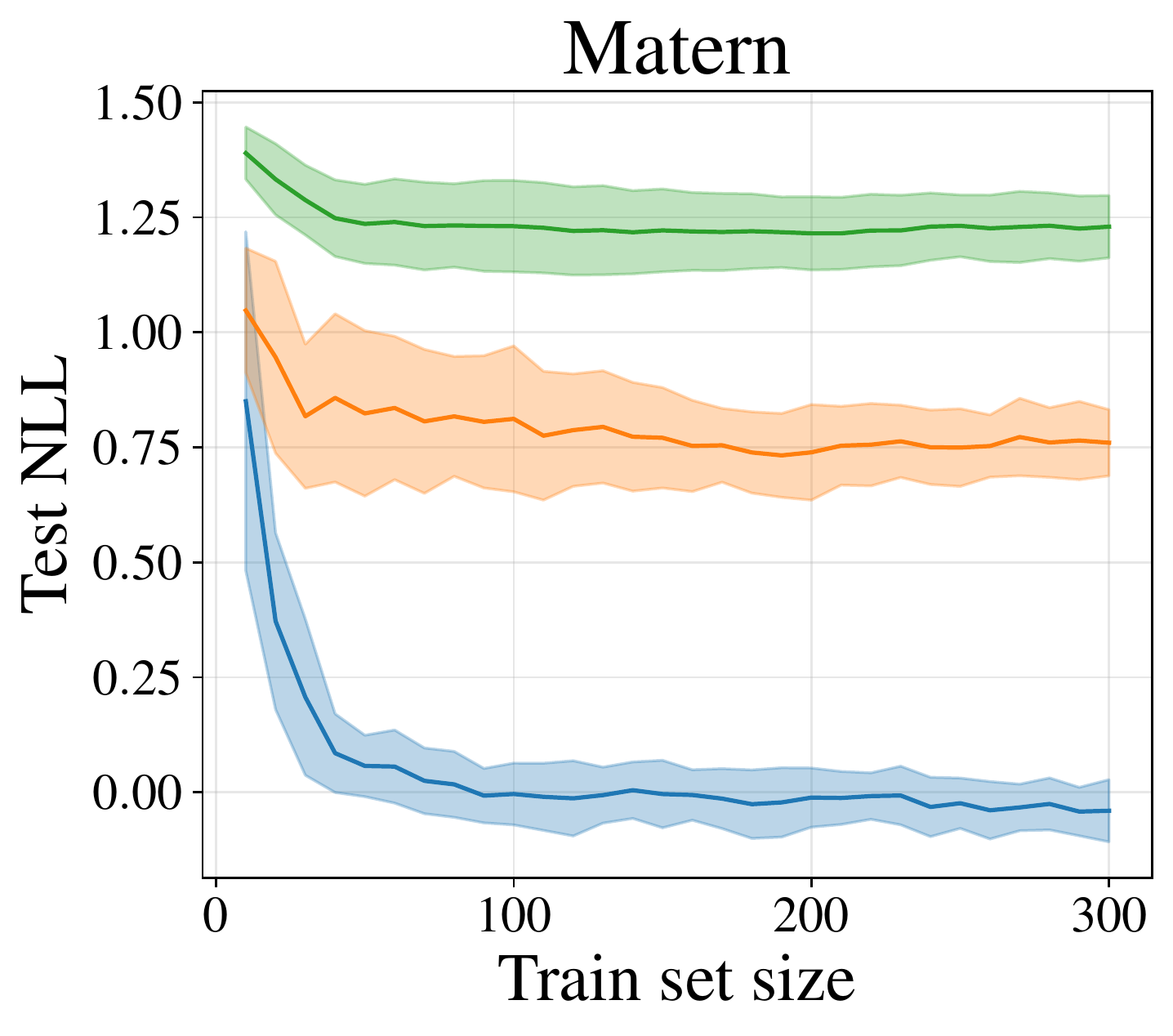}
    \end{subfigure} 
    \begin{subfigure}[b]{0.31\textwidth}
        \centering
        \includegraphics[width=\linewidth]{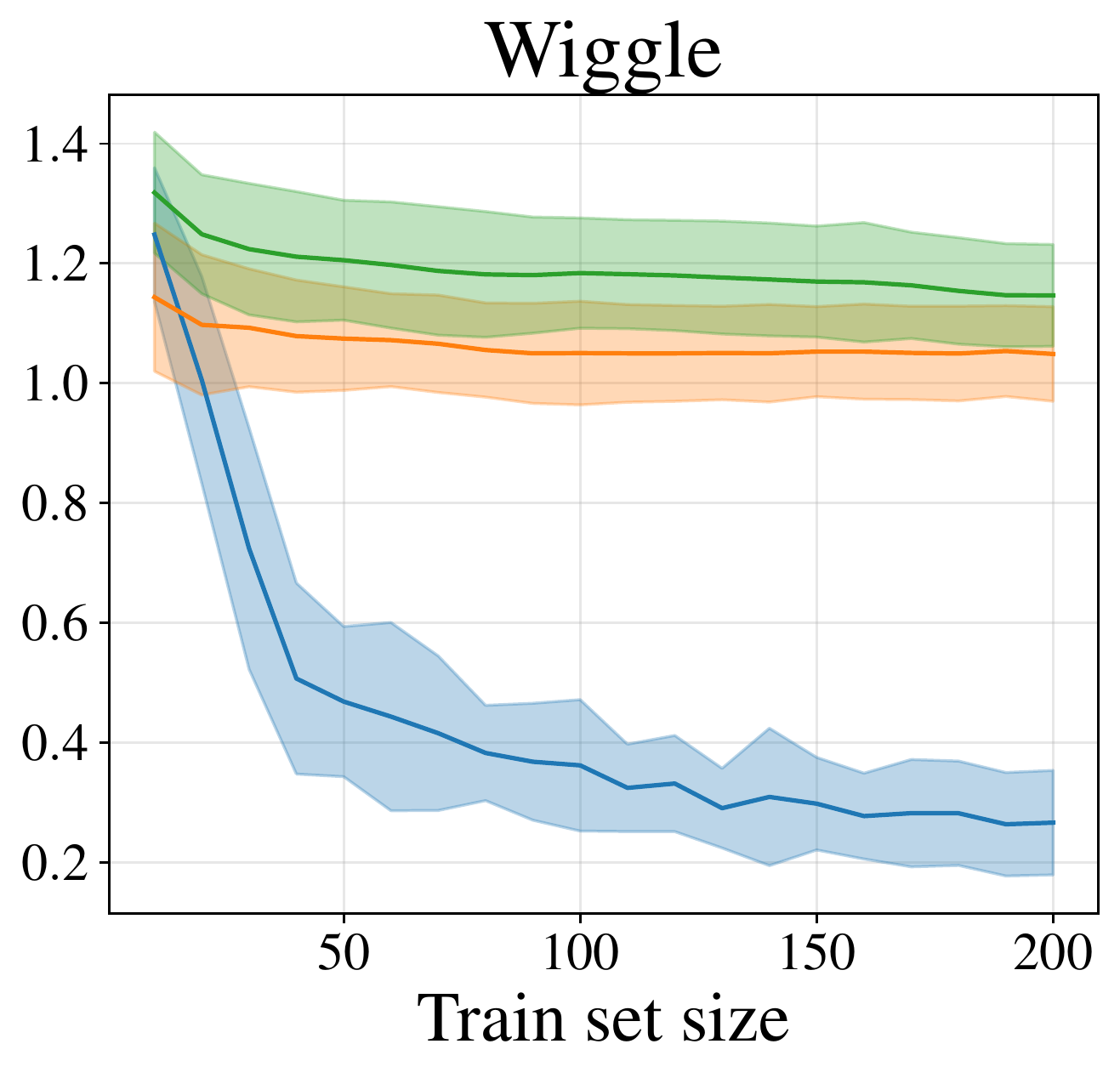}
    \end{subfigure} 
    \begin{subfigure}[b]{0.31\textwidth}
        \centering
        \includegraphics[width=\linewidth]{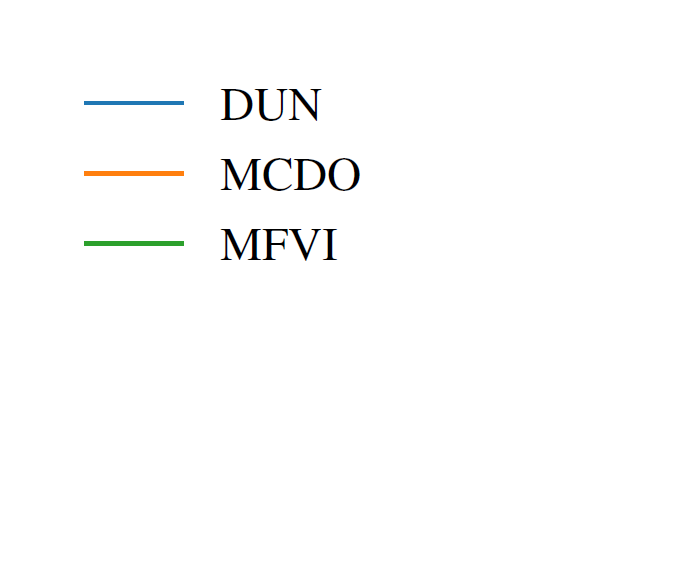}
    \end{subfigure} \\
    \caption{Test NLL vs. number of training points evaluated on toy datasets. The performance of \textbf{\textcolor{mplblue}{DUNs}}, \textbf{\textcolor{mplorange}{MCDO}} and \textbf{\textcolor{mplgreen}{MFVI}} is compared.}
    \label{fig:res_toy_methods_nll}
\end{figure}

The following two figures present results for MNIST (\cref{fig:app_res_mnist}) and the toy regression datasets (\cref{fig:res_toy_methods_nll}). These results are consistent with the UCI regression results---DUNs either outperform the baseline methods (in MNIST, \citet{foong2019between}, Matern and Wiggle), or perform comparably to MCDO (Simple 1d, \citet{izmailov2020subspace}).

\subsubsection{MNIST results}

\subsubsection{Toy dataset results}

\FloatBarrier
\newpage
\subsection{RMSE results}\label{app:rmse}

\begin{figure}[h] 
\centering
    \begin{subfigure}[b]{0.33\textwidth}
        \centering
        \includegraphics[width=\linewidth]{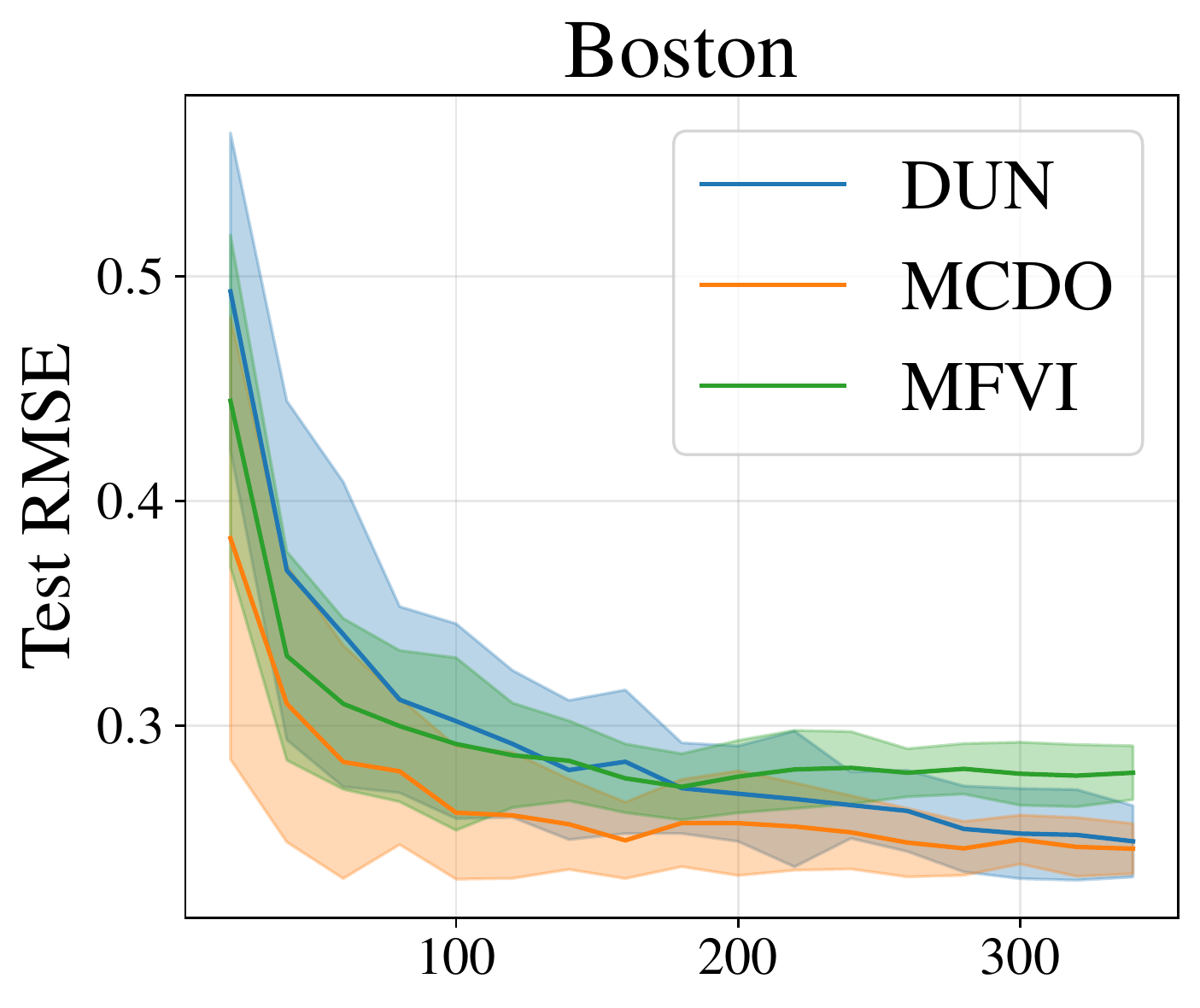}
    \end{subfigure}
    \begin{subfigure}[b]{0.31\textwidth}
        \centering
        \includegraphics[width=\linewidth]{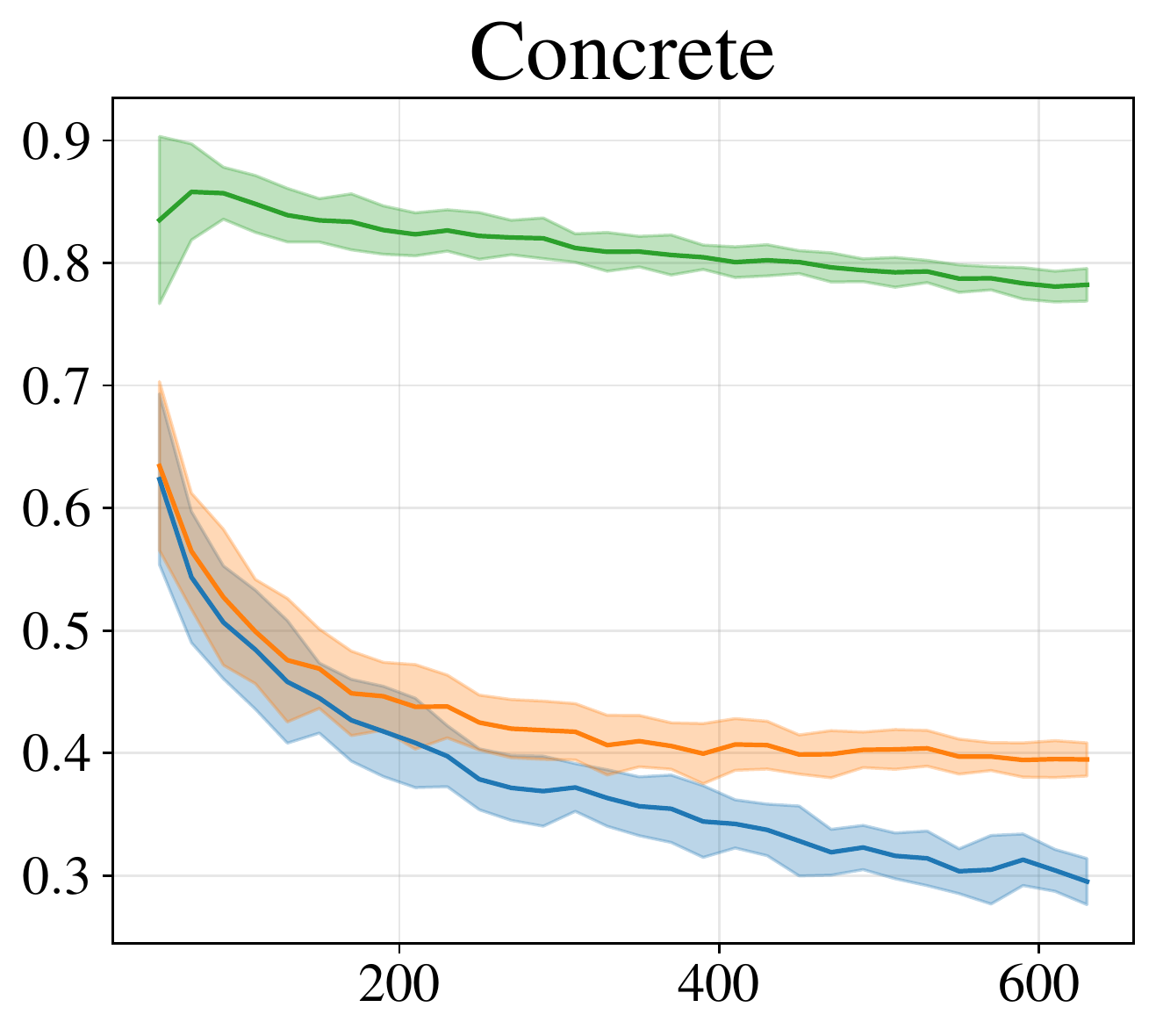}
    \end{subfigure} 
    \begin{subfigure}[b]{0.32\textwidth}
        \centering
        \includegraphics[width=\linewidth]{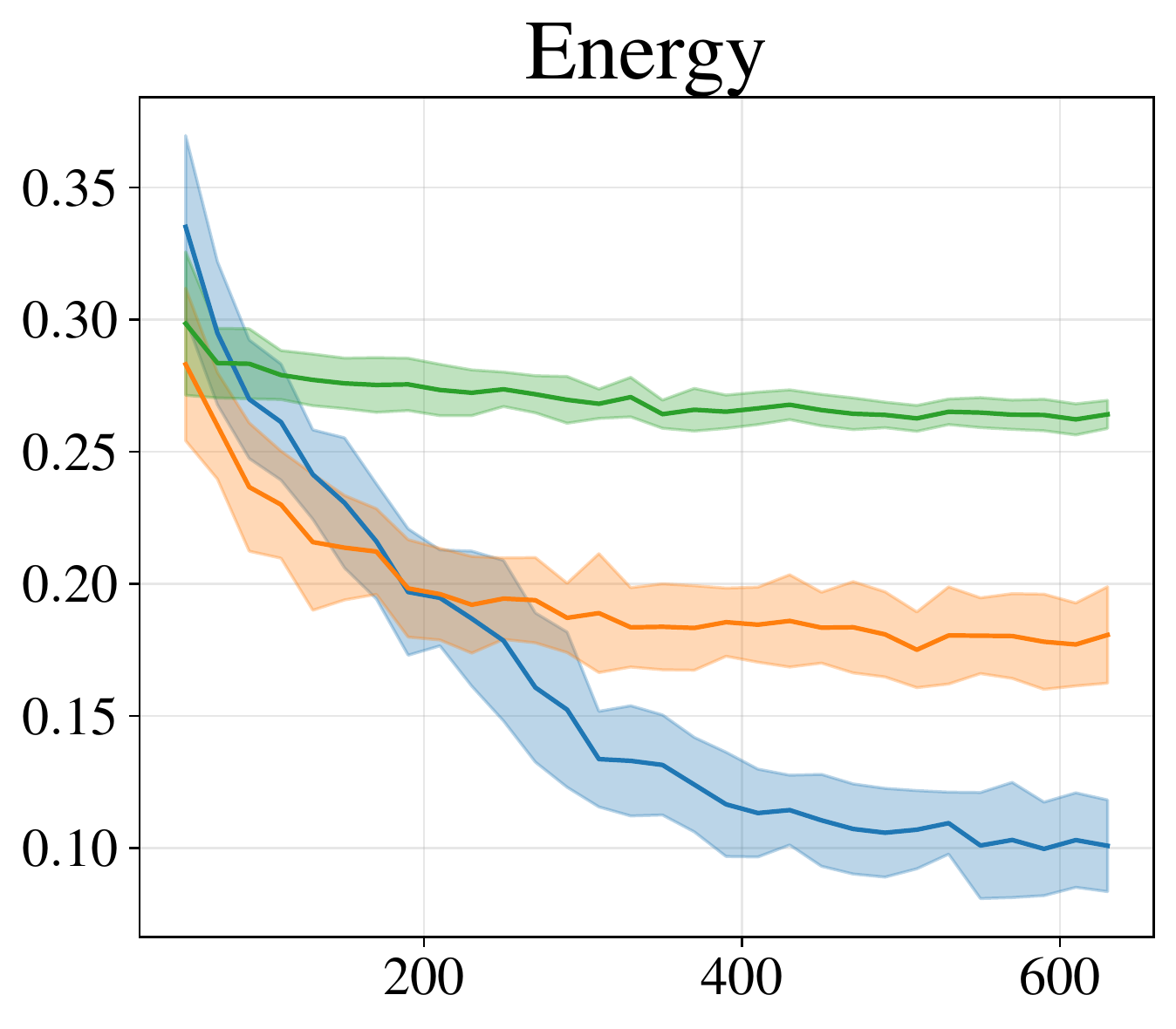} 
    \end{subfigure} \\
    \begin{subfigure}[b]{0.33\textwidth}
        \centering
        \includegraphics[width=\linewidth]{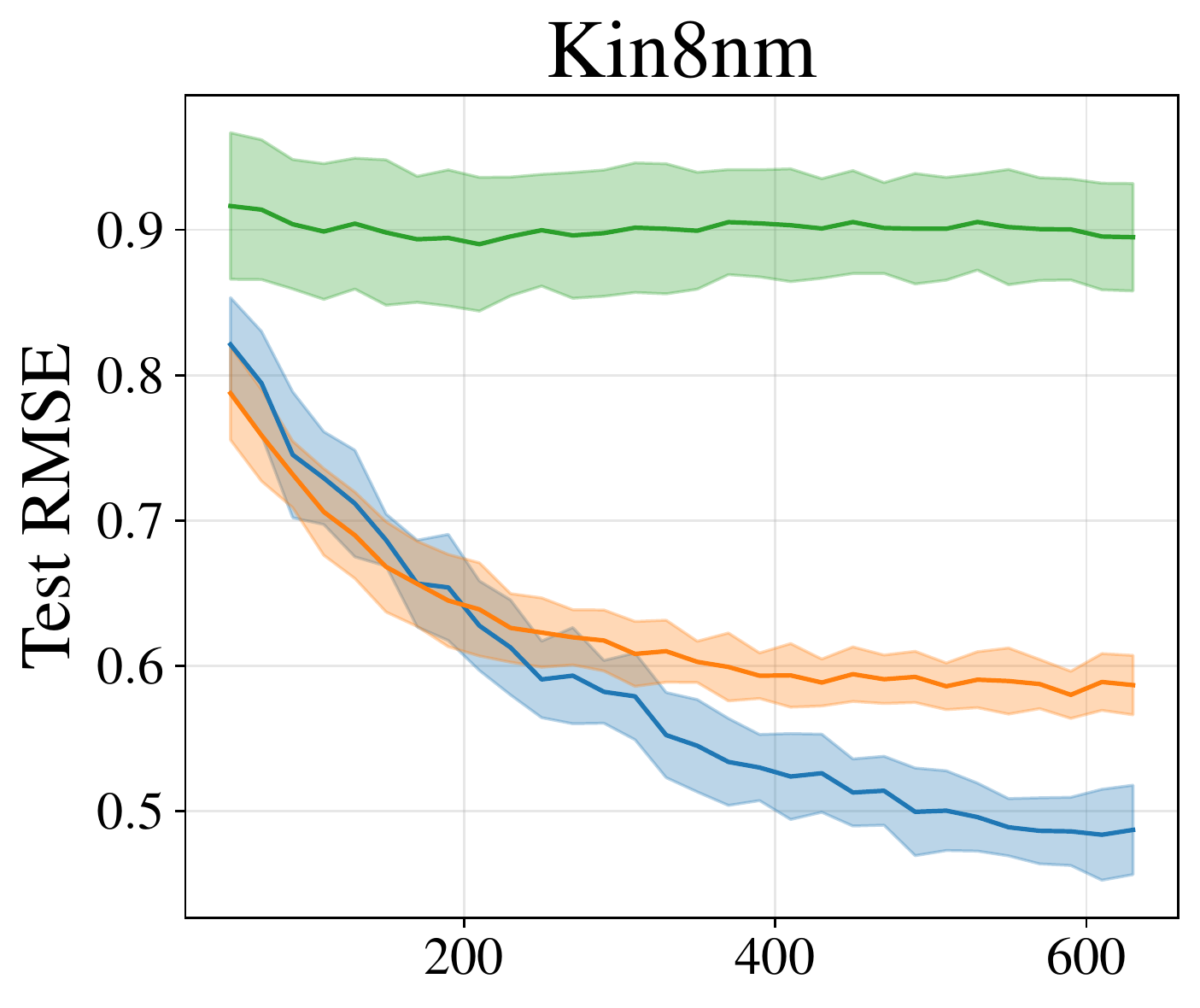}
    \end{subfigure} 
    \begin{subfigure}[b]{0.31\textwidth}
        \centering
        \includegraphics[width=\linewidth]{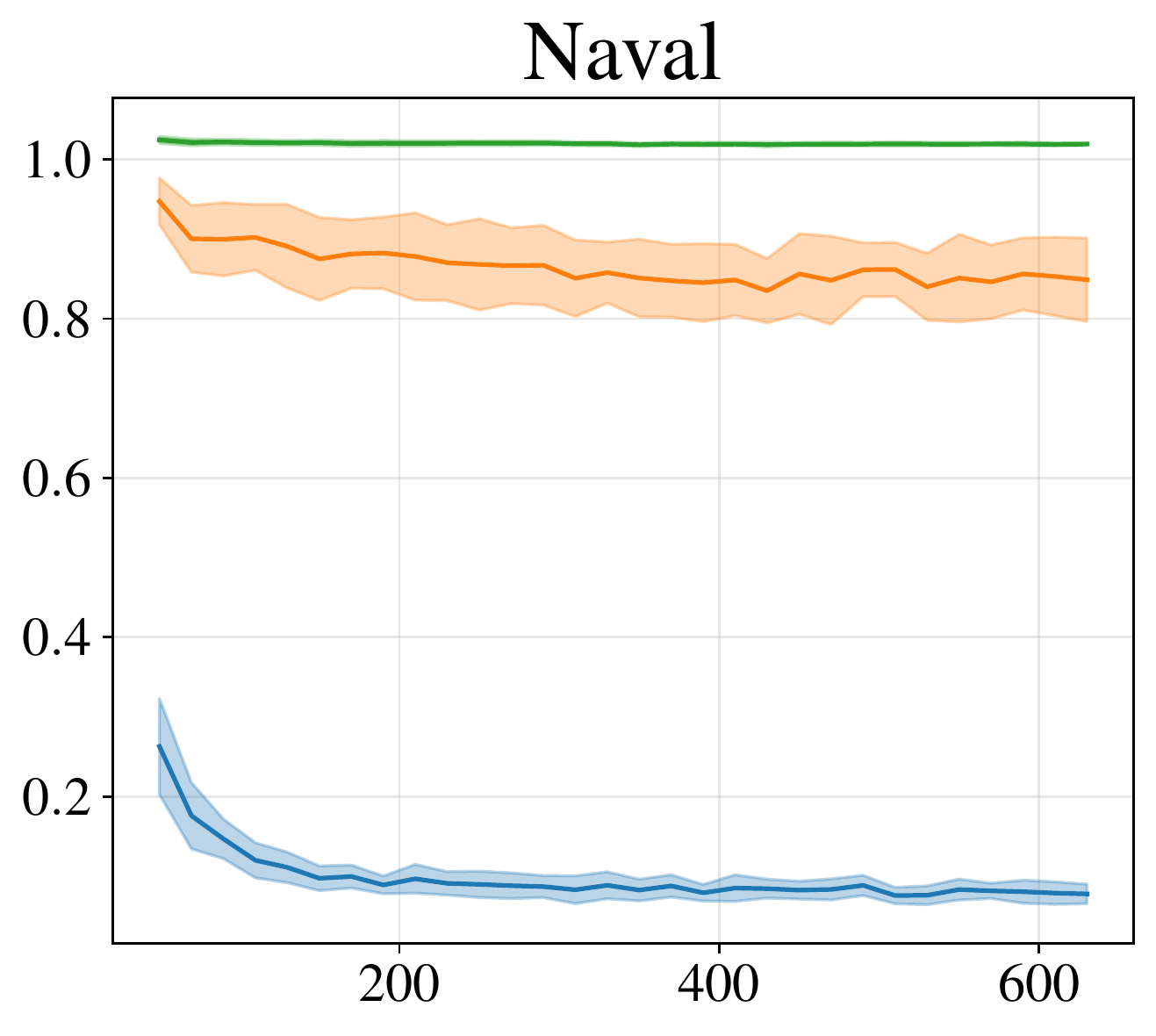}
    \end{subfigure} 
    \begin{subfigure}[b]{0.32\textwidth}
        \centering
        \includegraphics[width=\linewidth]{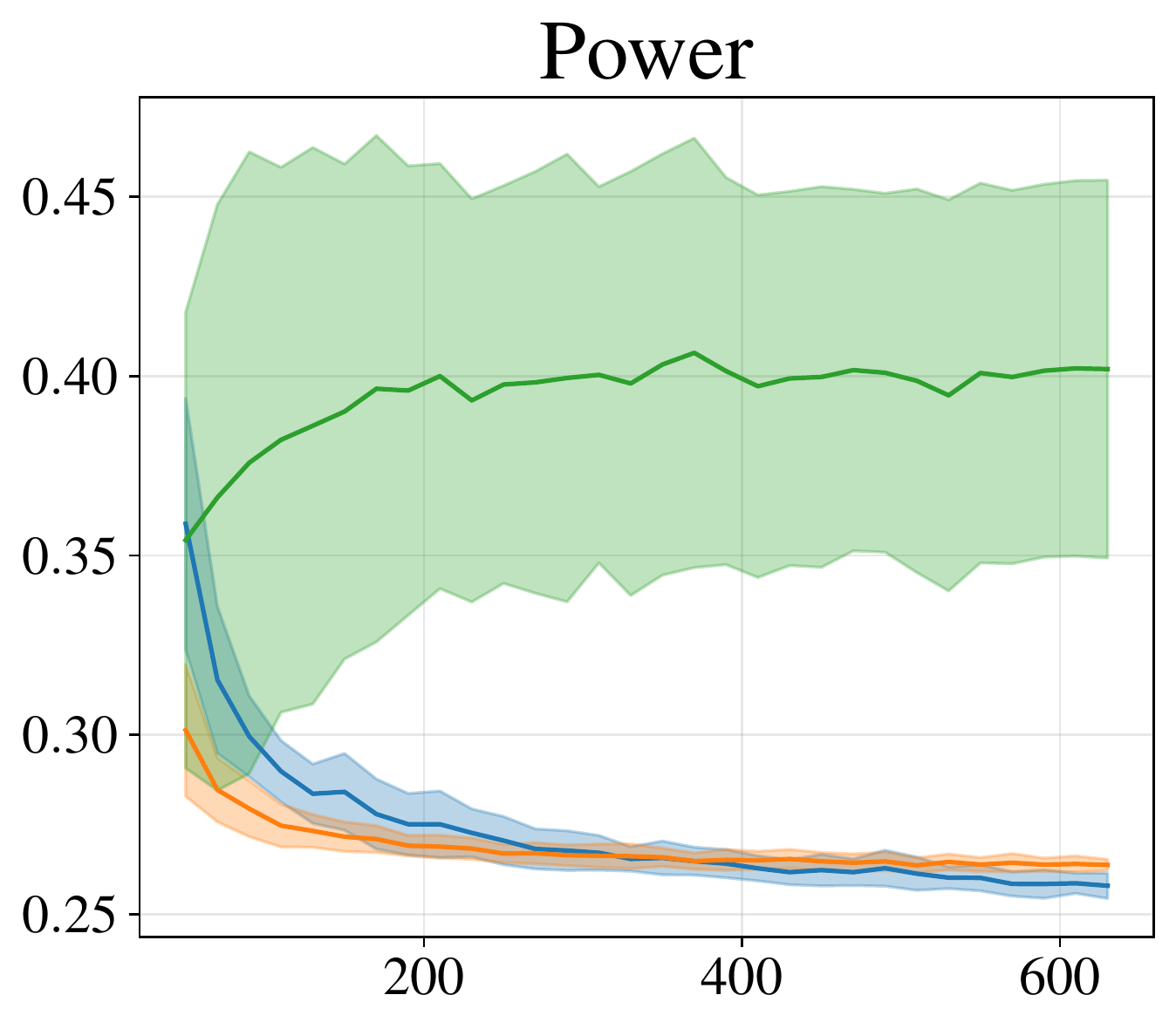}
    \end{subfigure} \\
    \begin{subfigure}[b]{0.333\textwidth}
        \centering
        \includegraphics[width=\linewidth]{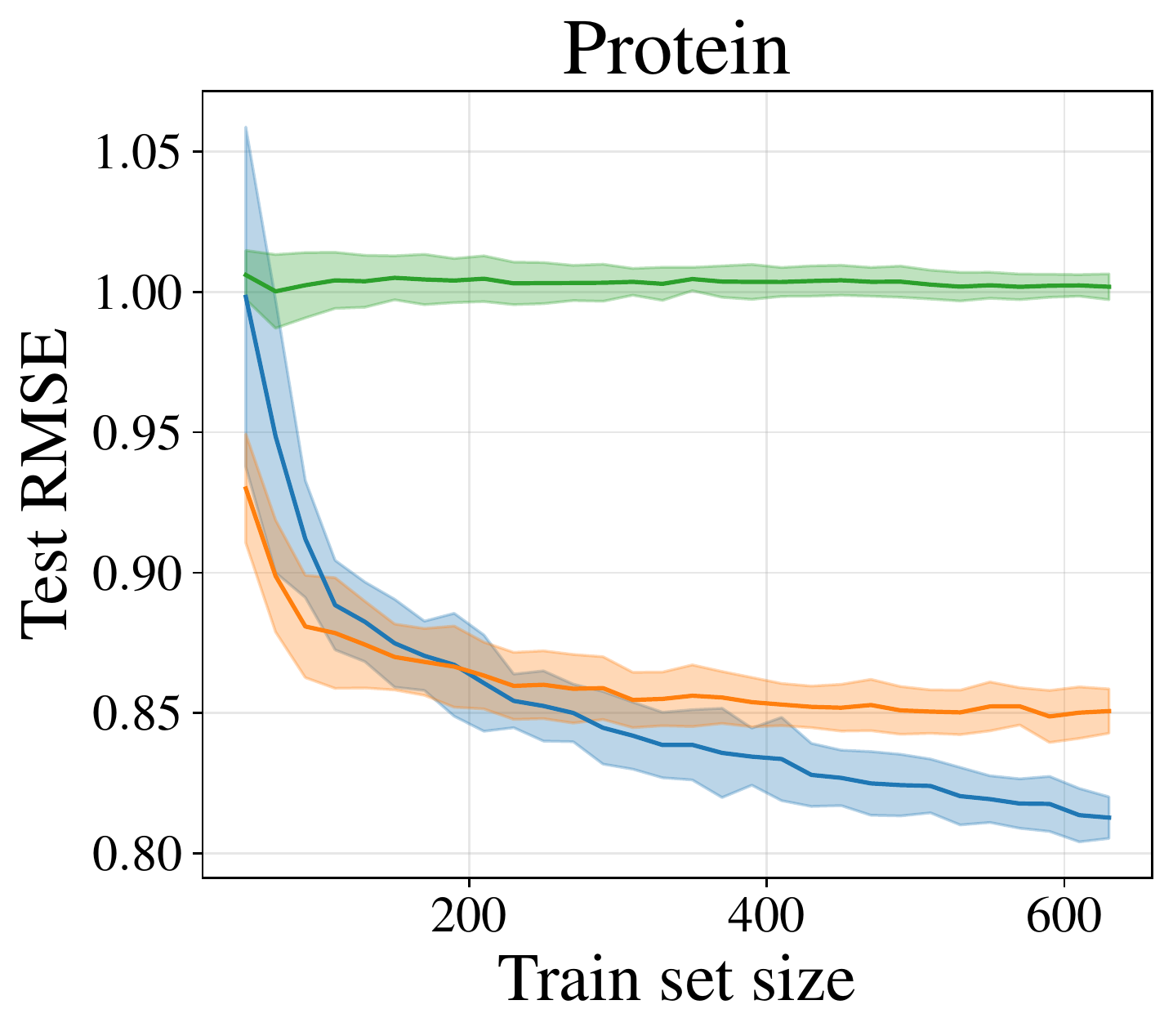}
    \end{subfigure} 
    \begin{subfigure}[b]{0.318\textwidth}
        \centering
        \includegraphics[width=\linewidth]{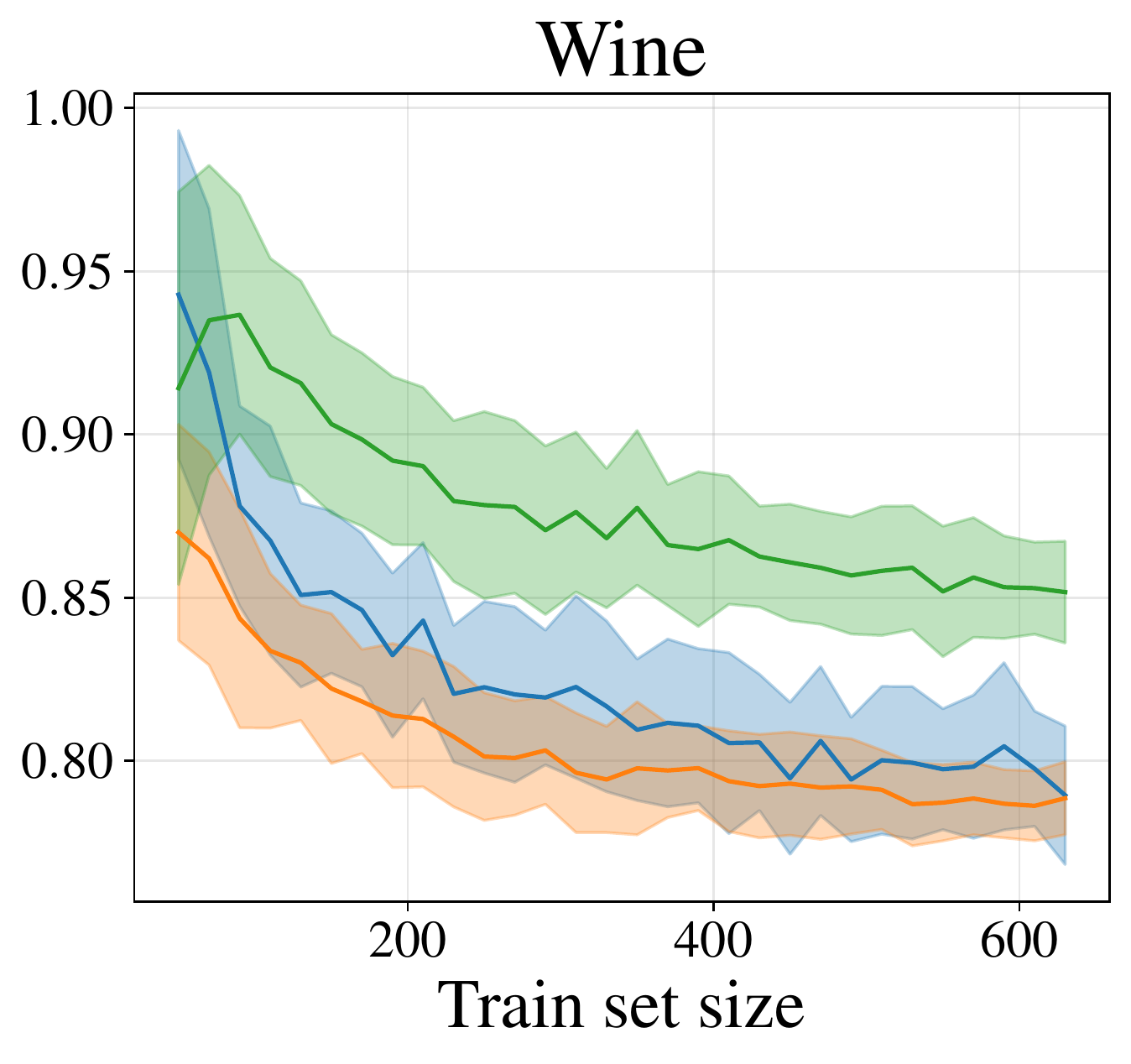}
    \end{subfigure} 
    \begin{subfigure}[b]{0.314\textwidth}
        \centering
        \includegraphics[width=\linewidth]{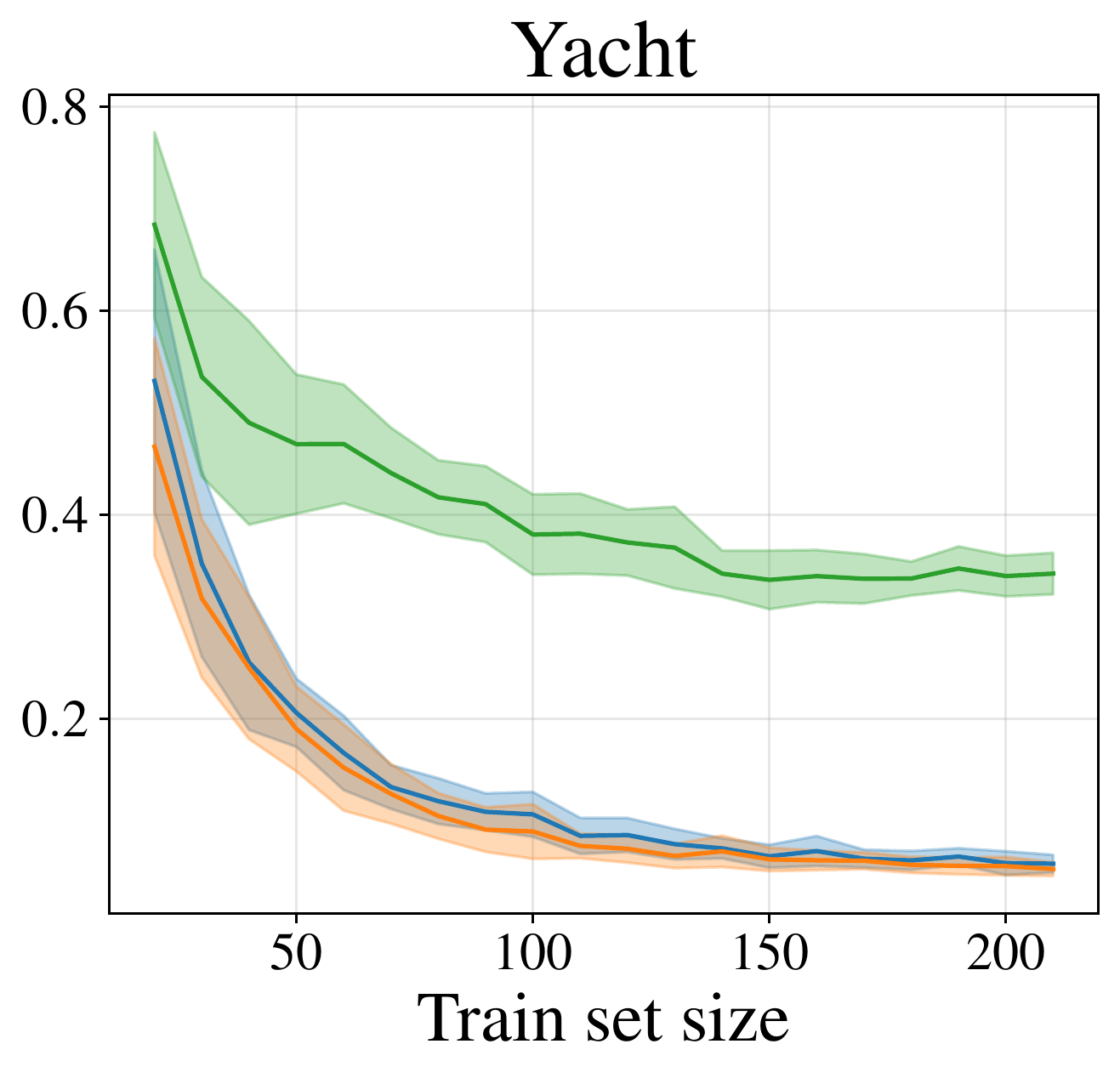}
    \end{subfigure} \\
    \caption{Test RMSE vs. number of training points evaluated on UCI datasets. The performance of \textbf{\textcolor{mplblue}{DUNs}}, \textbf{\textcolor{mplorange}{MCDO}} and \textbf{\textcolor{mplgreen}{MFVI}} is compared.}
    \label{fig:res_reg_methods_err}
\end{figure}

\Cref{fig:res_reg_methods_err} shows the root mean squared error (RMSE) performance of DUNs, MCDO and MFVI on the UCI regression datasets. It is equivalent to \cref{fig:res_reg_methods_nll} and \cref{fig:app_res_reg_methods_nll} but for RMSE rather than NLL. The RMSE results are consistent with the NLL results, with DUNs outperforming the baseline methods in five of the nine datasets, and performing comparably to MCDO in the remaining four datasets.

\FloatBarrier
\newpage
\subsection{Temperature of the proposal distribution}\label{app:temp}

\begin{figure}[h] 
\centering
    \begin{subfigure}[b]{0.33\textwidth}
        \centering
        \includegraphics[width=\linewidth]{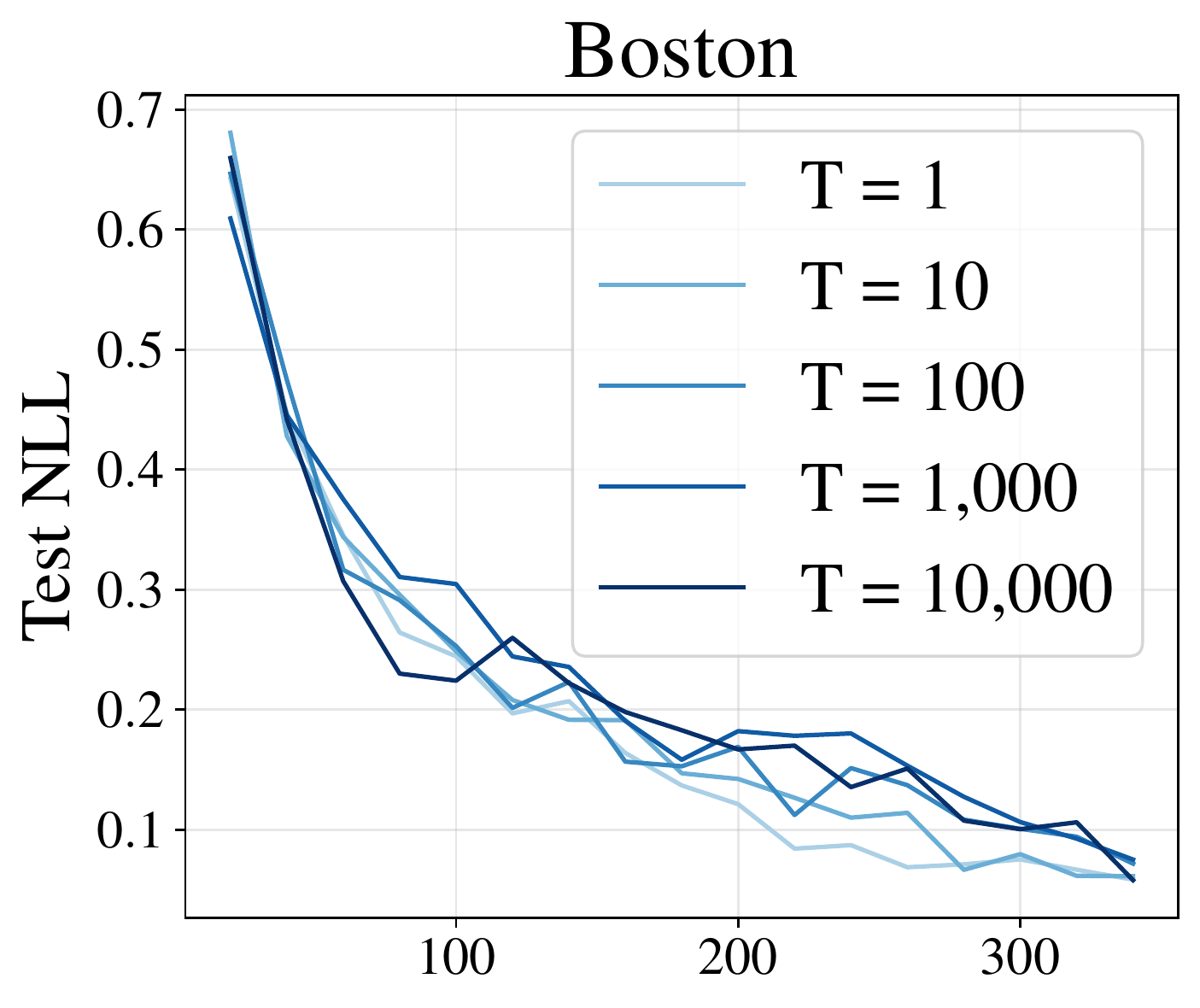}
    \end{subfigure}
    \begin{subfigure}[b]{0.31\textwidth}
        \centering
        \includegraphics[width=\linewidth]{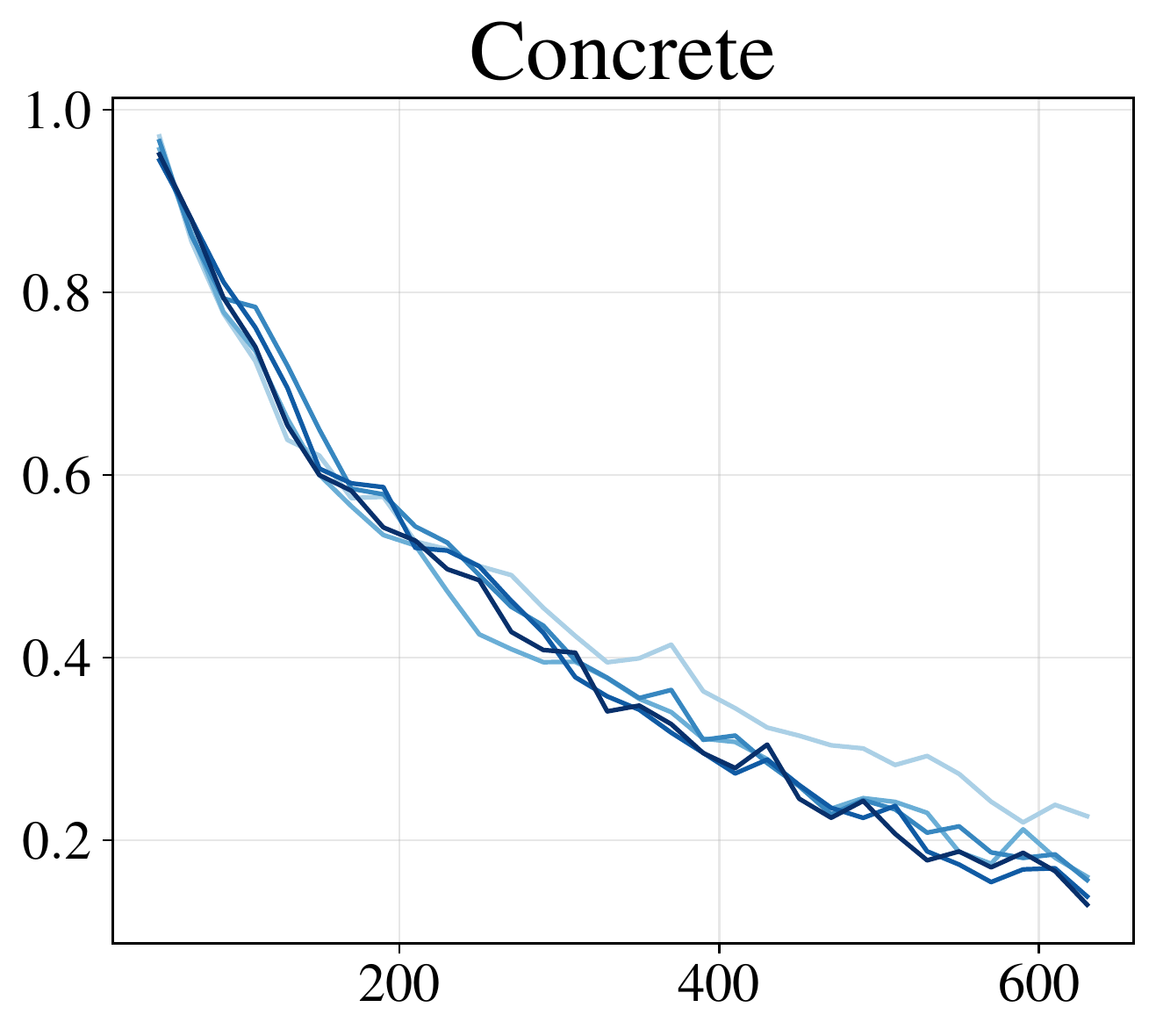}
    \end{subfigure} 
    \begin{subfigure}[b]{0.325\textwidth}
        \centering
        \includegraphics[width=\linewidth]{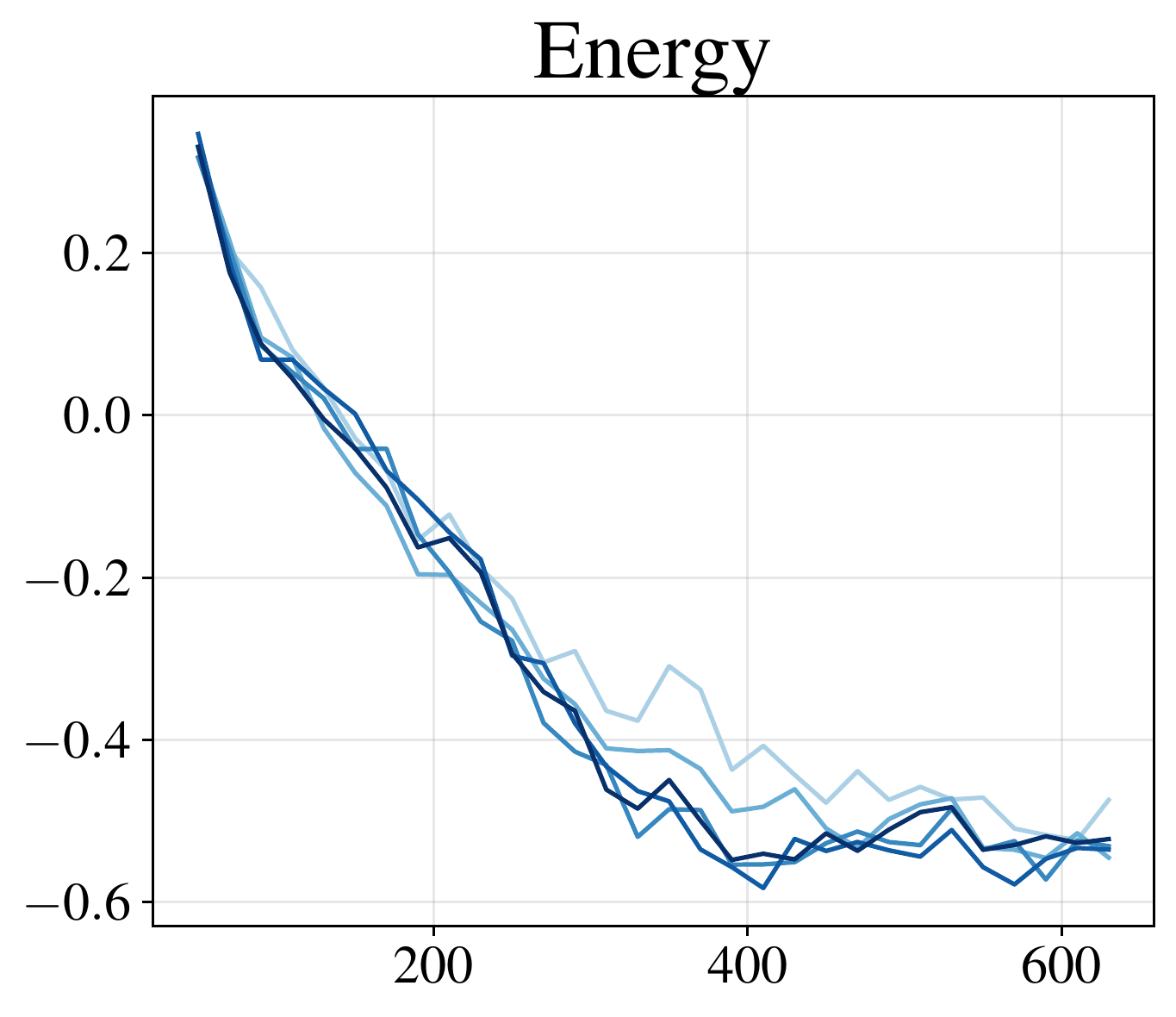} 
    \end{subfigure} \\
    \begin{subfigure}[b]{0.33\textwidth}
        \centering
        \includegraphics[width=\linewidth]{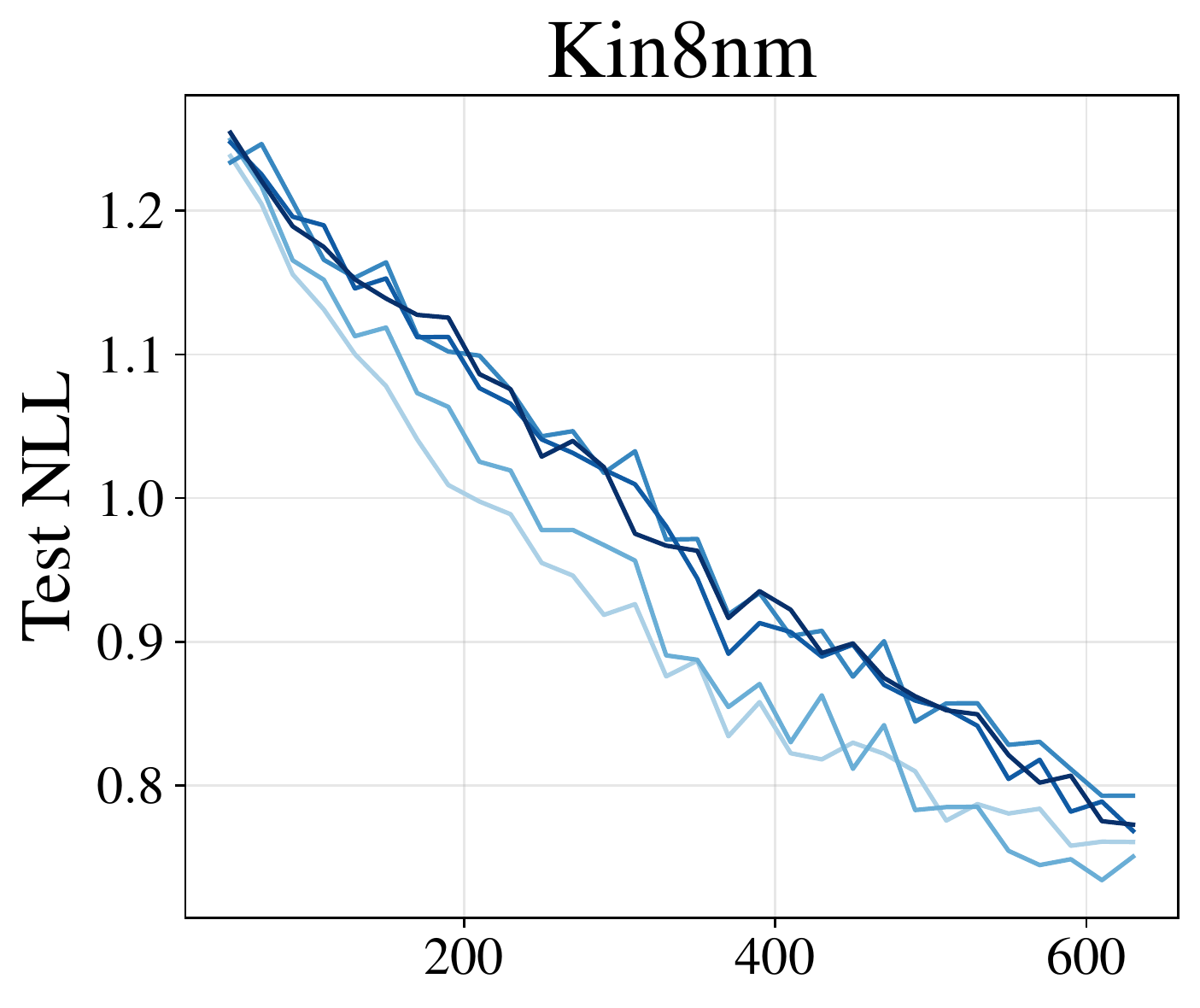}
    \end{subfigure} 
    \begin{subfigure}[b]{0.32\textwidth}
        \centering
        \includegraphics[width=\linewidth]{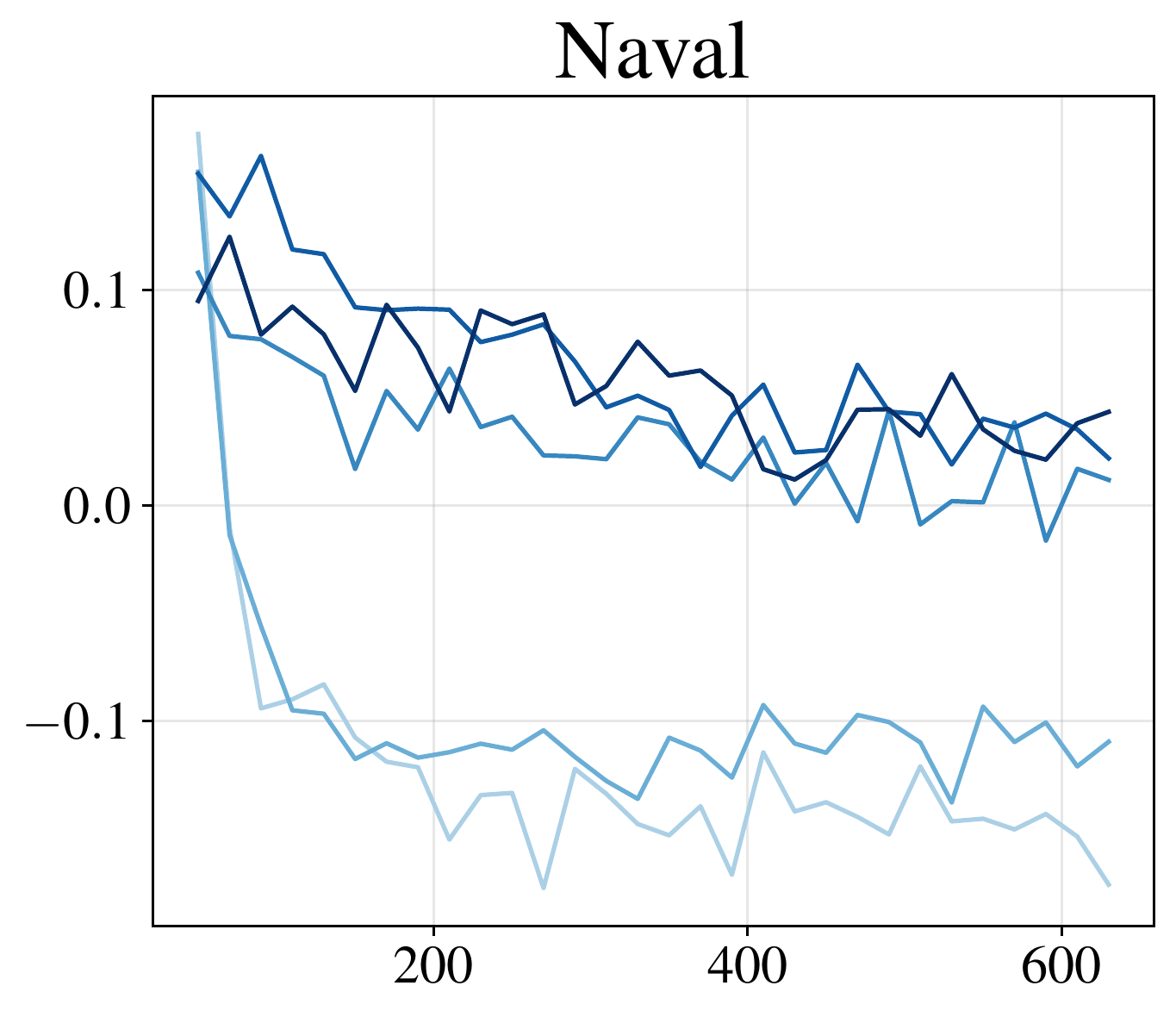}
    \end{subfigure} 
    \begin{subfigure}[b]{0.315\textwidth}
        \centering
        \includegraphics[width=\linewidth]{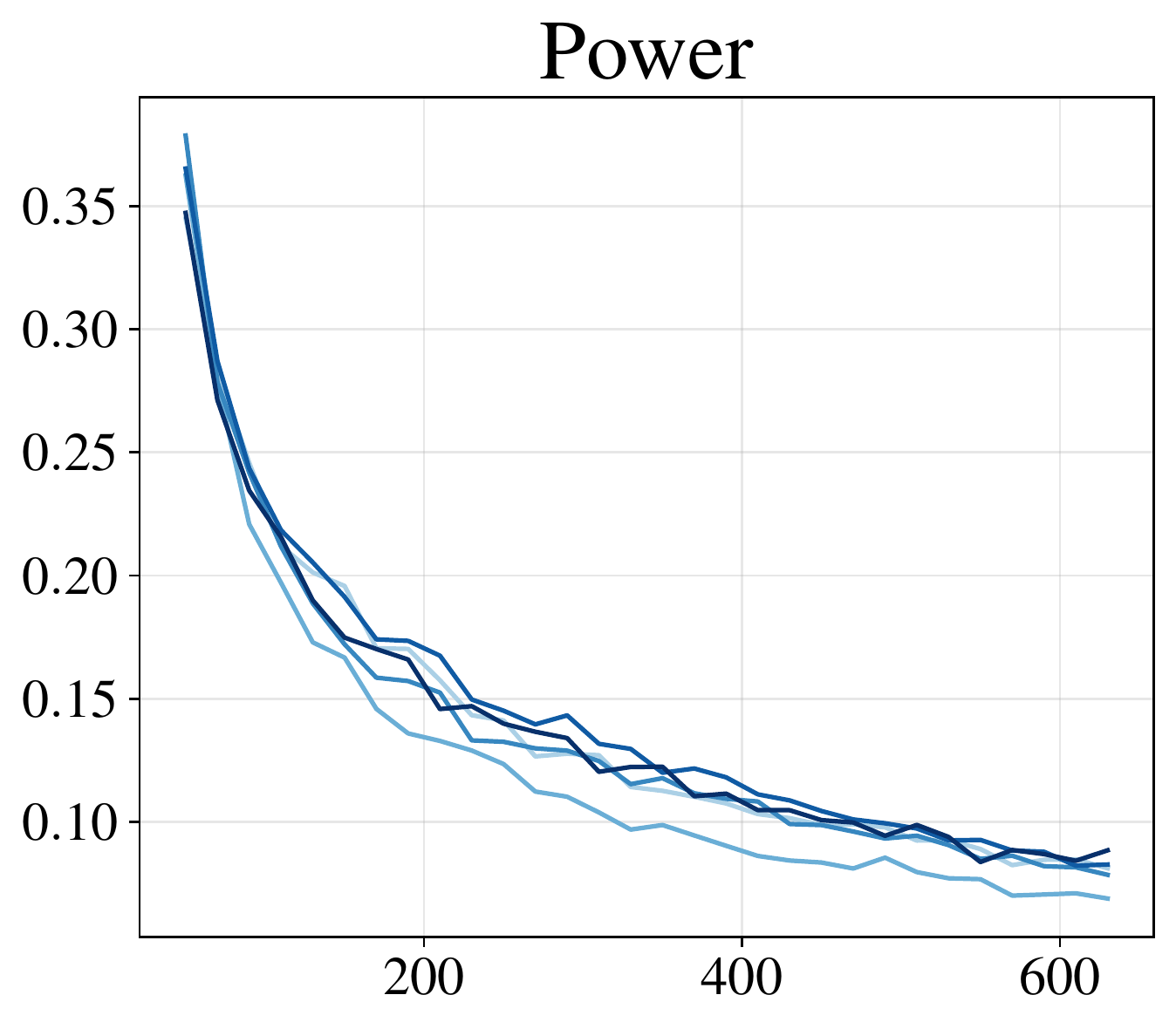}
    \end{subfigure} \\
    \begin{subfigure}[b]{0.33\textwidth}
        \centering
        \includegraphics[width=\linewidth]{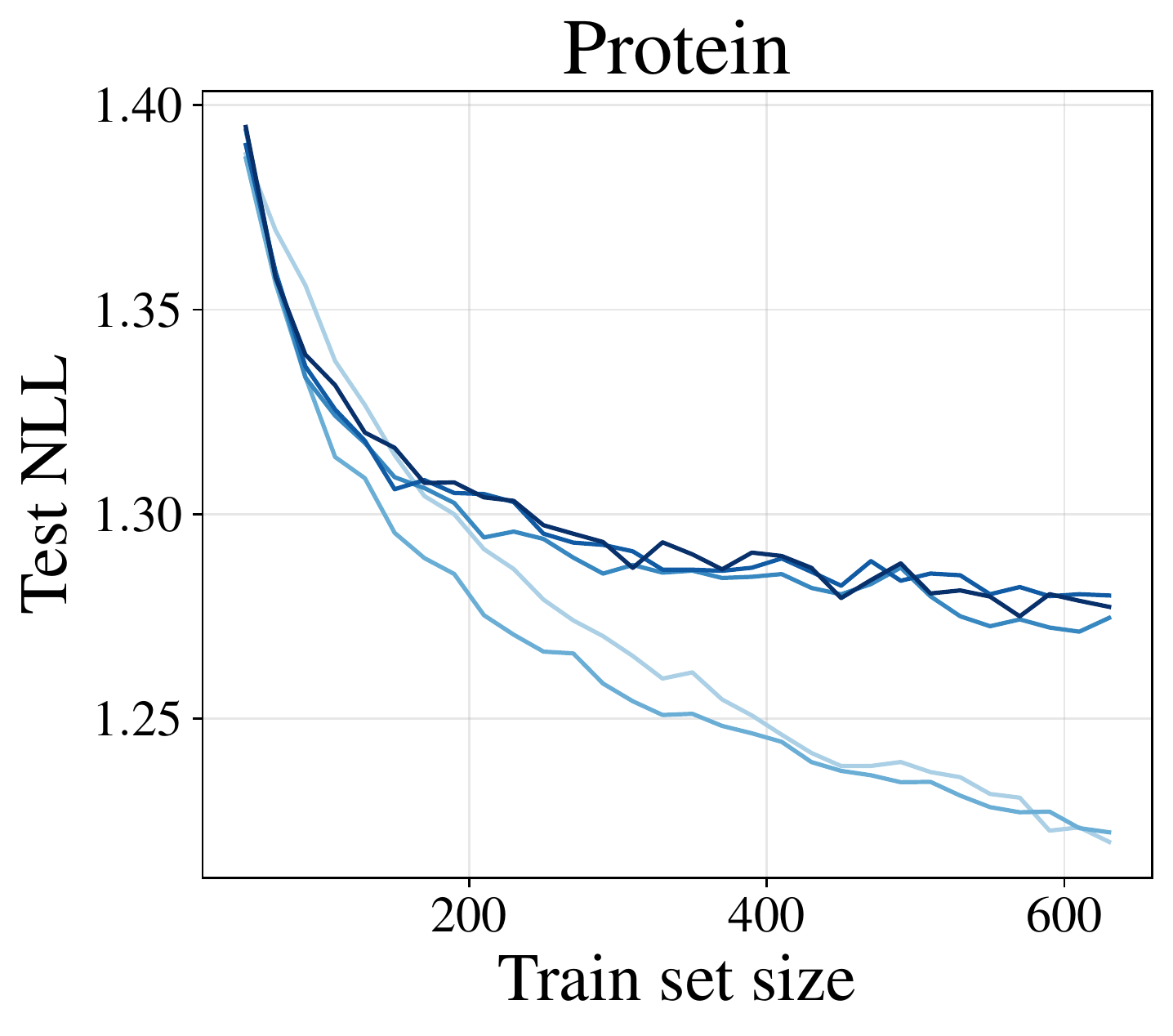}
    \end{subfigure} 
    \begin{subfigure}[b]{0.31\textwidth}
        \centering
        \includegraphics[width=\linewidth]{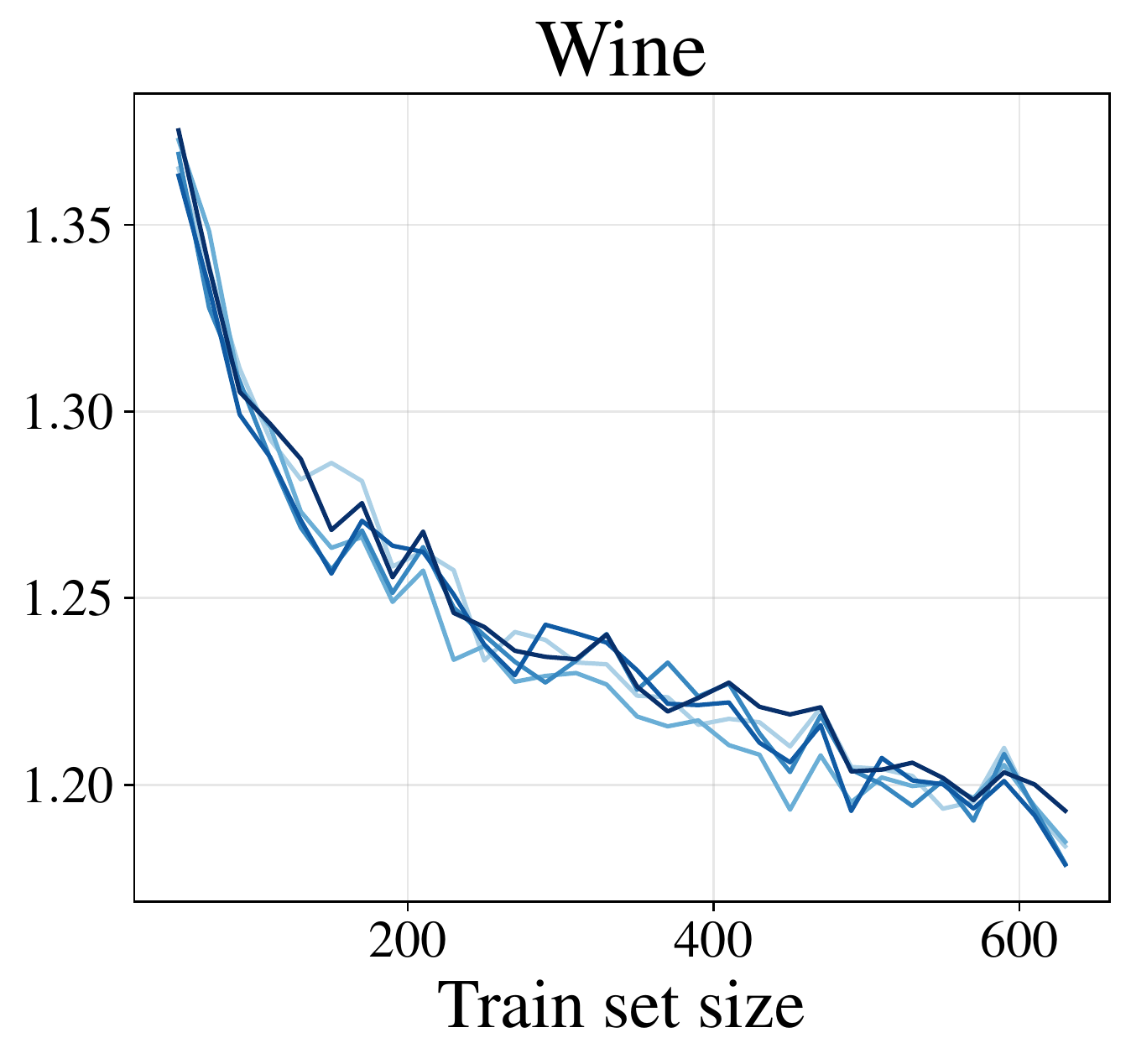}
    \end{subfigure} 
    \begin{subfigure}[b]{0.32\textwidth}
        \centering
        \includegraphics[width=\linewidth]{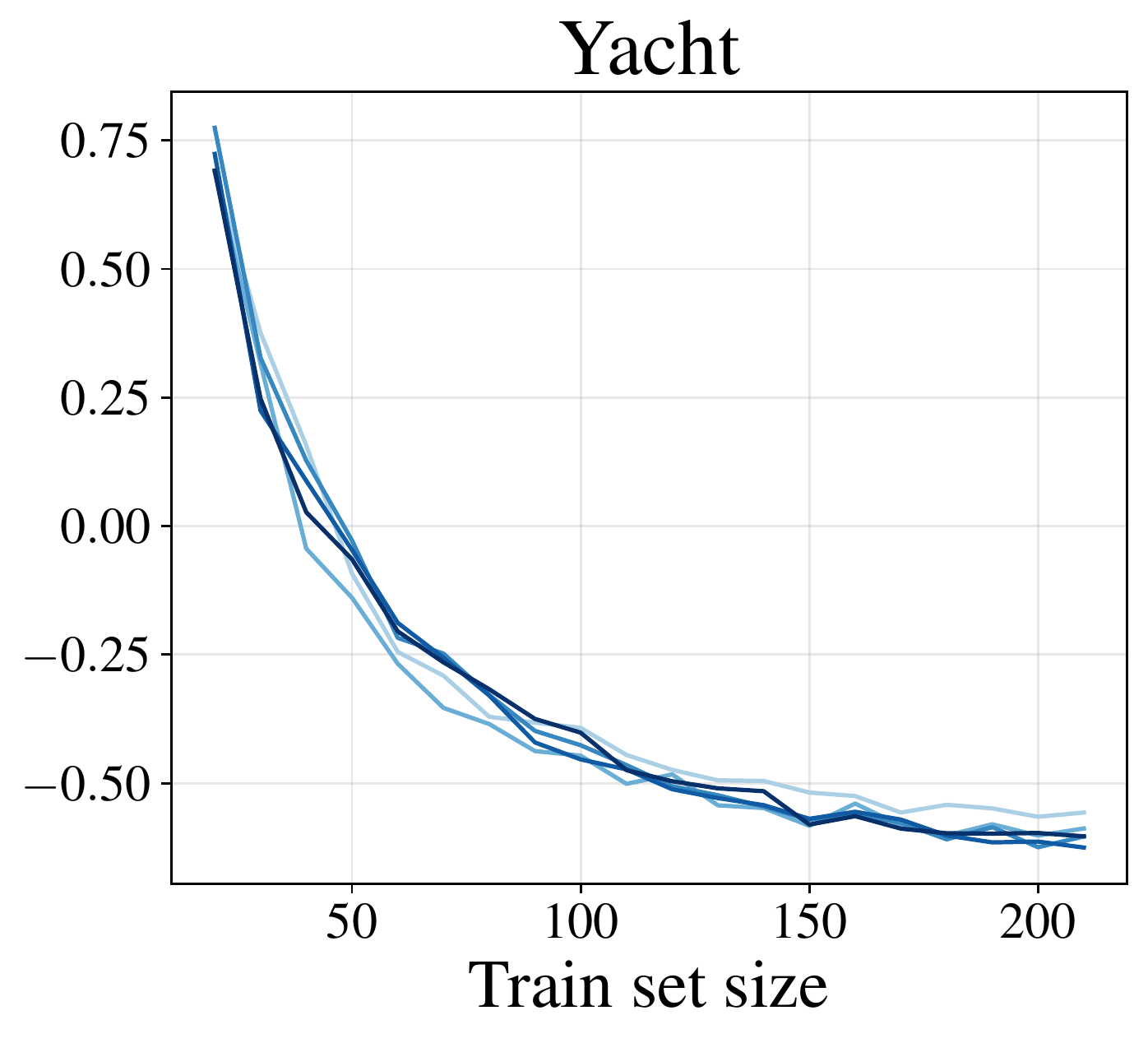}
\end{subfigure} \\
    \caption{Test NLL of DUNs using a stochastic relaxation of BALD. Different temperatures of the proposal distribution are compared. Means of 40 runs of the experiments are shown; standard deviations are not plotted for clarity.}
    \label{fig:res_stoch_bald_Ts_nll}
\end{figure}

\Cref{fig:res_stoch_bald_Ts_nll} shows the test NLL for DUNs using different temperatures $T$ for the proposal distribution. The magnitude of $T$ controls how deterministic the resulting sampling is---a larger $T$ corresponds to more certainly selecting the point with the highest BALD score, while a smaller $T$ is closer to uniform sampling. A temperature of $T=10$ yields the best performance for most datasets.

\FloatBarrier
\clearpage
\subsection{Uniform and decaying depth priors}

\begin{figure}[h] 
\centering
    \begin{subfigure}[b]{0.33\textwidth}
        \centering
        \includegraphics[width=\linewidth]{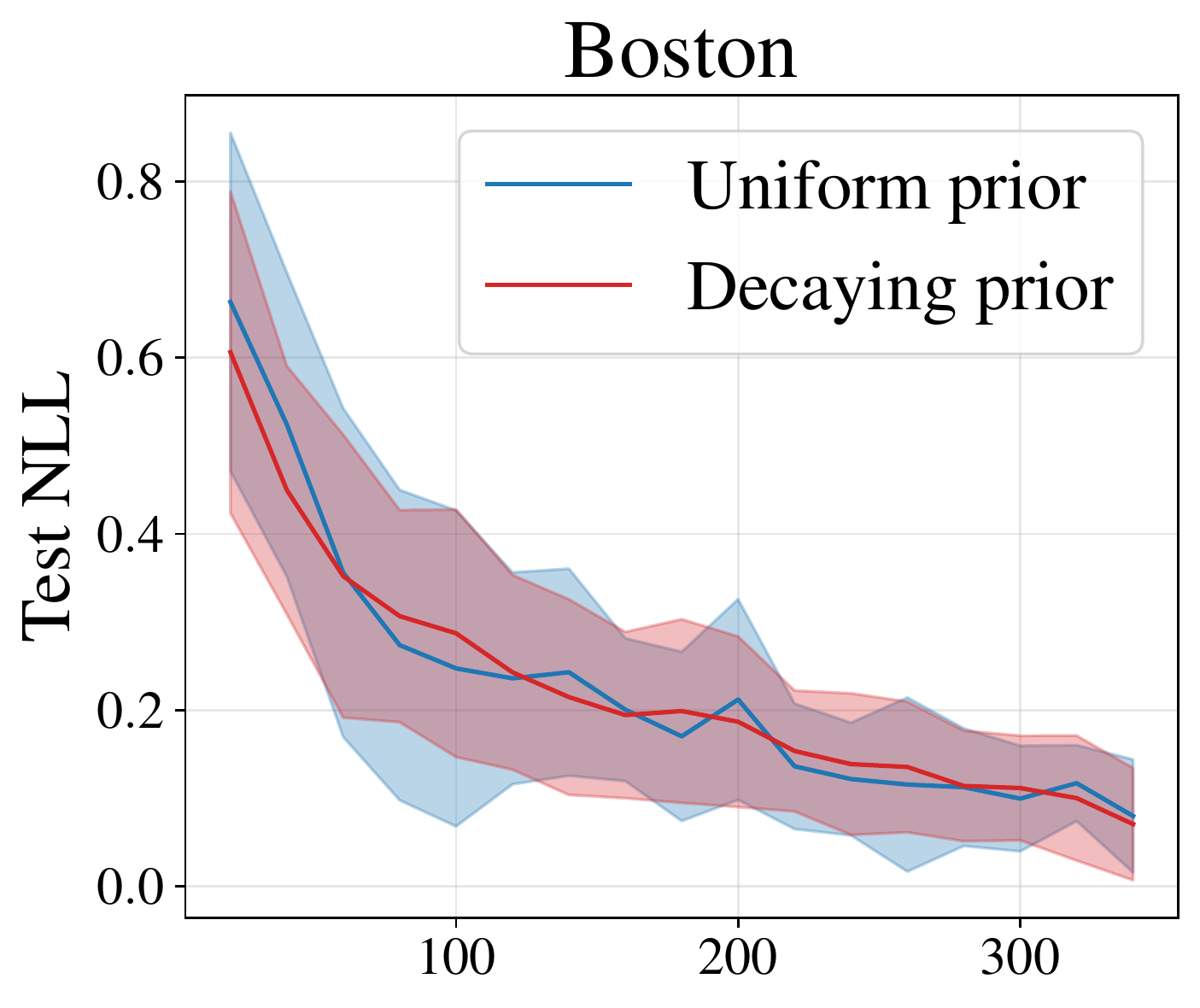}
    \end{subfigure}
    \begin{subfigure}[b]{0.31\textwidth}
        \centering
        \includegraphics[width=\linewidth]{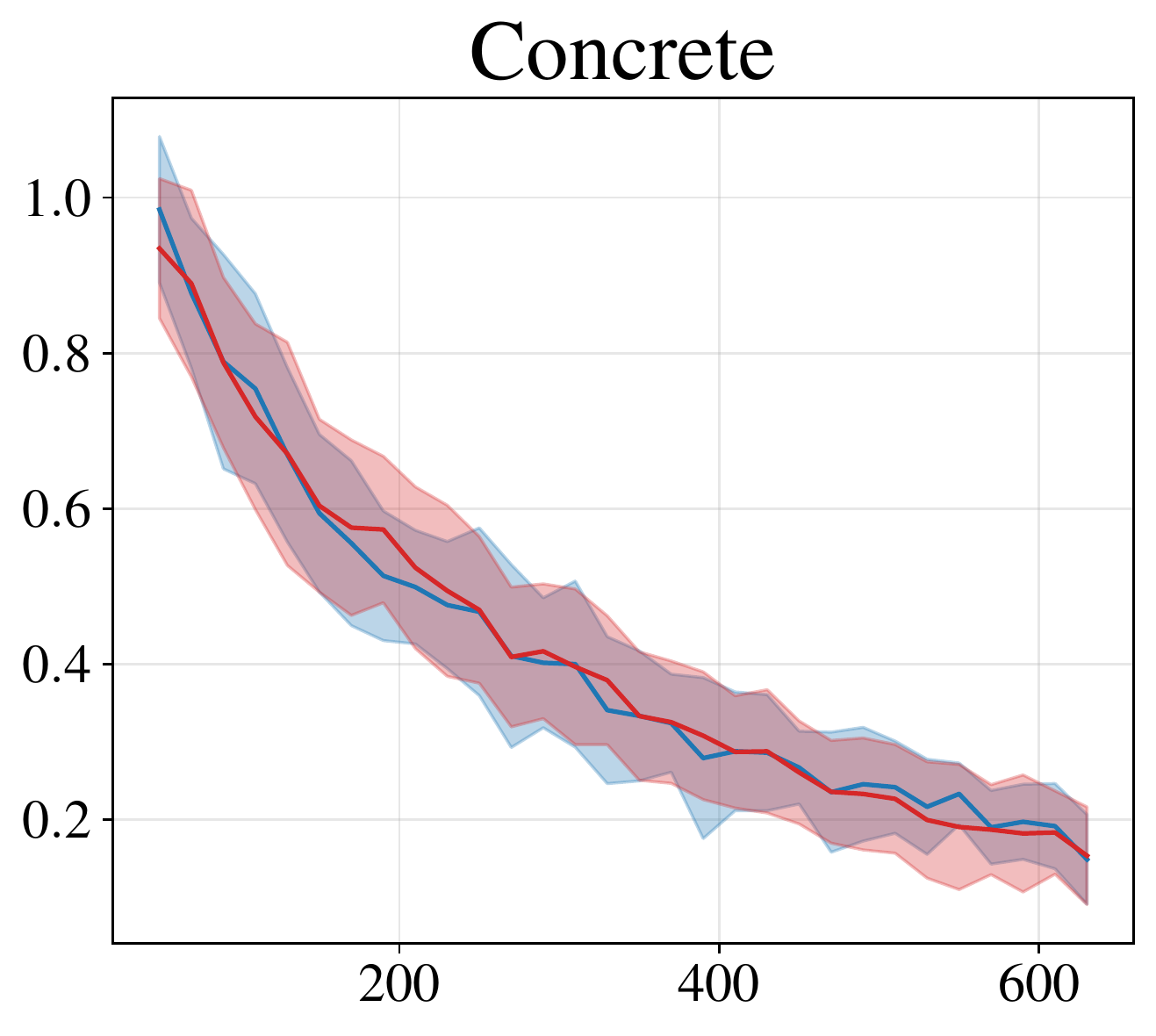}
    \end{subfigure} 
    \begin{subfigure}[b]{0.32\textwidth}
        \centering
        \includegraphics[width=\linewidth]{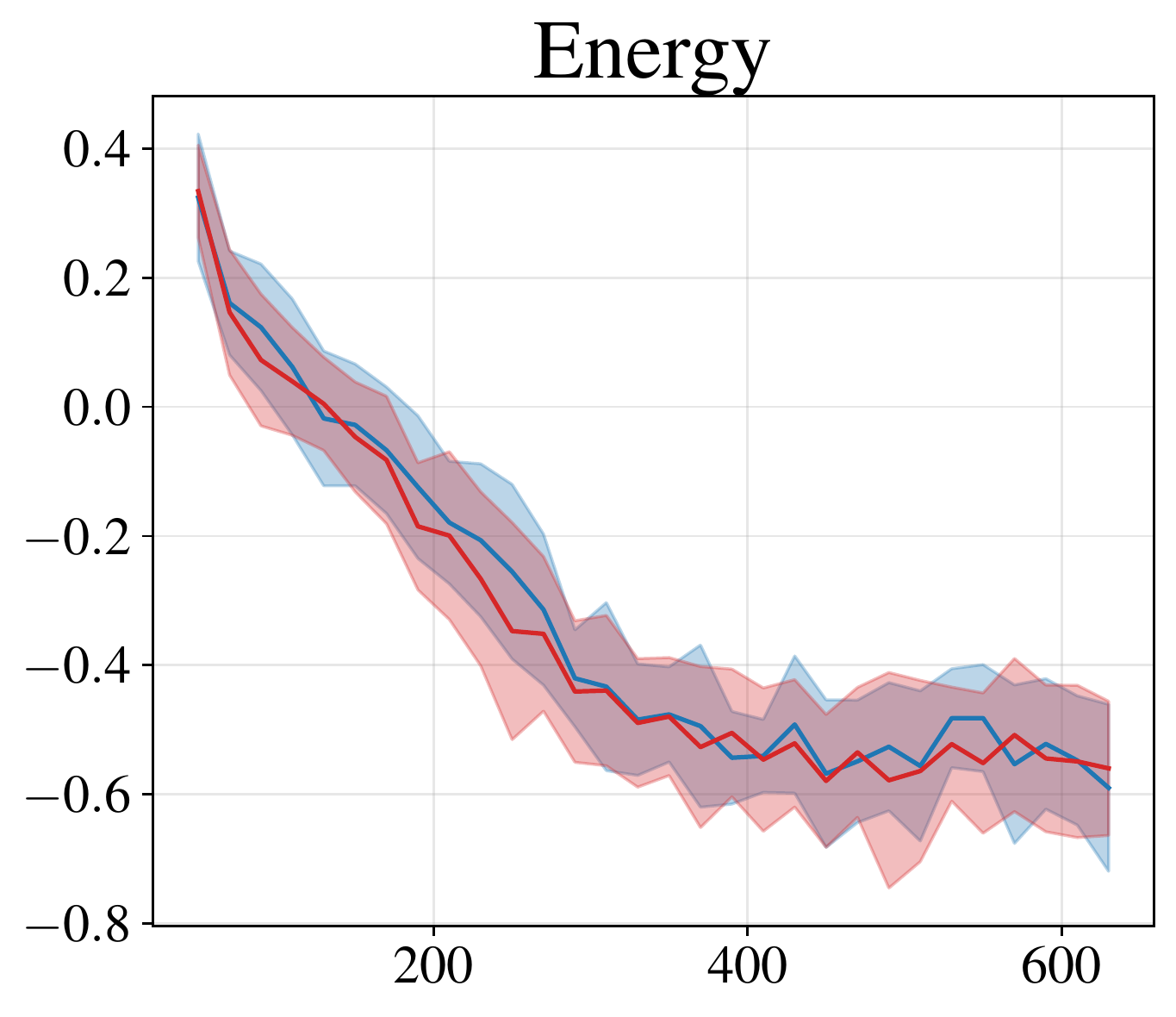} 
    \end{subfigure} \\
    \begin{subfigure}[b]{0.33\textwidth}
        \centering
        \includegraphics[width=\linewidth]{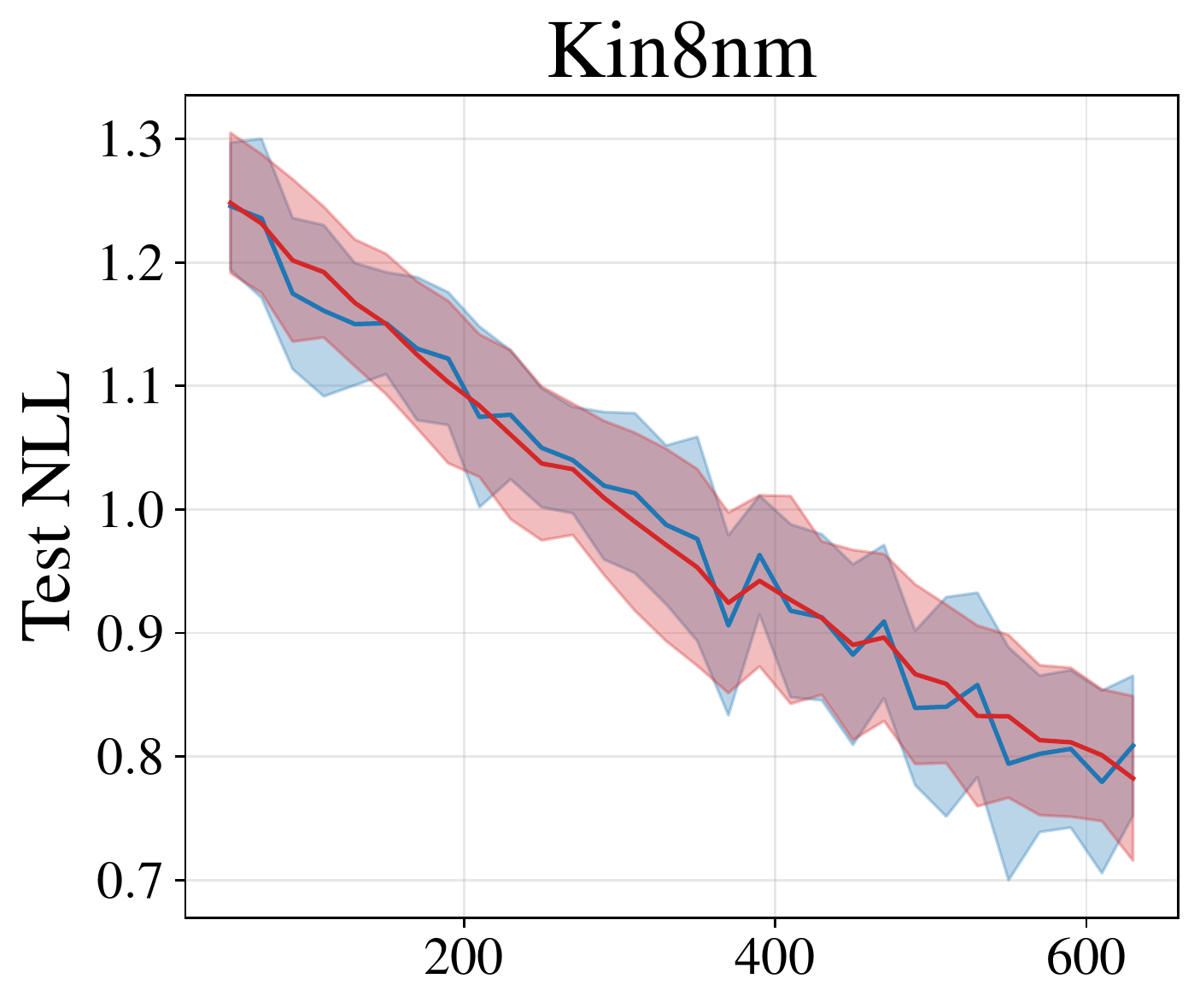}
    \end{subfigure} 
    \begin{subfigure}[b]{0.32\textwidth}
        \centering
        \includegraphics[width=\linewidth]{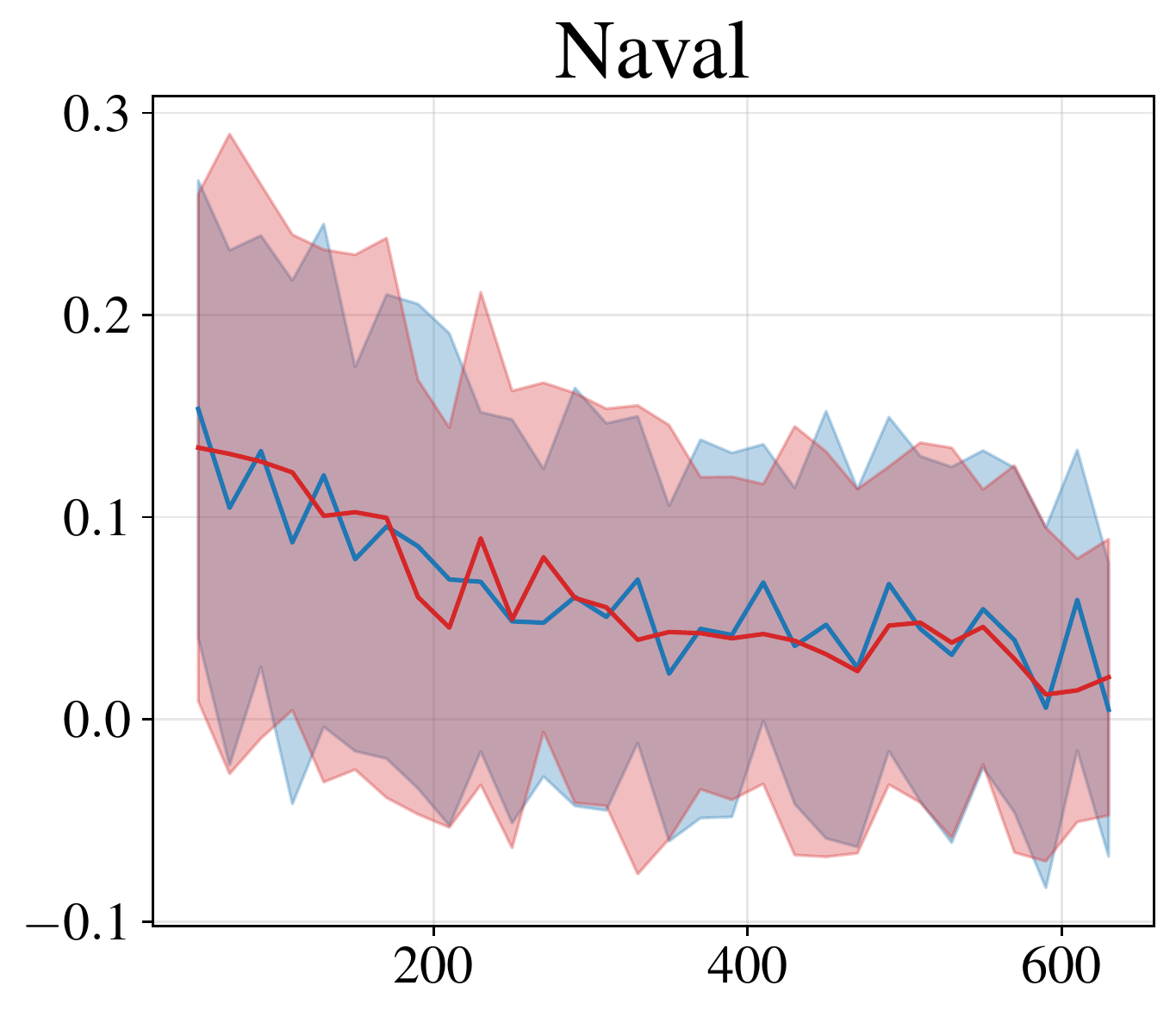}
    \end{subfigure} 
    \begin{subfigure}[b]{0.31\textwidth}
        \centering
        \includegraphics[width=\linewidth]{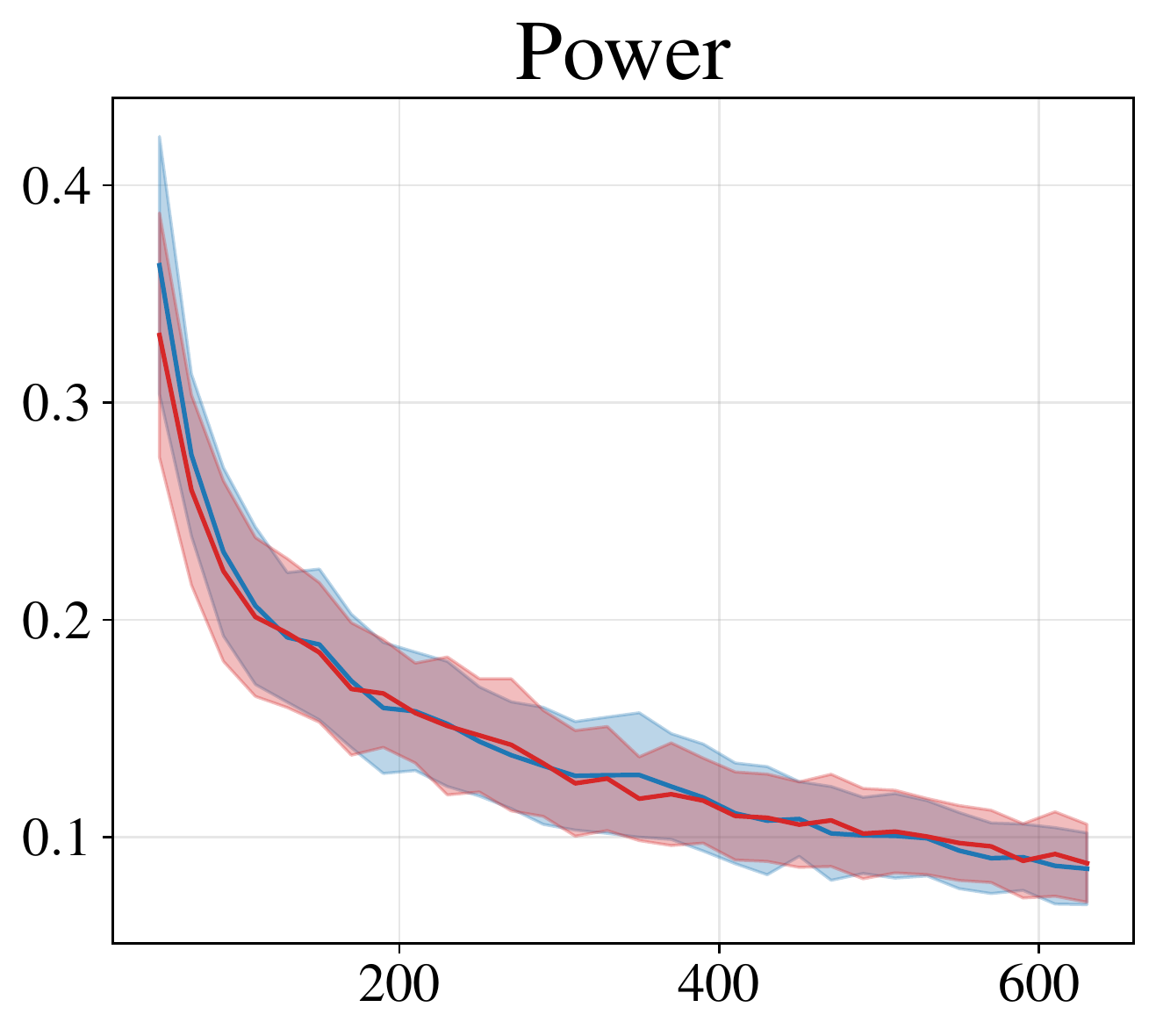}
    \end{subfigure} \\
    \begin{subfigure}[b]{0.332\textwidth}
        \centering
        \includegraphics[width=\linewidth]{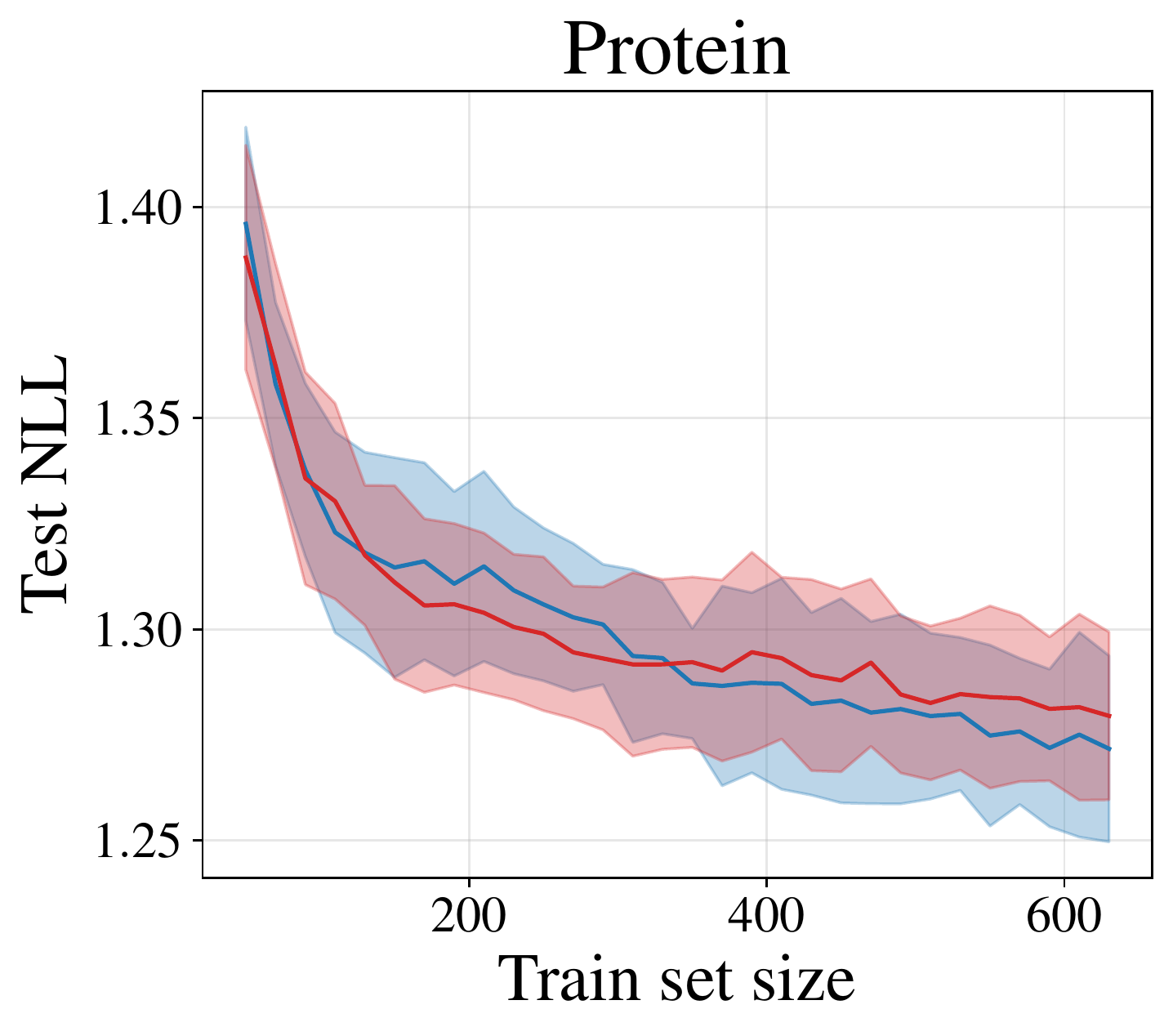}
    \end{subfigure} 
    \begin{subfigure}[b]{0.312\textwidth}
        \centering
        \includegraphics[width=\linewidth]{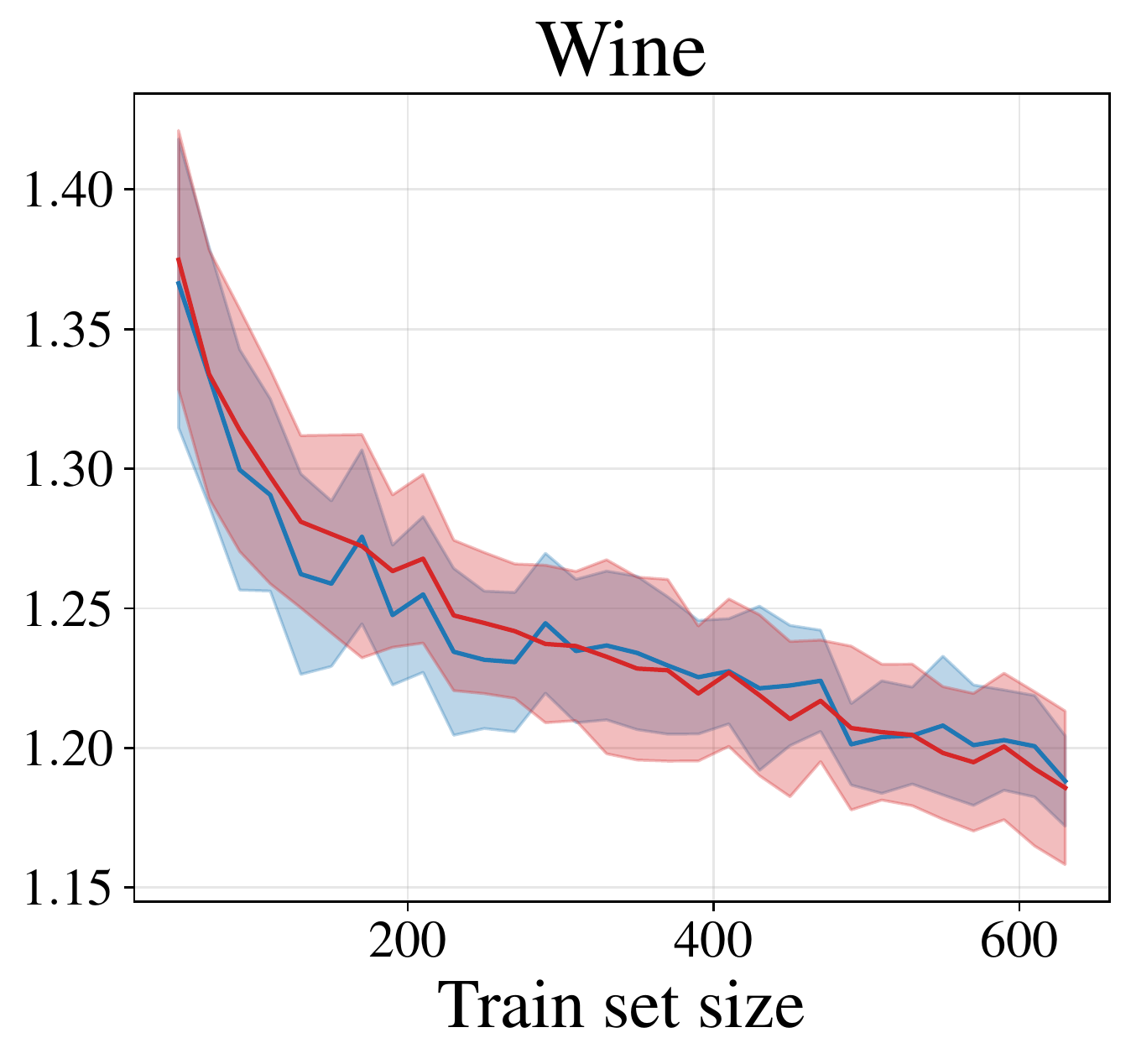}
    \end{subfigure} 
    \begin{subfigure}[b]{0.317\textwidth}
        \centering
        \includegraphics[width=\linewidth]{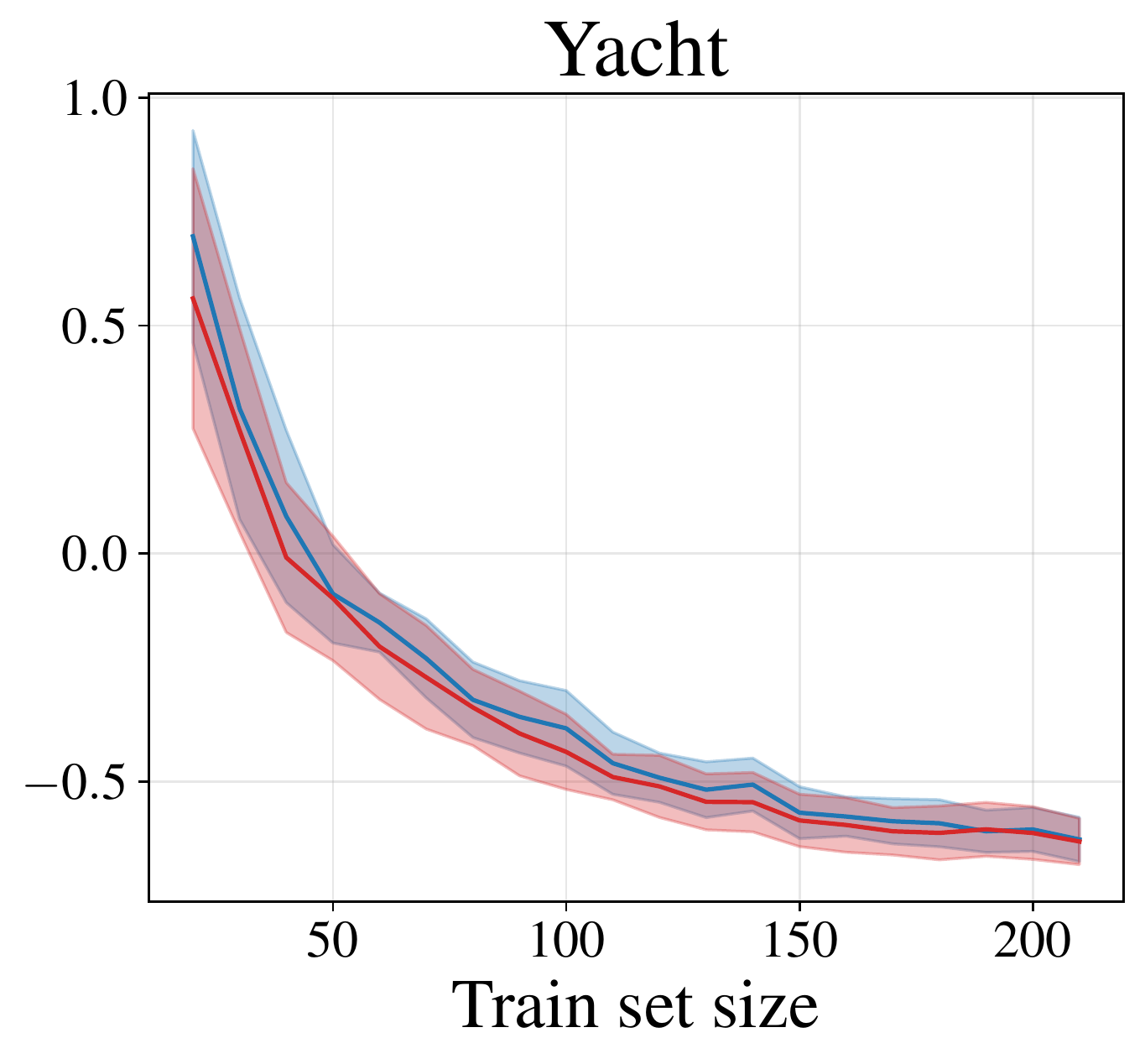}
\end{subfigure} \\
    \caption{Test NLL vs. number of training points for DUNs. A \textbf{\textcolor{mplblue}{uniform}} depth prior is compared to a \textbf{\textcolor{mplred}{decaying}} prior.}
    \label{fig:res_reg_prior_decay_nll}
\end{figure}

\Cref{fig:res_reg_prior_decay_nll} compares DUNs' performance with a uniform prior over depth to that with exponentially decaying prior probabilities over depth. Much larger prior probabilities for earlier layers relative to later layers in the network should encourage a posterior that prefers shallower networks. This may, in turn, help to reduce overfitting in the early stages of active learning. In practice, no difference in performance is observed between the two prior distributions.  

\clearpage
\subsection{Alternative depths for MCDO and MFVI}

\Cref{fig:app_depths} compares the performance of DUNs to MCDO, MFVI and regular NNs trained with stochastic gradient descent (SGD), using different depth networks for the baseline methods. For each dataset and baseline method, a single hidden layer network and a network with depth equal to the highest posterior probability depth found by the DUN for that dataset are shown.

\begin{figure}[h] 
\centering
    \begin{subfigure}{0.33\textwidth}
        \centering
        \includegraphics[width=\linewidth]{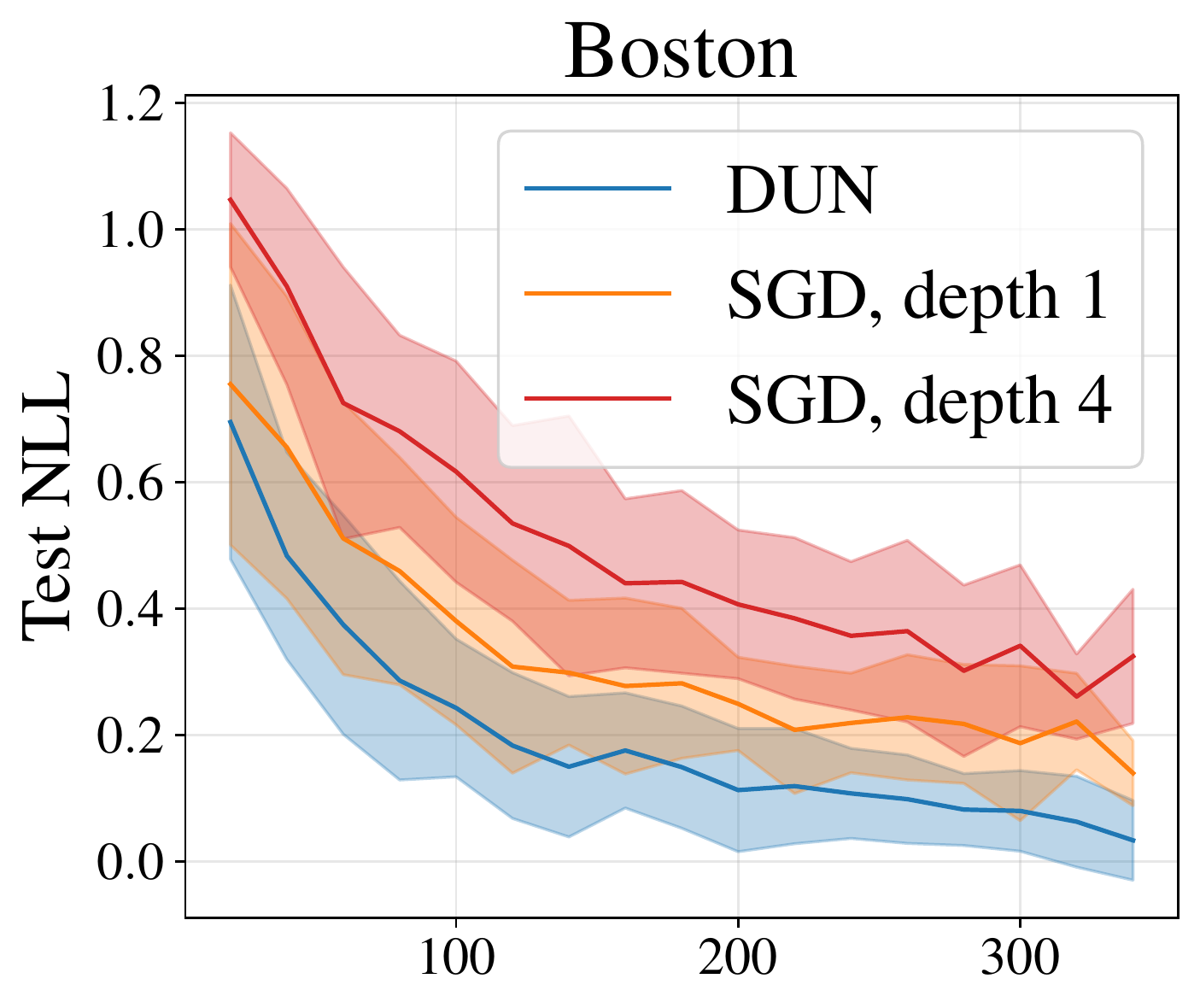}
    \end{subfigure}
    \begin{subfigure}{0.32\textwidth}
        \centering
        \includegraphics[width=\linewidth]{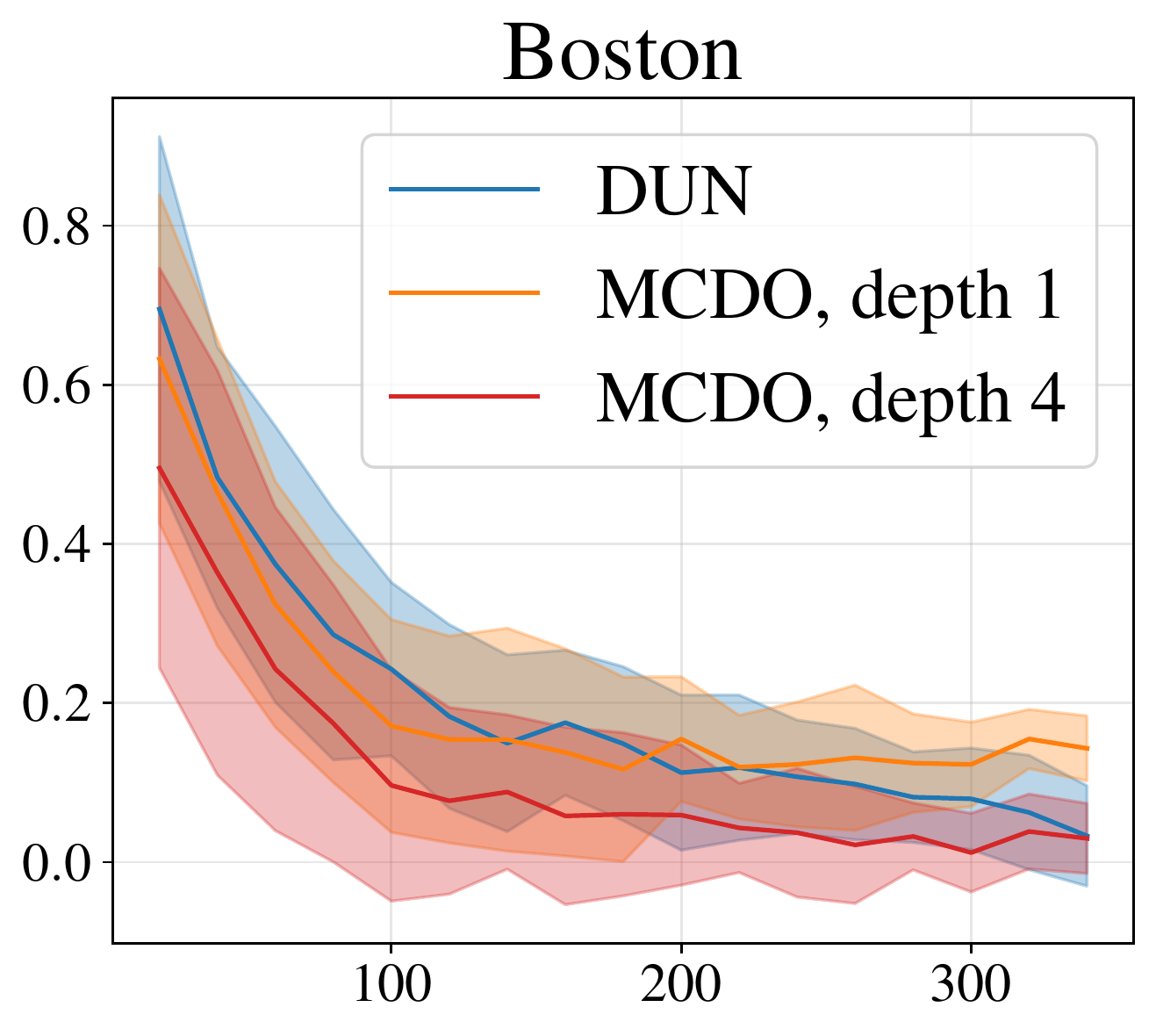}
    \end{subfigure} 
    \begin{subfigure}{0.32\textwidth}
        \centering
        \includegraphics[width=\linewidth]{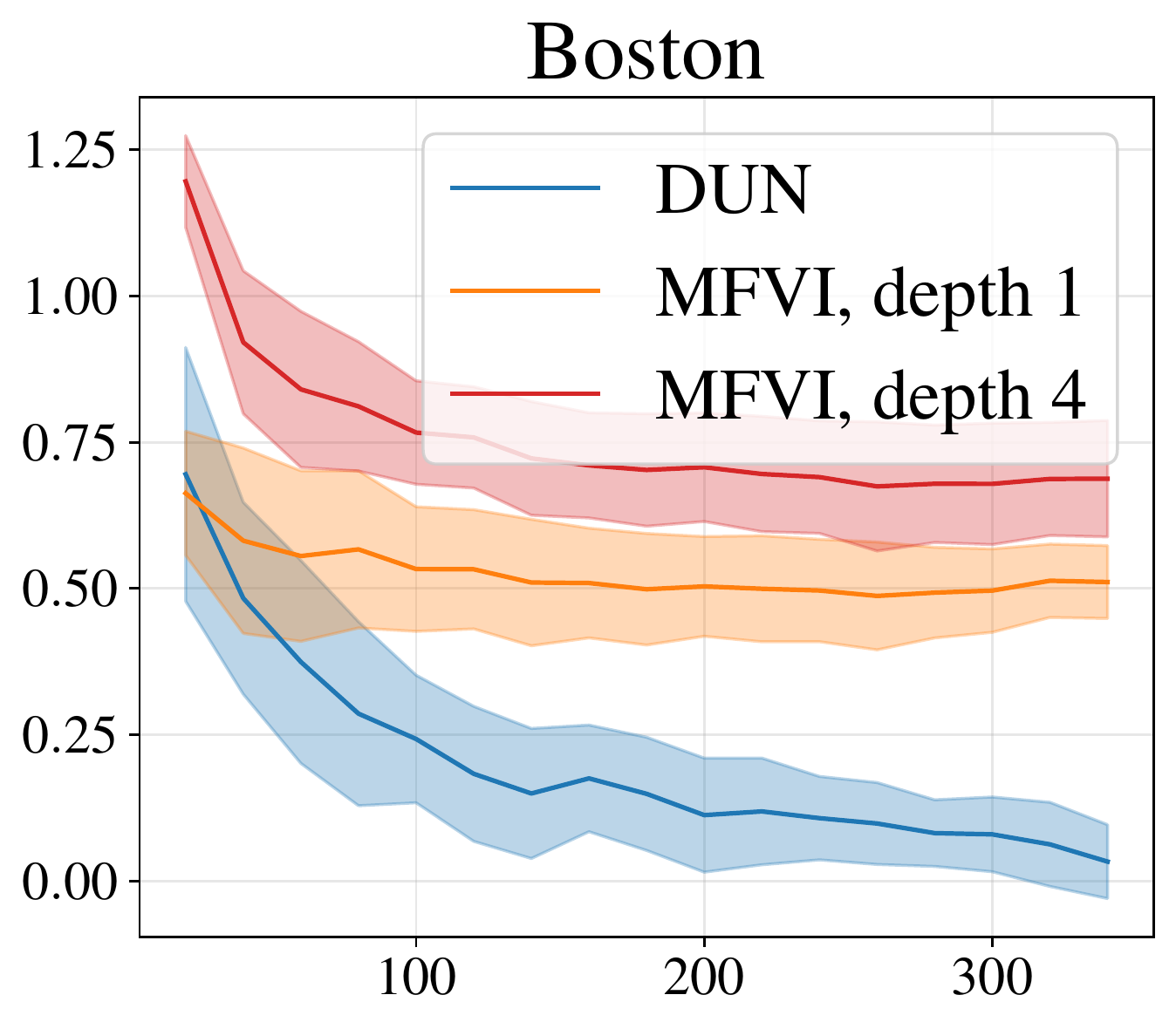} 
    \end{subfigure} \\
    \begin{subfigure}{0.33\textwidth}
        \centering
        \includegraphics[width=\linewidth]{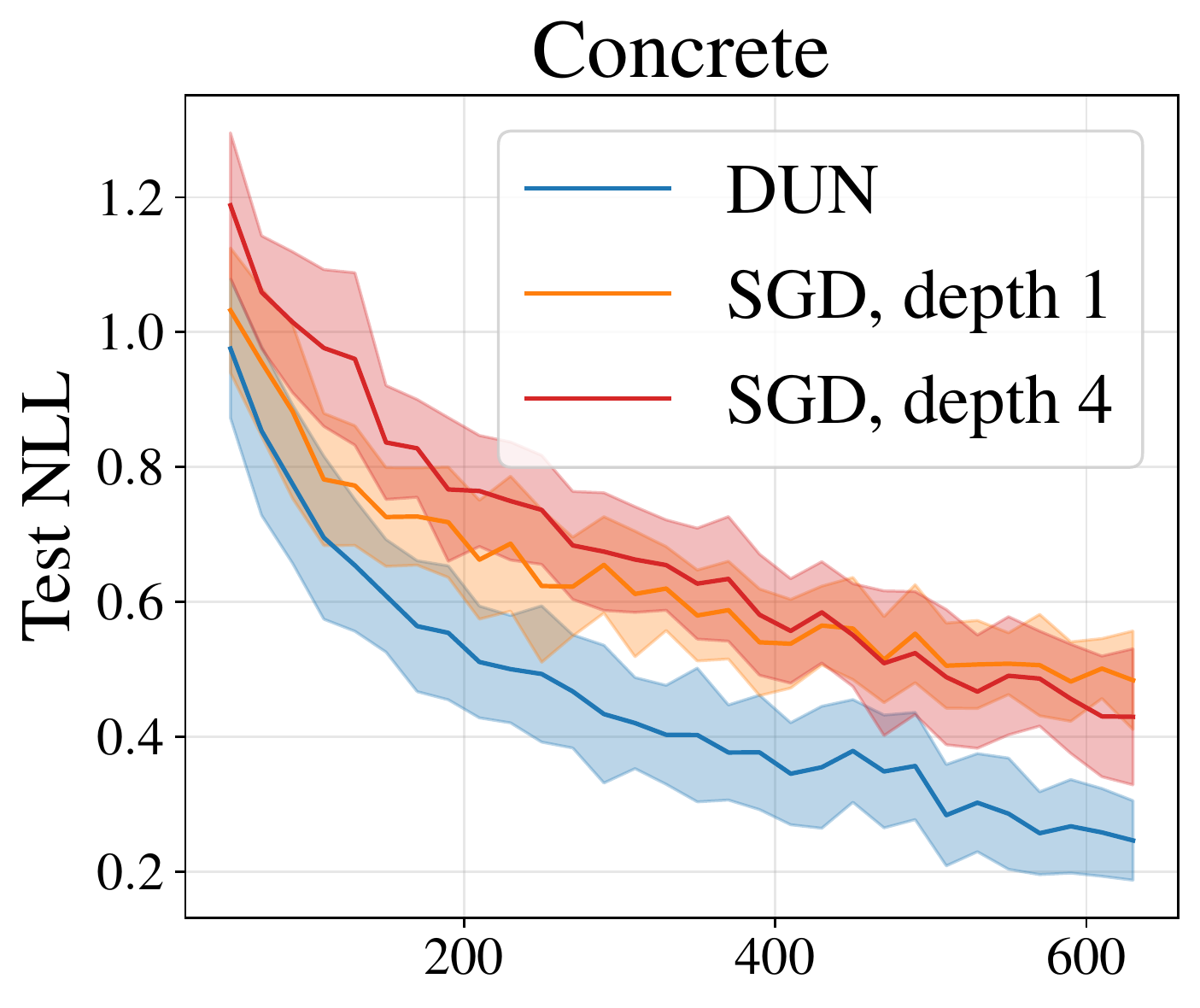}
    \end{subfigure}
    \begin{subfigure}{0.32\textwidth}
        \centering
        \includegraphics[width=\linewidth]{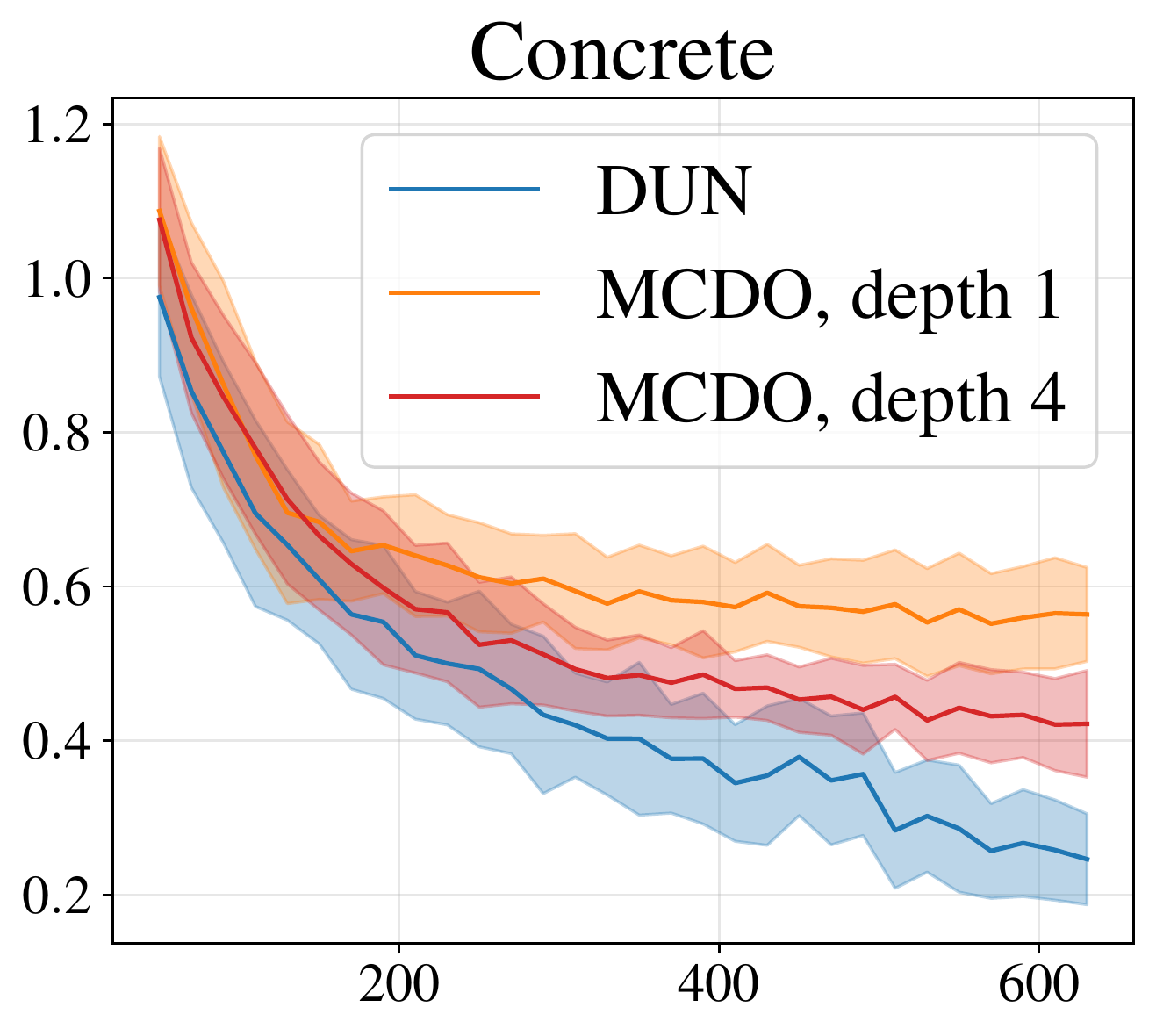}
    \end{subfigure} 
    \begin{subfigure}{0.32\textwidth}
        \centering
        \includegraphics[width=\linewidth]{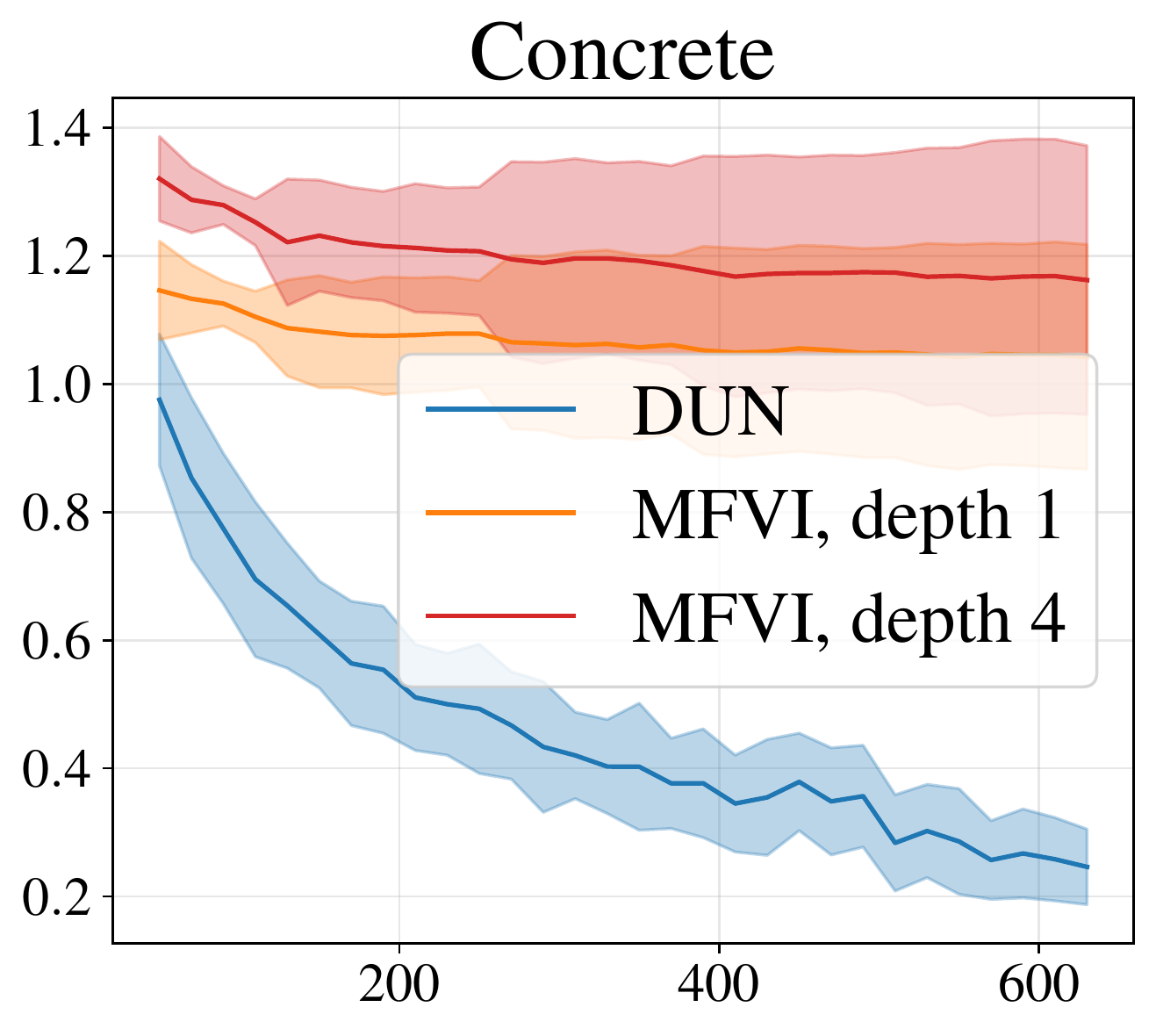} 
    \end{subfigure} \\  
    \begin{subfigure}{0.33\textwidth}
        \centering
        \includegraphics[width=\linewidth]{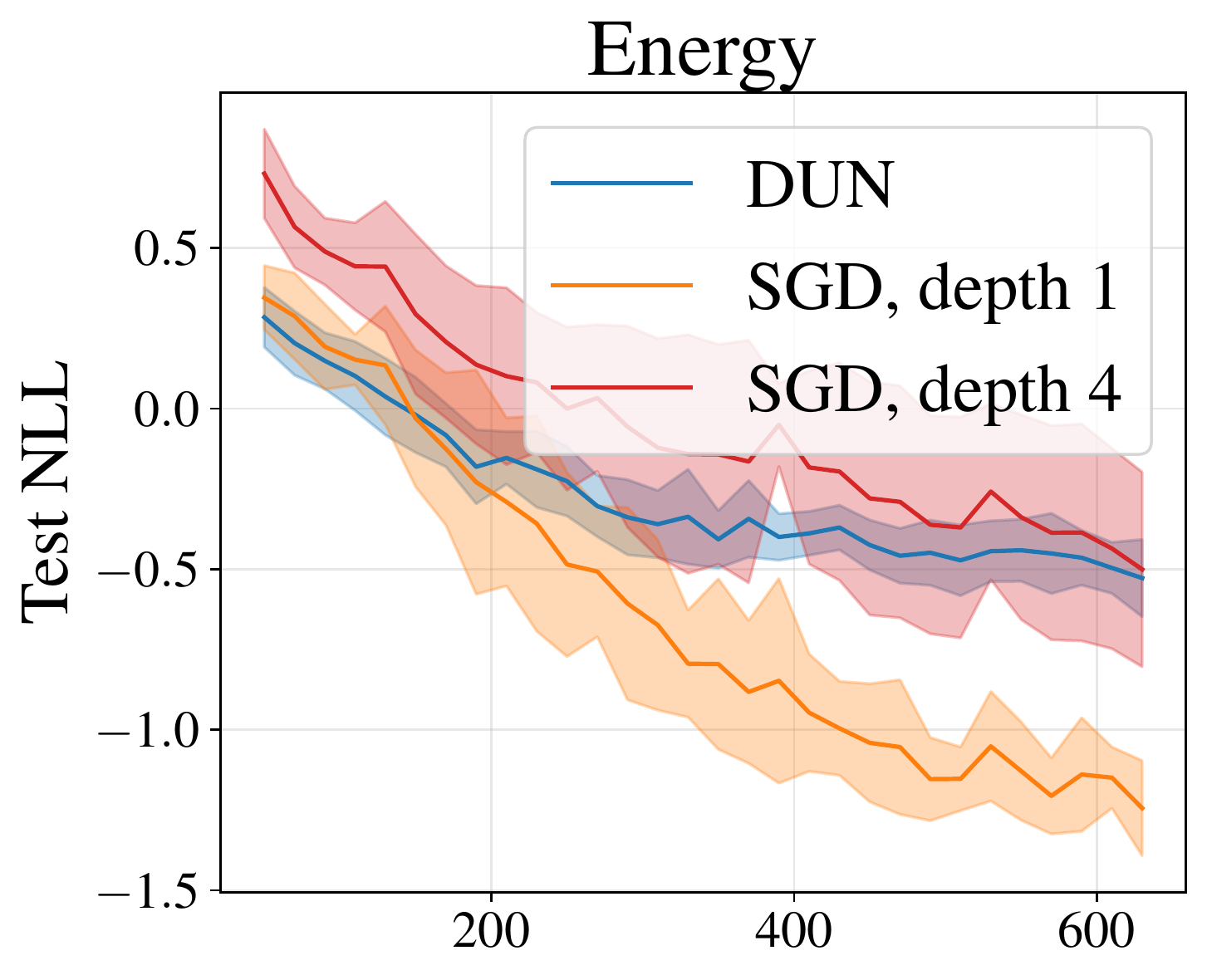}
    \end{subfigure}
    \begin{subfigure}{0.32\textwidth}
        \centering
        \includegraphics[width=\linewidth]{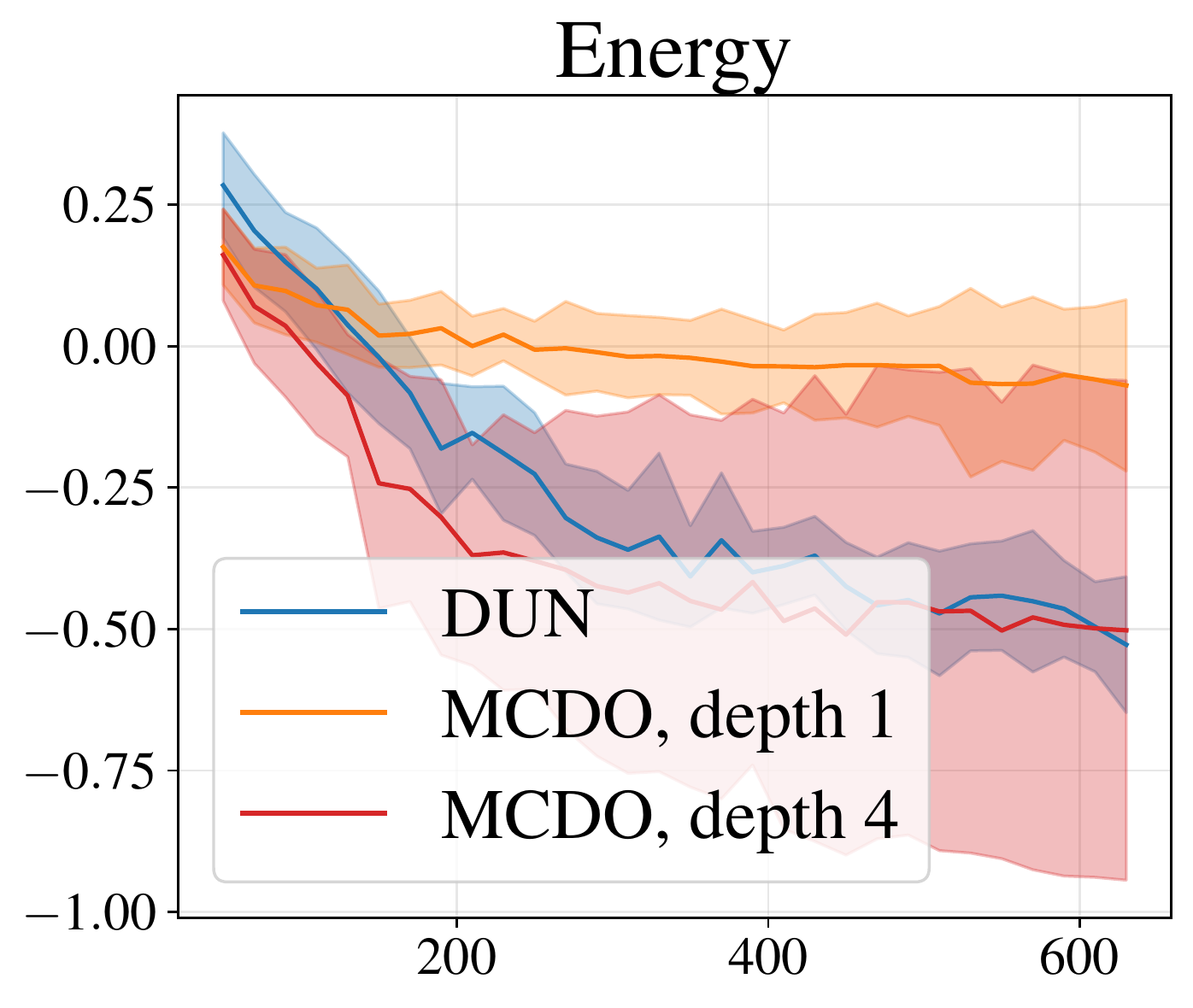}
    \end{subfigure} 
    \begin{subfigure}{0.32\textwidth}
        \centering
        \includegraphics[width=\linewidth]{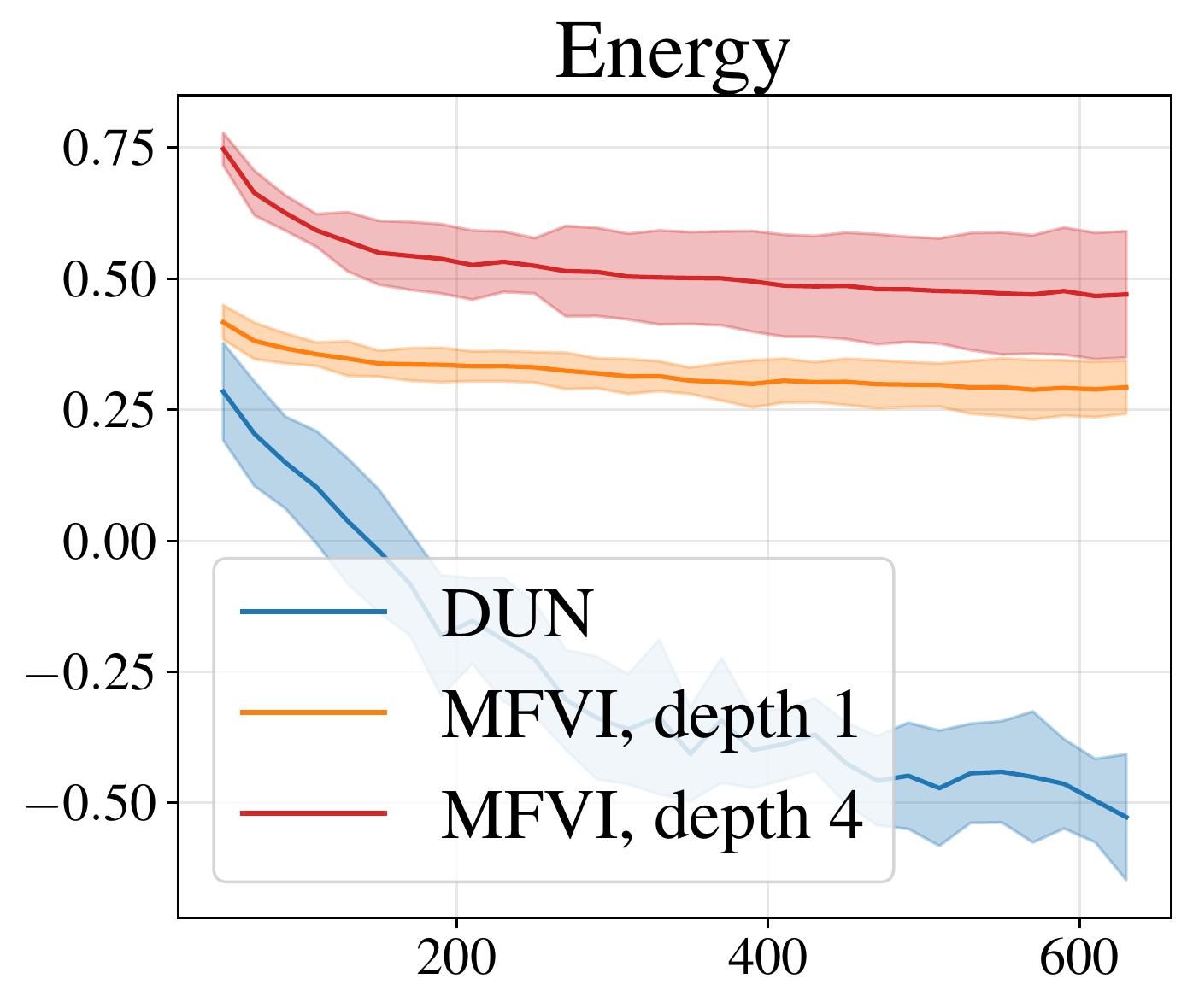} 
    \end{subfigure} \\
    \begin{subfigure}{0.33\textwidth}
        \centering
        \includegraphics[width=\linewidth]{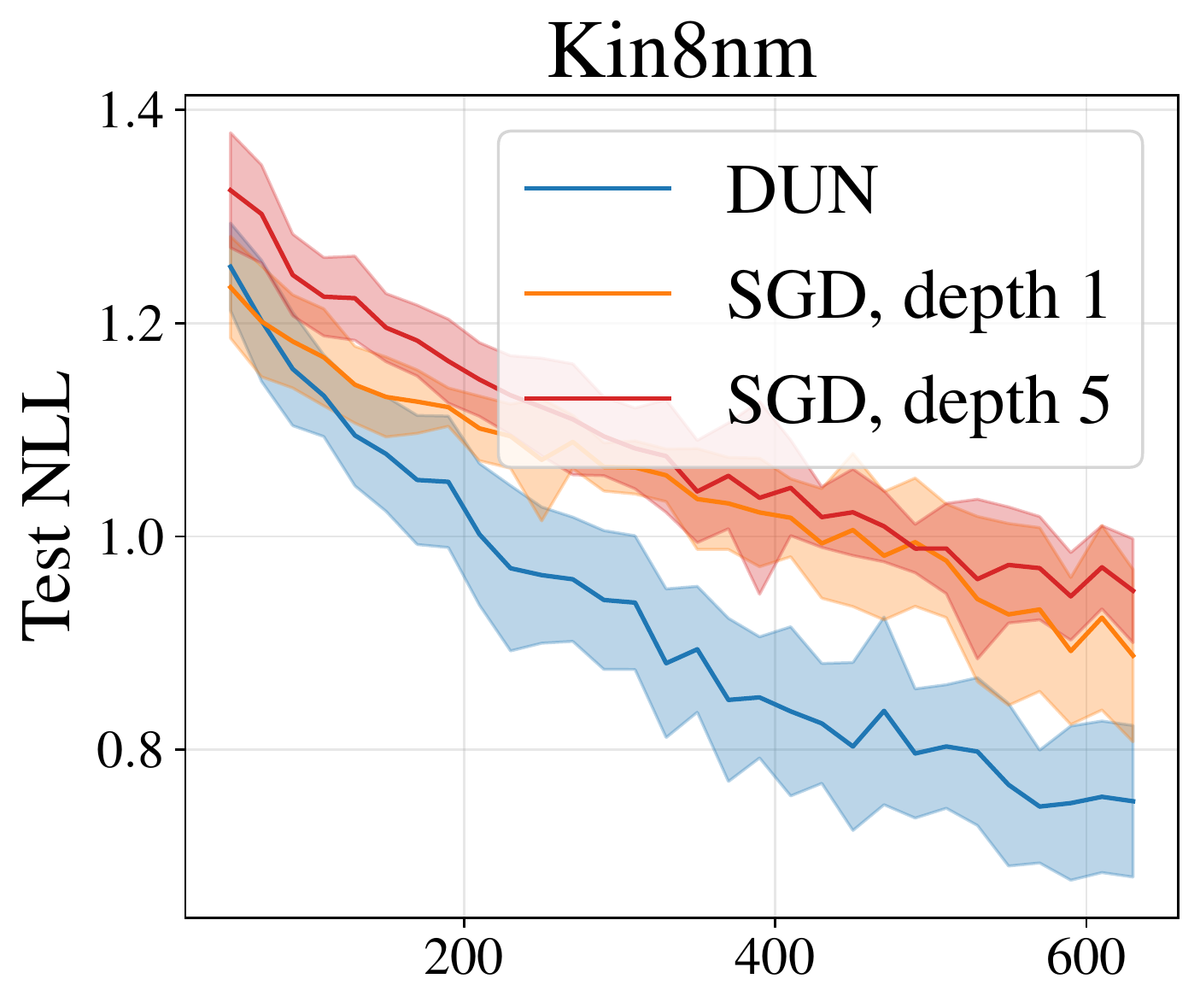}
    \end{subfigure}
    \begin{subfigure}{0.32\textwidth}
        \centering
        \includegraphics[width=\linewidth]{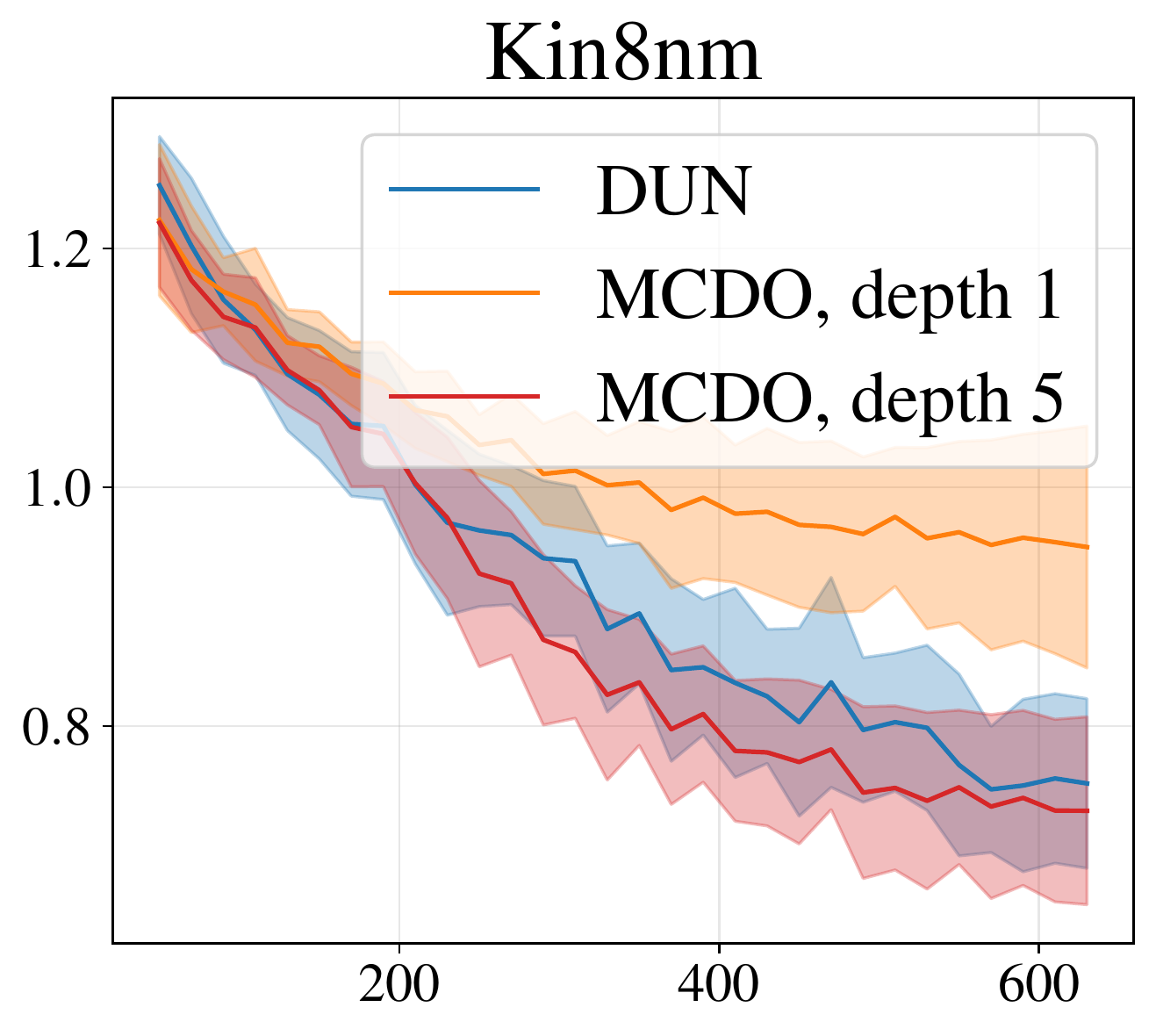}
    \end{subfigure} 
    \begin{subfigure}{0.32\textwidth}
        \centering
        \includegraphics[width=\linewidth]{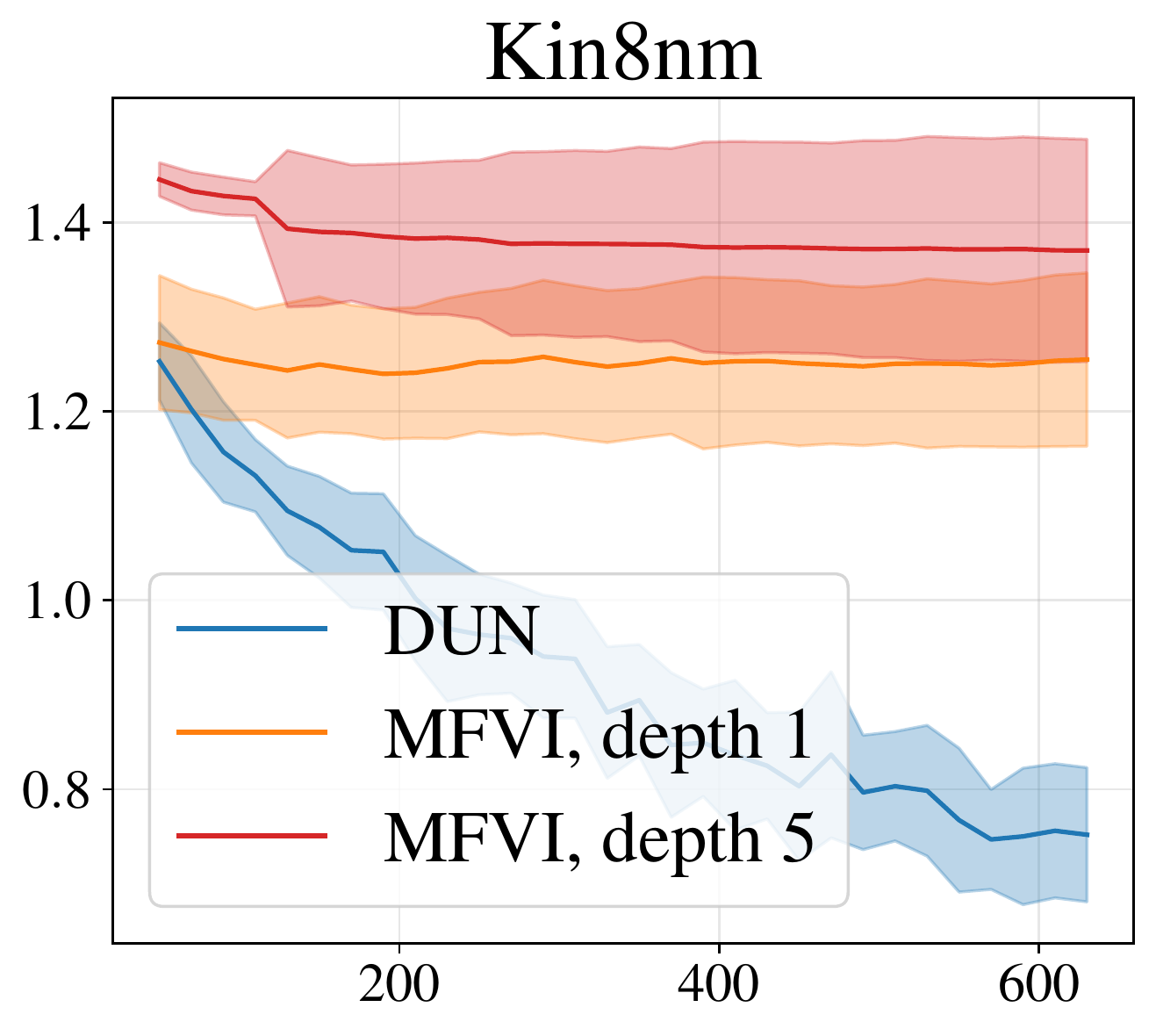} 
    \end{subfigure} \\
    \begin{subfigure}{0.33\textwidth}
        \centering
        \includegraphics[width=\linewidth]{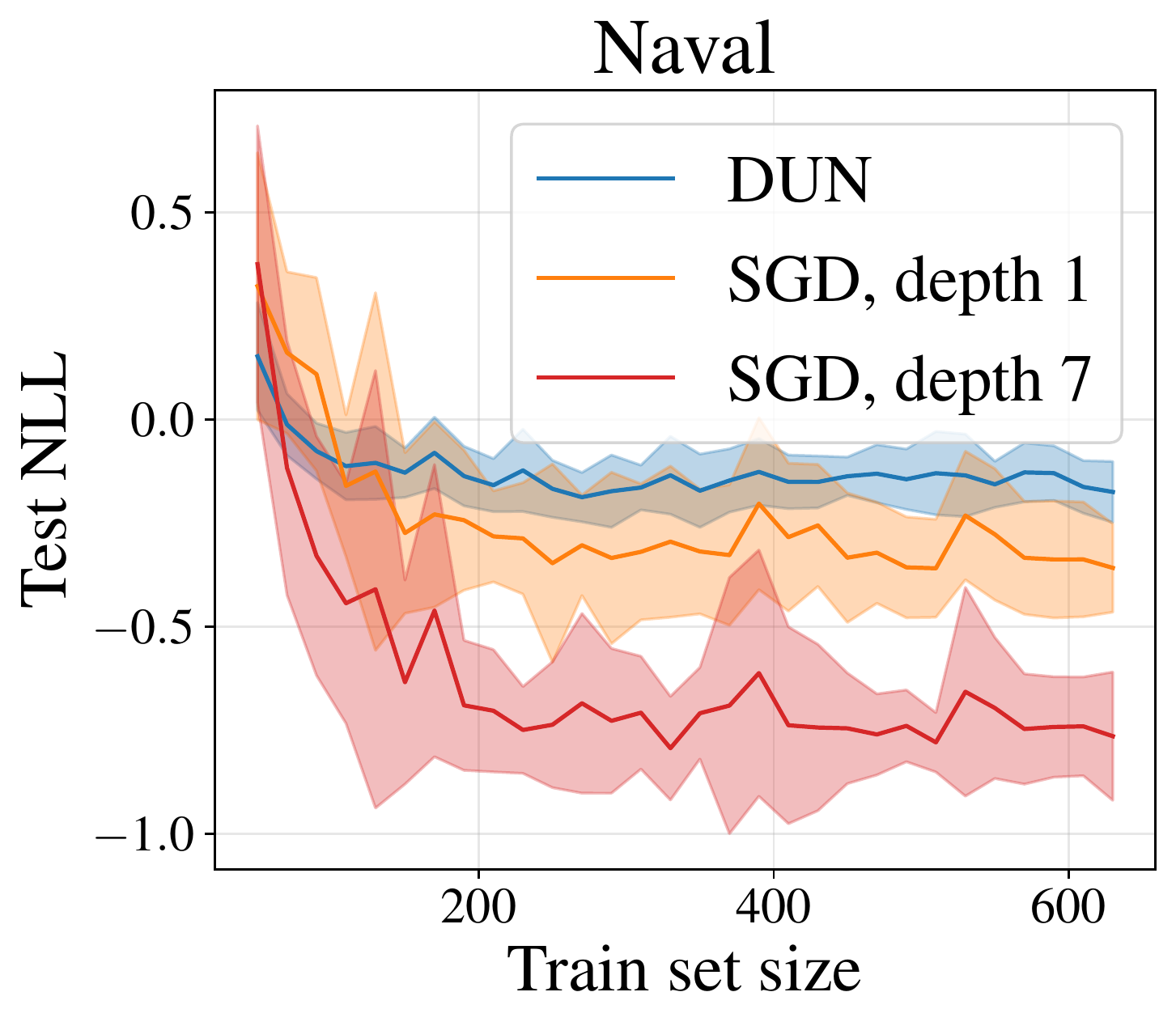}
    \end{subfigure}
    \begin{subfigure}{0.32\textwidth}
        \centering
        \includegraphics[width=\linewidth]{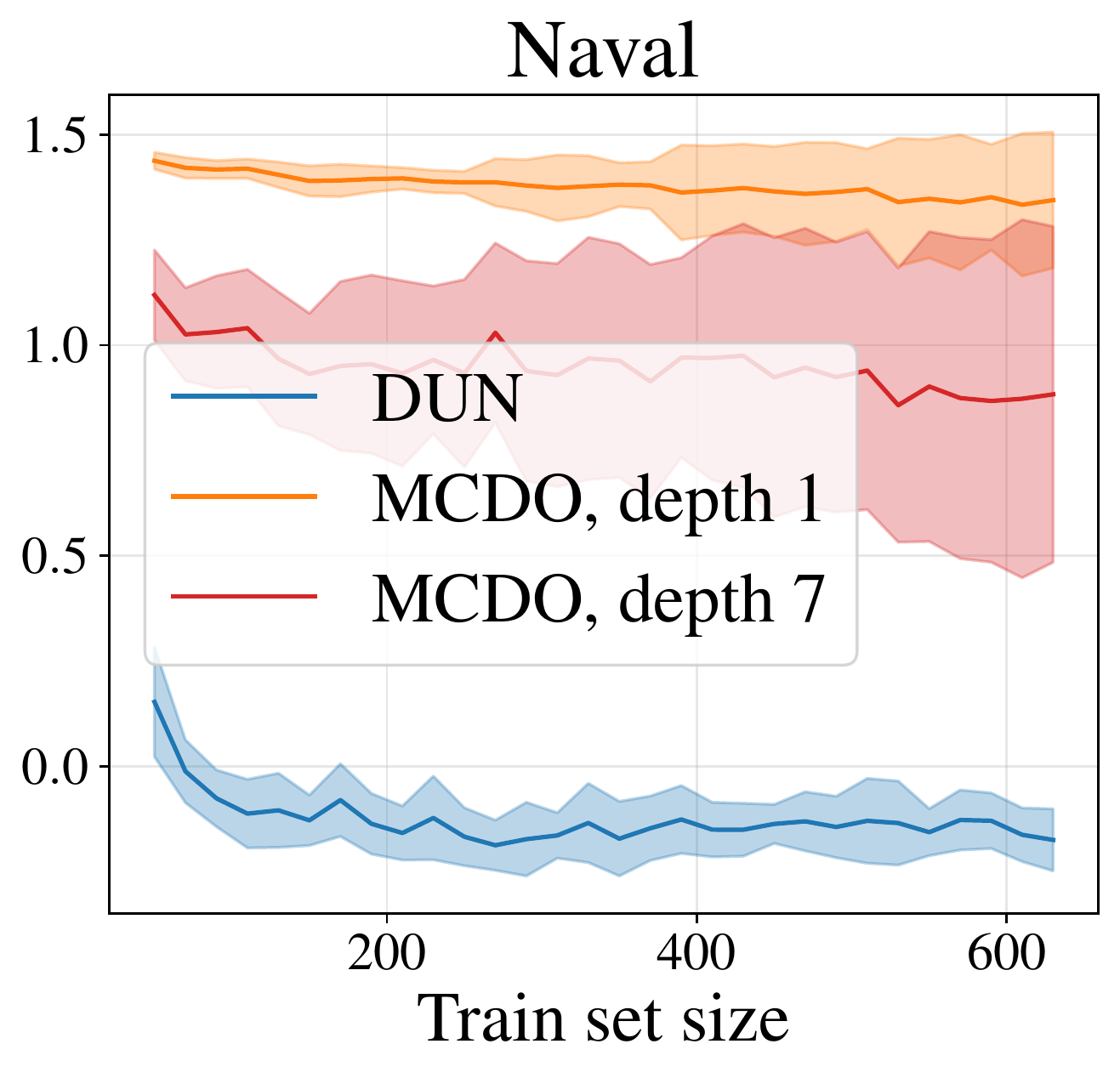}
    \end{subfigure} 
    \begin{subfigure}{0.32\textwidth}
        \centering
        \includegraphics[width=\linewidth]{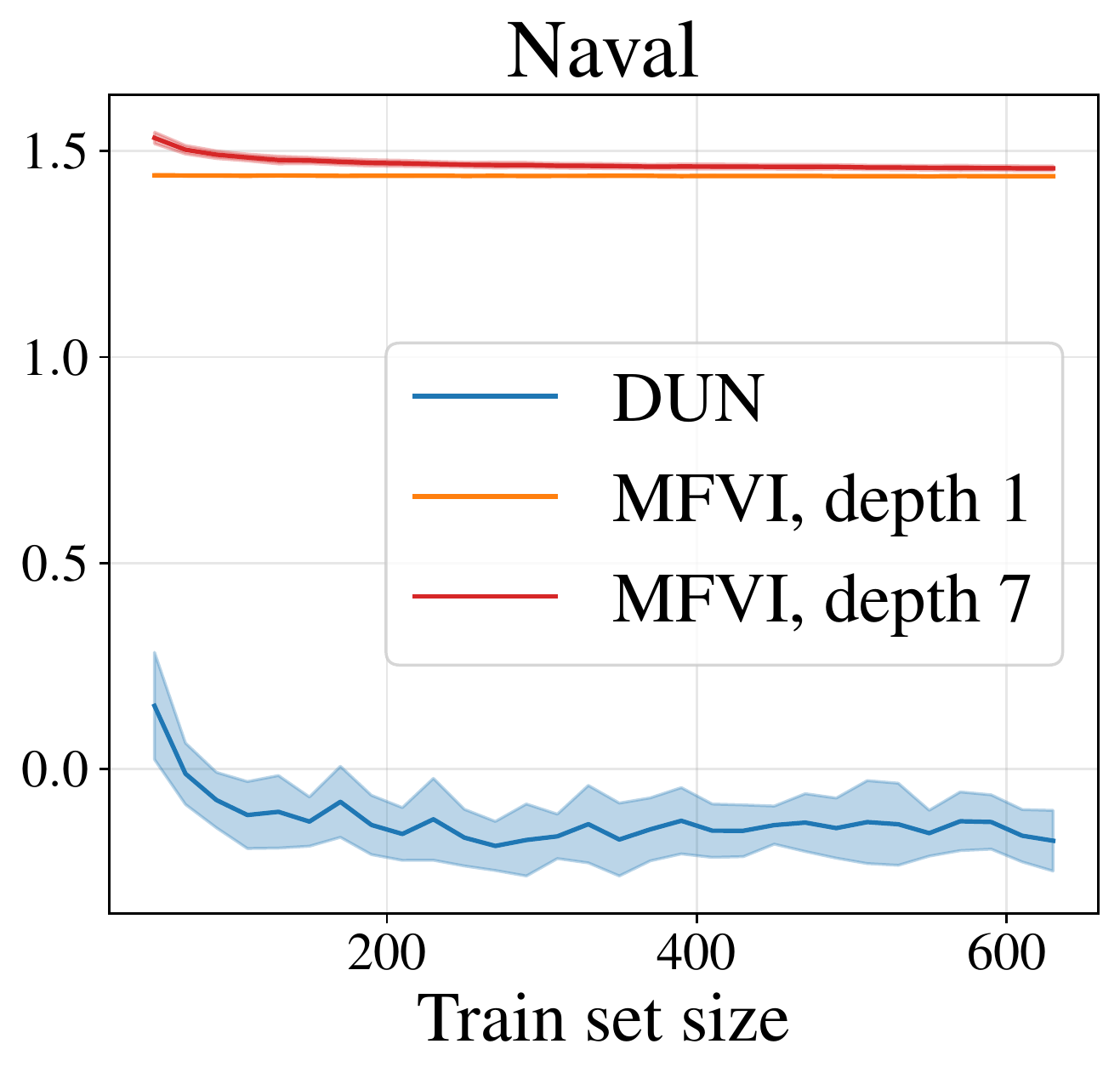} 
    \end{subfigure} \\
    \end{figure}%
    \begin{figure}[ht]\ContinuedFloat
    \begin{subfigure}{0.33\textwidth}
        \centering
        \includegraphics[width=\linewidth]{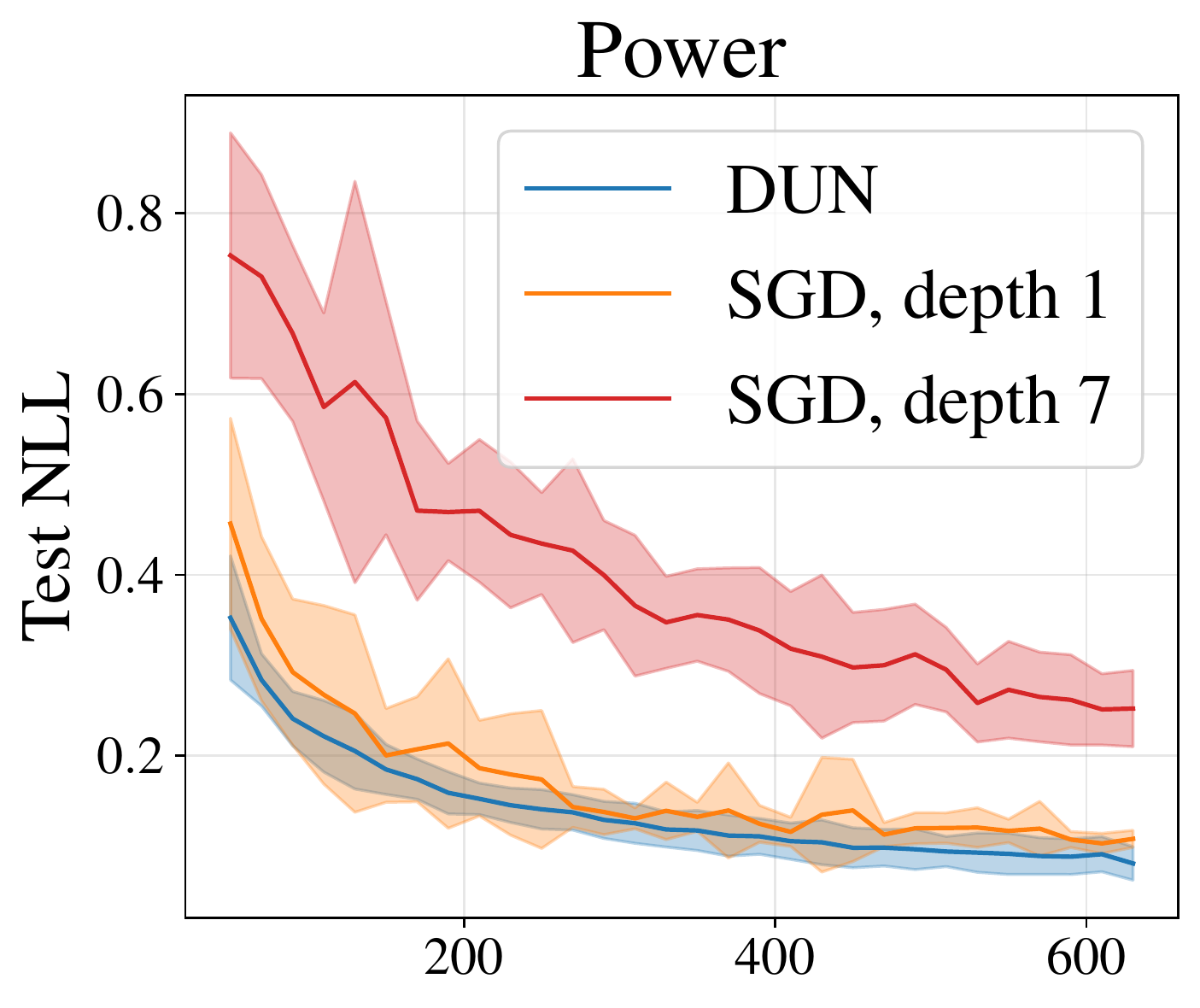}
    \end{subfigure}
    \begin{subfigure}{0.32\textwidth}
        \centering
        \includegraphics[width=\linewidth]{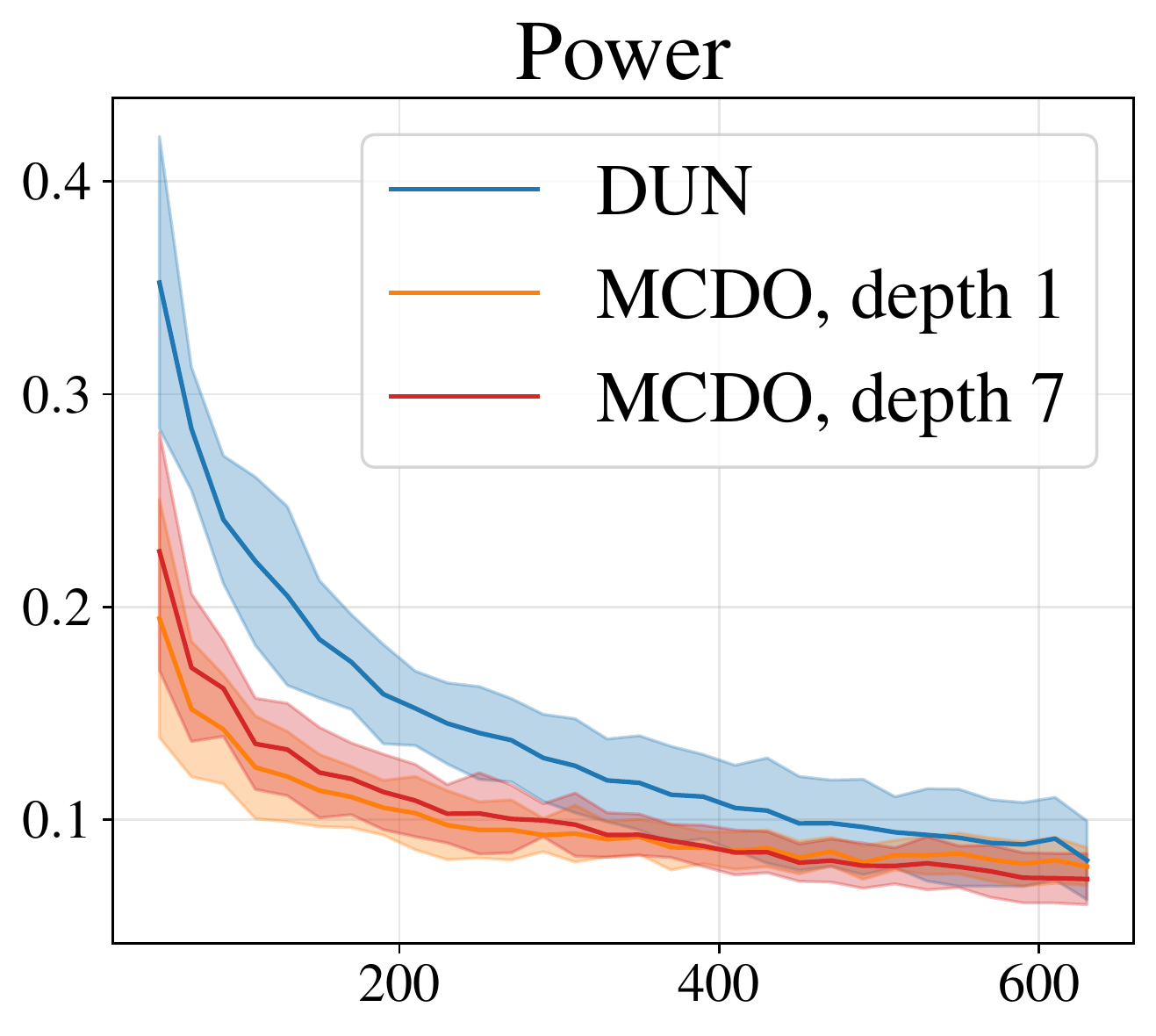}
    \end{subfigure} 
    \begin{subfigure}{0.32\textwidth}
        \centering
        \includegraphics[width=\linewidth]{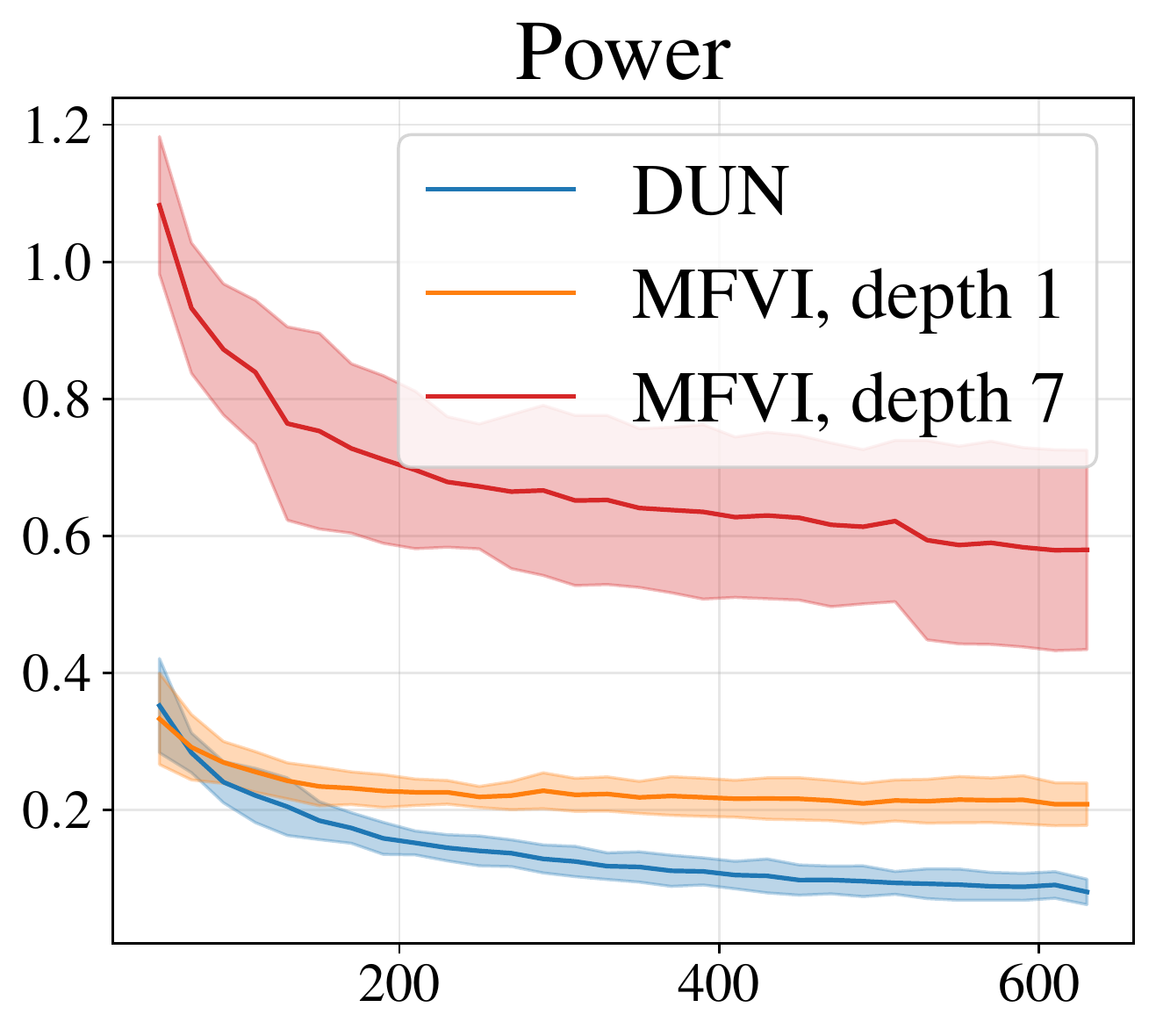} 
    \end{subfigure} \\
    \begin{subfigure}{0.33\textwidth}
        \centering
        \includegraphics[width=\linewidth]{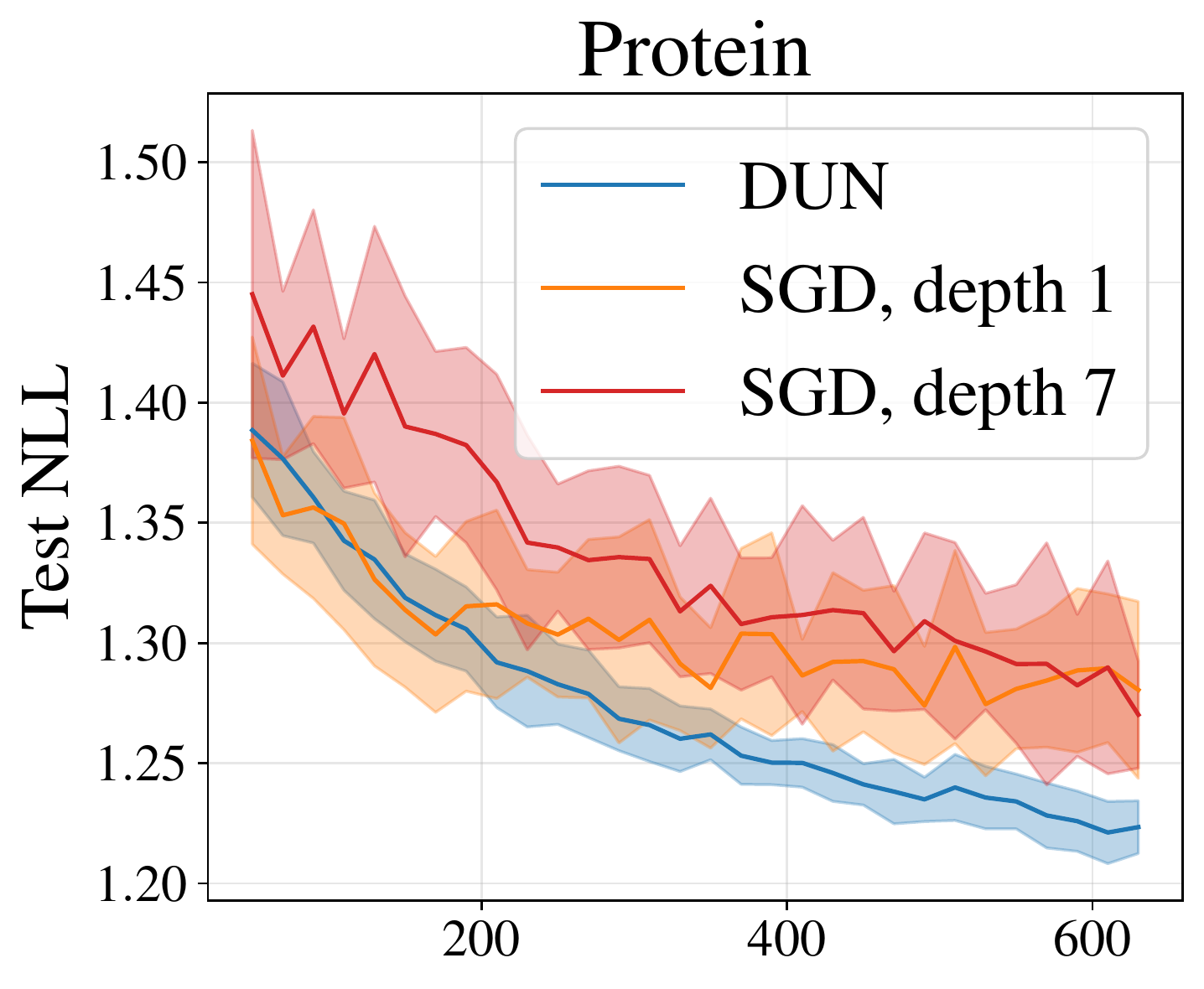}
    \end{subfigure}
    \begin{subfigure}{0.32\textwidth}
        \centering
        \includegraphics[width=\linewidth]{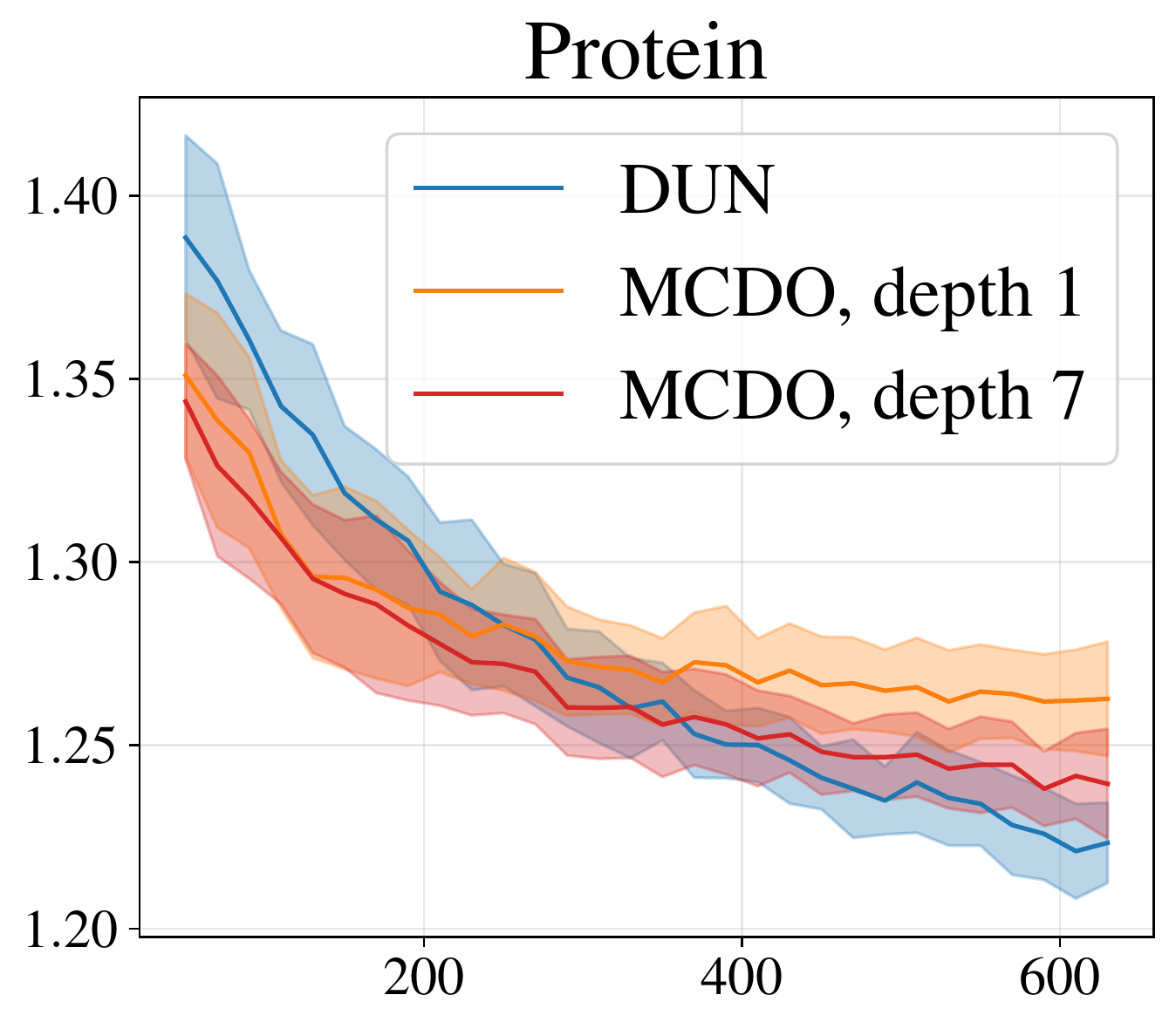}
    \end{subfigure} 
    \begin{subfigure}{0.32\textwidth}
        \centering
        \includegraphics[width=\linewidth]{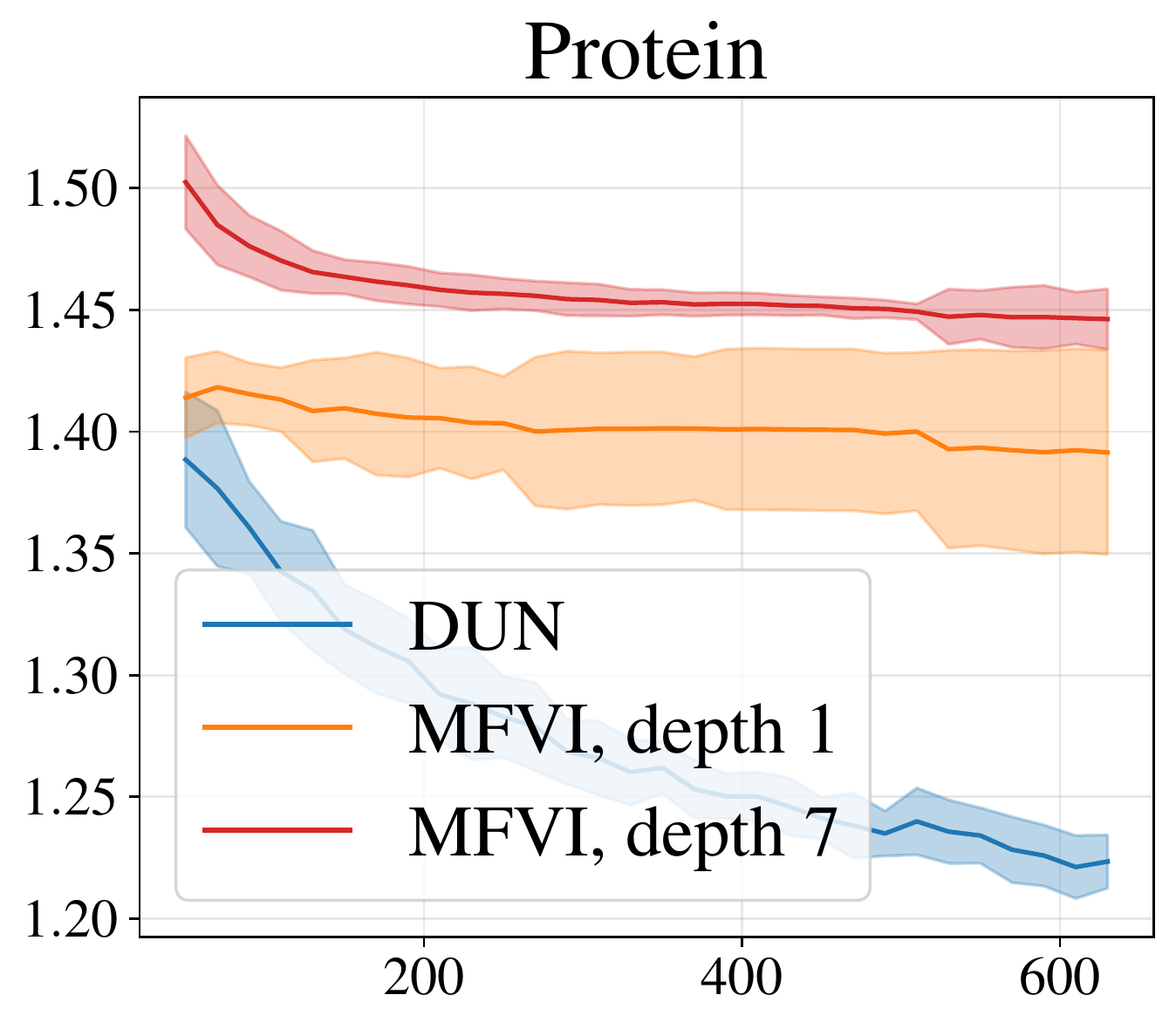} 
    \end{subfigure} \\
    \begin{subfigure}{0.33\textwidth}
        \centering
        \includegraphics[width=\linewidth]{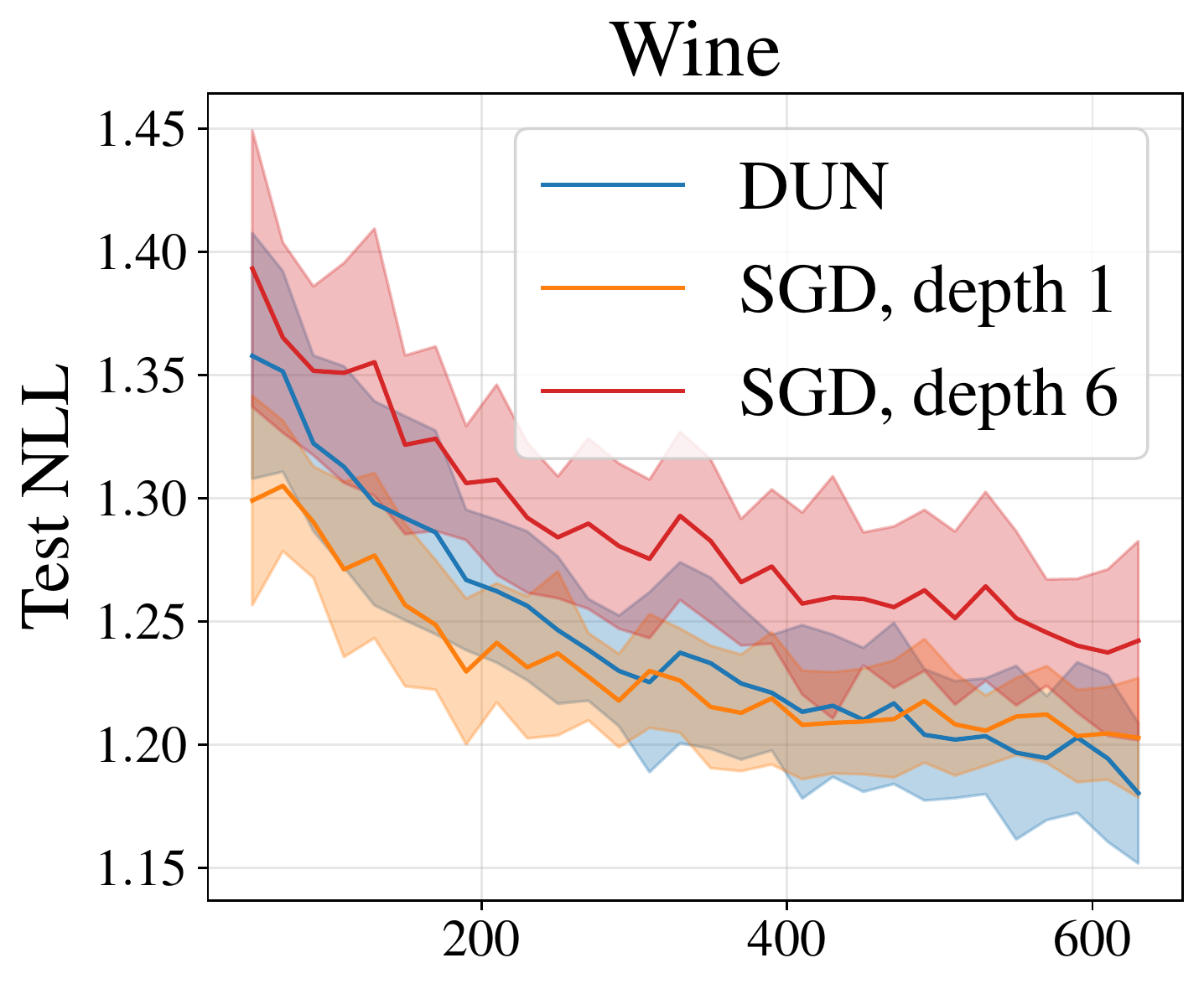}
    \end{subfigure}
    \begin{subfigure}{0.32\textwidth}
        \centering
        \includegraphics[width=\linewidth]{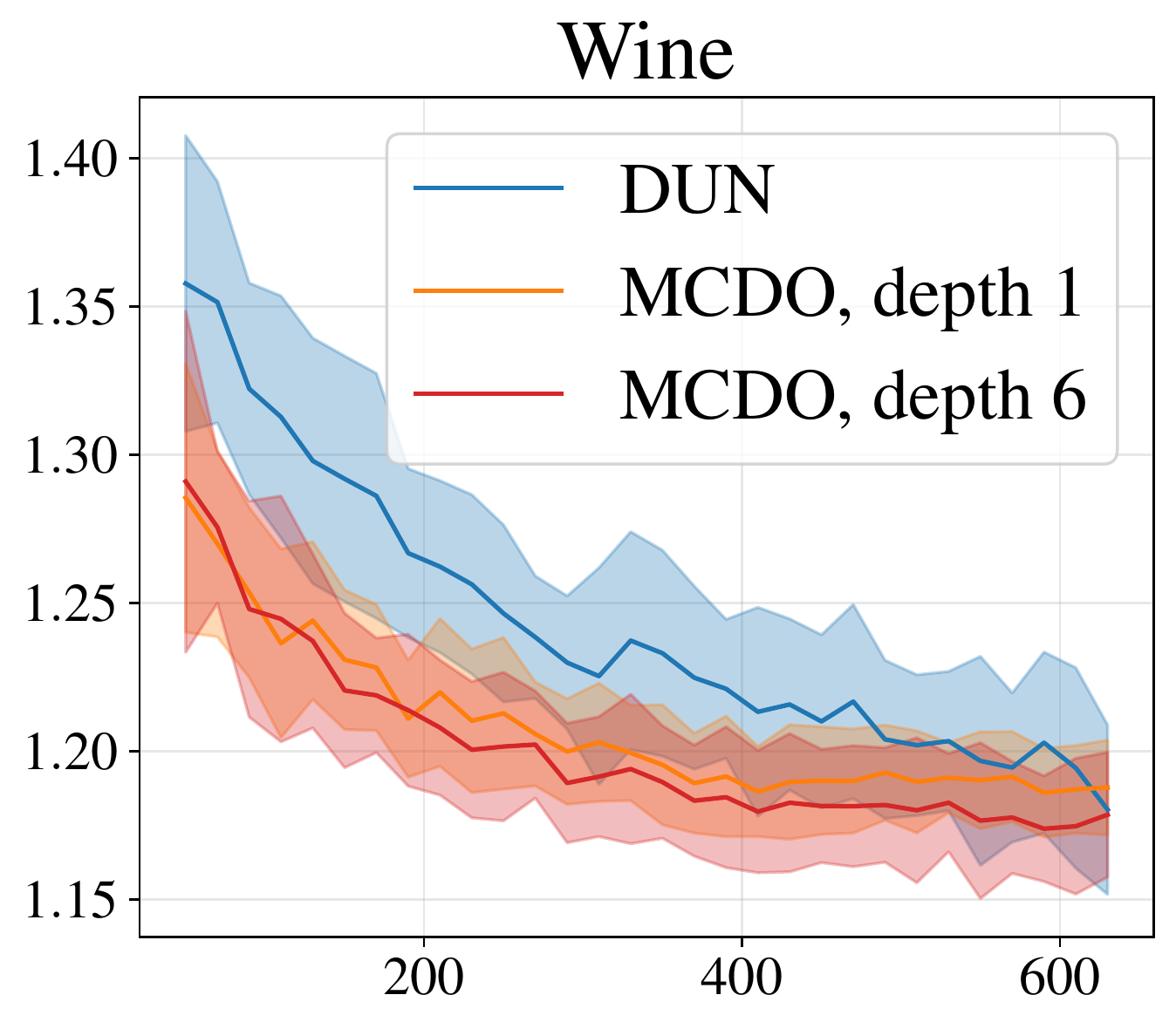}
    \end{subfigure} 
    \begin{subfigure}{0.32\textwidth}
        \centering
        \includegraphics[width=\linewidth]{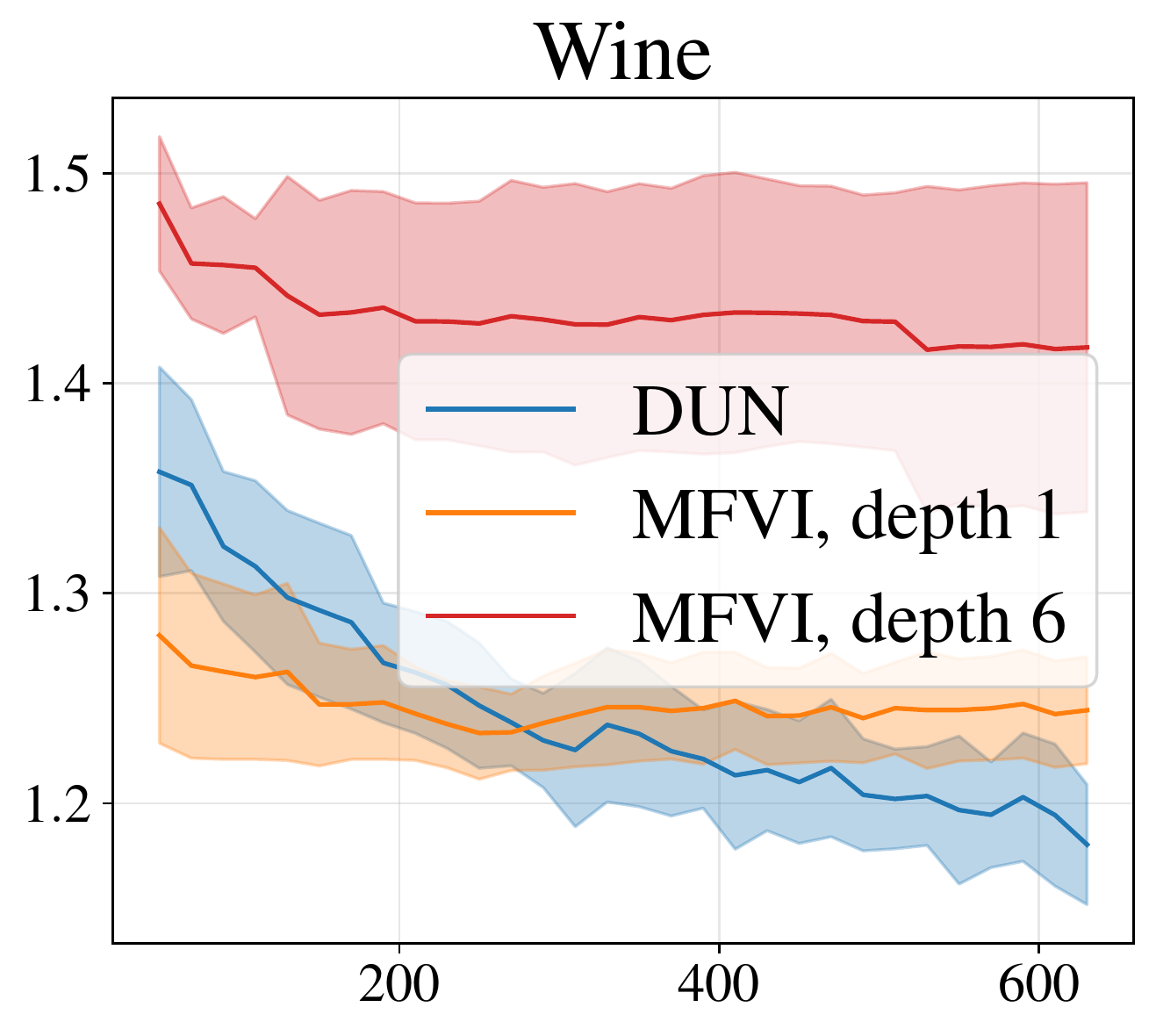} 
    \end{subfigure} \\
    \begin{subfigure}{0.33\textwidth}
        \centering
        \includegraphics[width=\linewidth]{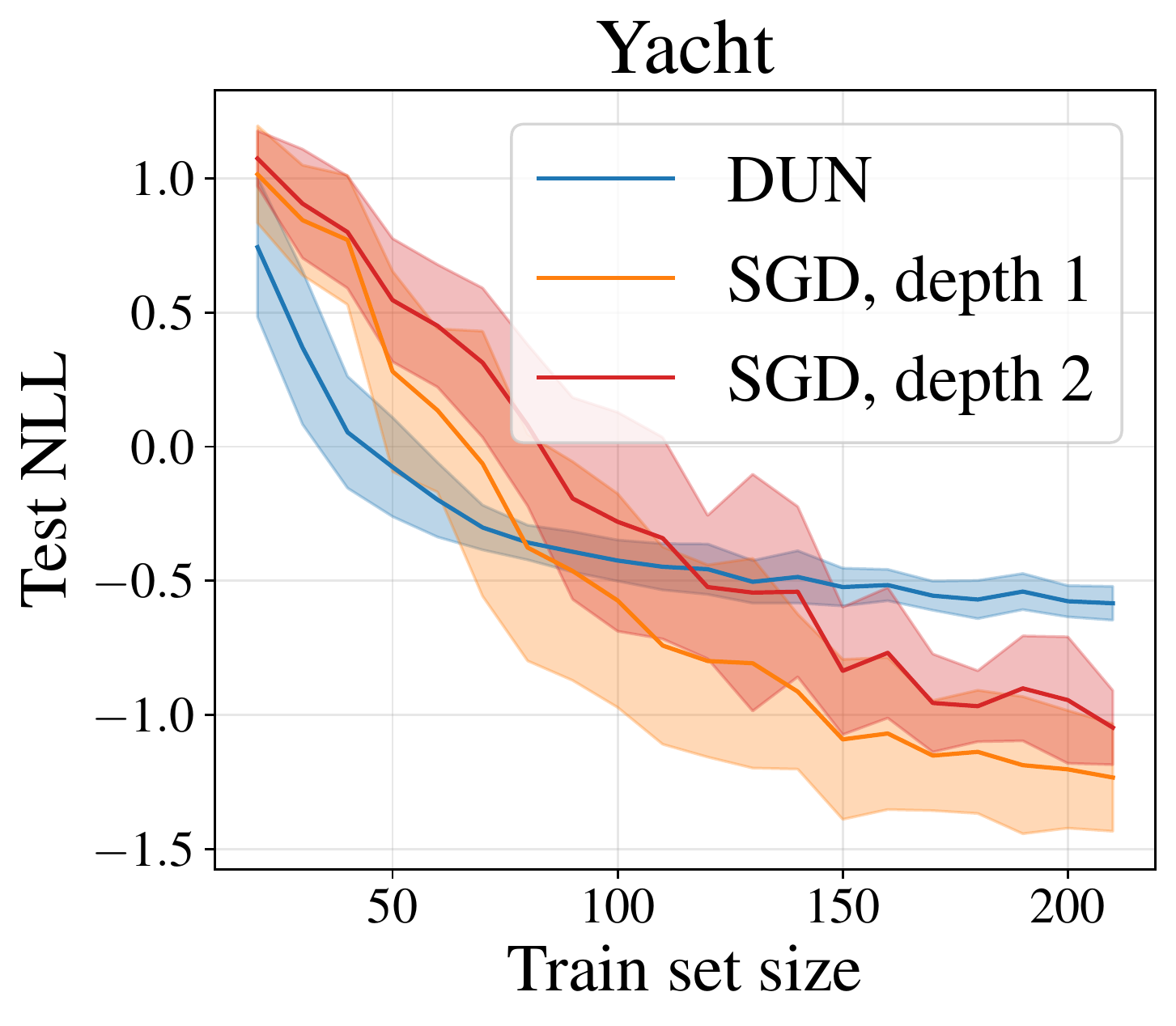}
    \end{subfigure}
    \begin{subfigure}{0.32\textwidth}
        \centering
        \includegraphics[width=\linewidth]{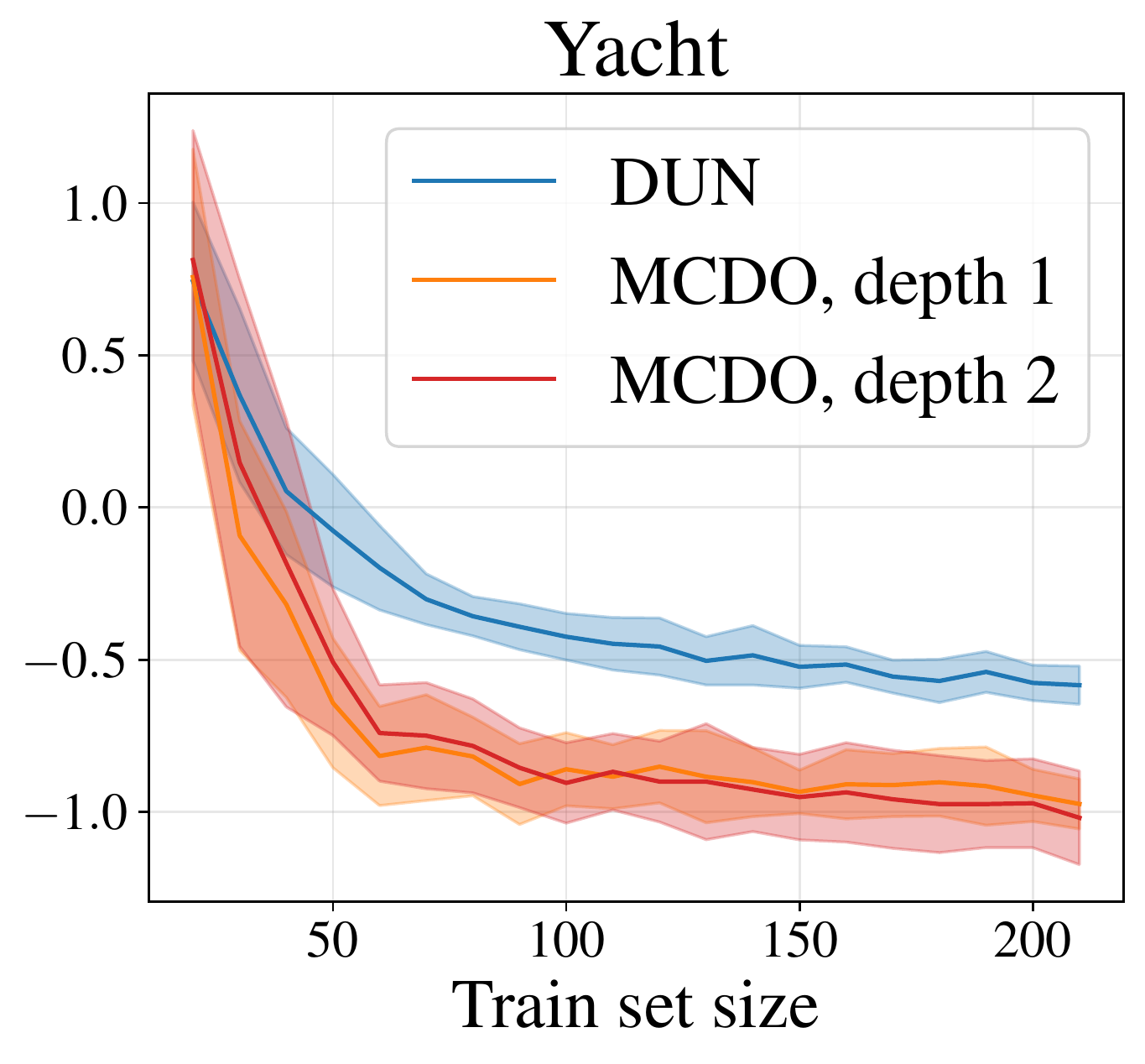}
    \end{subfigure} 
    \begin{subfigure}{0.32\textwidth}
        \centering
        \includegraphics[width=\linewidth]{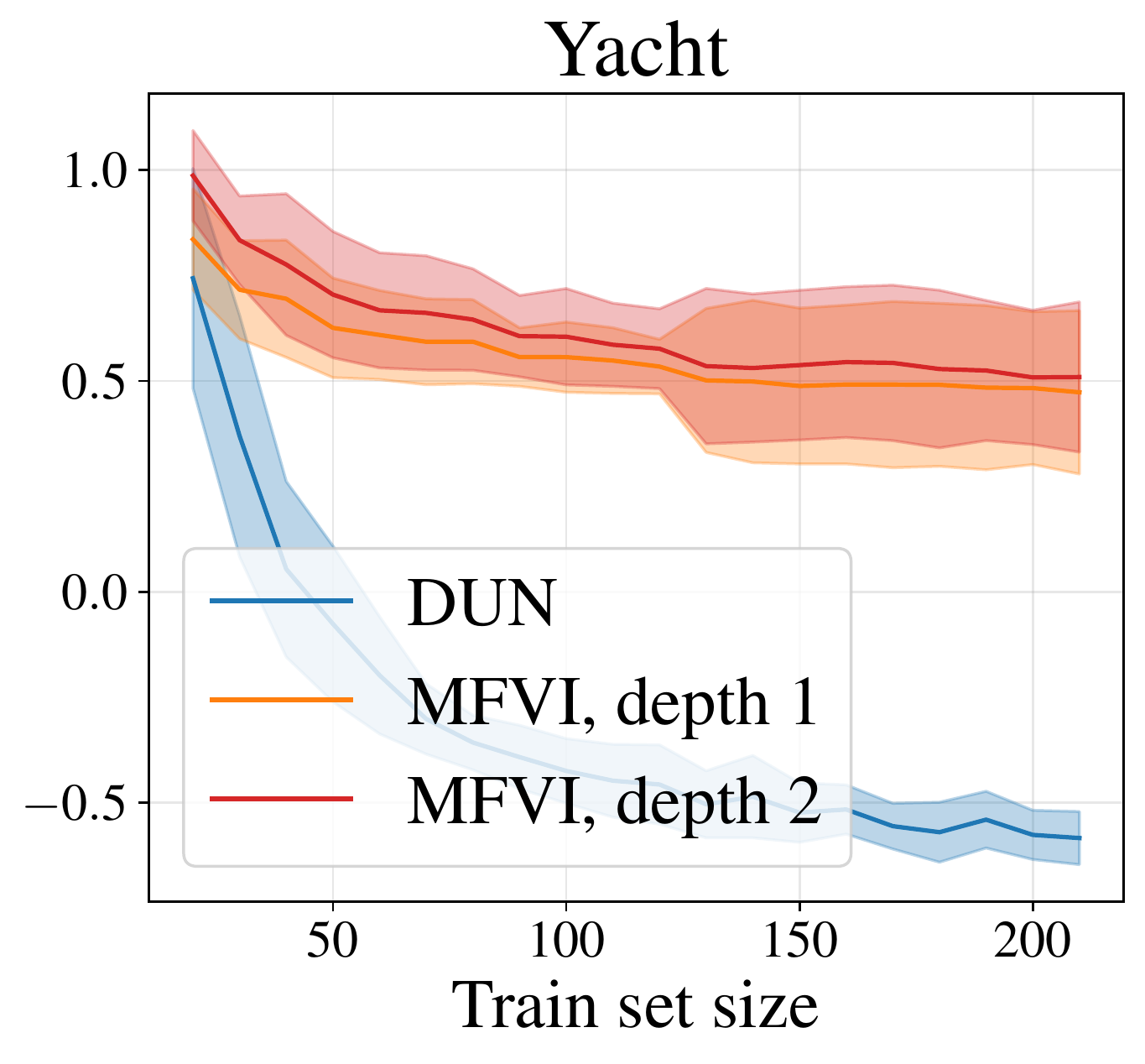} 
    \end{subfigure} \\
    \caption{NLL vs. number of training points evaluated on UCI datasets. The performance of SGD, MCDO and MFVI models with a single hidden layer and with the optimal number of layers found at the end of active learning by the DUN is compared to the performance of DUNs. A random acquisition strategy is used.}
    \label{fig:app_depths}
\end{figure}

\end{document}